%% file: main.tex
\documentclass[]{sensenova}

\usepackage{hyperref}
\usepackage{cleveref}
\usepackage{verbatim}

\usepackage{wrapfig}
\usepackage{graphicx}
\usepackage{floatrow}
\usepackage{placeins}
\usepackage{subcaption}
\usepackage{listings}
\usepackage{algorithm}
\usepackage{subcaption}
\usepackage{dsfont}

\usepackage{mmstyles}

\usepackage[toc,page,header]{appendix}

\usepackage{xspace}
\usepackage[utf8]{inputenc}  % Read UTF-8
\usepackage[T1]{fontenc}     % Better font encoding for European languages
\usepackage{CJKutf8} 

\usepackage[misc]{ifsym}

\usepackage{CJKutf8} 
\usepackage[colorinlistoftodos,prependcaption,textsize=small]{todonotes}
\usepackage{fontawesome5}

\usepackage{soul} 
\definecolor{HLGreen}{rgb}{0.78,0.95,0.78}
\definecolor{HLRed}{rgb}{0.9725,0.8431,0.8549}   % #F8D7DA
\definecolor{benchblue}{RGB}{58, 122, 252}

\newcommand{\colornum}[1]{%
  \footnotesize
  \ifdim #1 pt < 0pt
    \textcolor{sensepurple}{#1}%
  \else
    \textcolor{benchblue}{#1}%
  \fi
}

\usepackage{cuted} 

\definecolor{bestcolor}{RGB}{219, 208, 237}
\definecolor{secondcolor}{RGB}{241, 237, 248}
\definecolor{thirdcolor}{RGB}{211, 222, 190}
\definecolor{line-blue}{RGB}{243, 248, 252}
\definecolor{line-green}{RGB}{200,242,200}
\definecolor{line-red}{RGB}{255,215,215}

\usepackage{adjustbox}
\usepackage{makecell}
\usepackage{booktabs}

\newlength{\ModelW}
\newlength{\NonModelW}
\newcolumntype{M}{>{\centering\arraybackslash}p{\NonModelW}}
\usepackage{graphicx}
\usepackage{subcaption}
\graphicspath{{assets/figures/}}

\usepackage{array}      % m{<width>}
\usepackage{caption}
\usepackage{subcaption} % 提供 subfigure
\usepackage{enumitem}

\newcommand{\df}[1]{\textcolor{RoyalBlue}{{\footnotesize #1}}}
\newcommand{\ddf}[1]{\textcolor{TealBlue}{{\footnotesize #1}}}
\usepackage{multirow}
\usepackage{makecell}

\usepackage{booktabs,tabularx,array}
\newcolumntype{Y}{>{\centering\arraybackslash}X}
\newcolumntype{L}{>{\raggedright\arraybackslash}X}

\usepackage{framed}
\usepackage{algorithm}
\usepackage{algorithmic}
\usepackage{listings}
\usepackage{soul}          % 高亮/下划线等

\definecolor{cotredbg}{RGB}{255,220,220}   % 淡红
\definecolor{cotgreenbg}{RGB}{220,255,220} % 淡绿
\newcommand{\cotfailhl}[1]{%
  {\sethlcolor{cotredbg}\hl{#1}}%
}
\newcommand{\cotsuchl}[1]{%
  {\sethlcolor{cotgreenbg}\hl{#1}}%
}

\newcommand{\name}{SenseNova-SI\xspace}
\newcommand{\benchname}{EASI\xspace}
\newcommand{\Supp}{Supplementary Material}

\title{Scaling Spatial Intelligence with Multimodal Foundation Models}

\author{
Zhongang Cai$^{*,1}$, Ruisi Wang$^{*,1}$, Chenyang Gu$^{*,1}$, Fanyi Pu$^{*,1,2}$, Junxiang Xu$^{*,1}$, Yubo Wang$^{*,1}$, \\ 
Wanqi Yin$^{*,1}$, Zhitao Yang$^{*,1}$, Chen Wei$^{*,1}$, Qingping Sun$^{*,1}$, Tongxi Zhou$^{*,1}$, Jiaqi Li$^{*,1}$,\\ 
Hui En Pang$^{*,2}$, Oscar Qian$^{*,1,2}$, Yukun Wei$^{1}$, Zhiqian Lin$^{1}$, Xuanke Shi$^{1}$, Kewang Deng$^{1}$, \\
Xiaoyang Han$^{1}$,  Zukai Chen$^{1}$, Xiangyu Fan$^{1}$, Hanming Deng$^{1}$, Lewei Lu$^{1}$, Liang Pan$^{1}$, \\ 
Bo Li$^{2}$, Ziwei Liu$^{\textrm{\Letter},2}$, Quan Wang$^{\textrm{\Letter},1}$, Dahua Lin$^{\textrm{\Letter},1}$, Lei Yang$^{*,\textrm{\Letter},1}$ 
\parbox{\textwidth}{\centering\small
    $*$ Core Contributors,
    $\textrm{\Letter}$ Corresponding Authors \\
    $^1$SenseTime Research,
    $^2$Nanyang Technological University
}}

\input{sec/0_abstract} 

\begin{document}
\maketitle

\input{sec/1_intro}
\input{sec/2_related_works}
\input{sec/3_data}

\input{sec/4_training}
\input{sec/5_results}
\input{sec/6_conclusion}

\clearpage

\bibliographystyle{plainnat}
\bibliography{main}

\clearpage

% %%%%%%%%%% Appendix %%%%%%%%%%

\beginappendix
\input{sec/x_suppl}

\end{document}

%% file: sec/0_abstract.tex
\abstract{

Despite remarkable progress, multimodal foundation models still exhibit surprising deficiencies in spatial intelligence. In this work, we explore scaling up multimodal foundation models to cultivate spatial intelligence within the \textbf{\name} family\textbf{*}, built upon established multimodal foundations including visual understanding models (\ie, Qwen3-VL and InternVL3) and unified understanding and generation models (\ie, Bagel). We take a principled approach to constructing high-performing and robust spatial intelligence by systematically curating \name-8M: eight million diverse data samples under a rigorous taxonomy of spatial capabilities. \name demonstrates unprecedented performance across a broad range of spatial intelligence benchmarks: 68.8\% on VSI-Bench, 43.3\% on MMSI, 85.7\% on MindCube, 54.7\% on ViewSpatial, 47.7\% on SITE, 63.9\% on BLINK, 55.5\% on 3DSR, and 72.0\% on EmbSpatial, while maintaining strong general multimodal understanding (\eg, 84.9\% on MMBench-En). More importantly, we analyze the impact of data scaling, discuss early signs of emergent generalization capabilities enabled by diverse data training, analyze the risk of overfitting and language shortcuts, present a preliminary study on spatial chain-of-thought reasoning, and validate the potential downstream application. 
All newly trained multimodal foundation models are publicly released.

\textit{\small* This report is based on the \textbf{v1.1} version of \name.}

\checkdata[Codebase]{\url{https://github.com/OpenSenseNova/SenseNova-SI}}
\checkdata[Models]{\url{https://huggingface.co/collections/sensenova/sensenova-si}} 
}

%% file: sec/1_intro.tex
\section{Introduction}
\label{sec:intro}
\input{figures/teaser}

% Context and research gap
In recent years, multimodal foundation models~\cite{Qwen3-VL, zhu2025internvl3, deng2025bagel} have achieved groundbreaking progress across a wide spectrum of tasks. However, it has become evident that even the most advanced models still struggle with spatial intelligence: the ability to understand, reason about, and act within three-dimensional space, which is fundamental to embodied AGI that can perceive, adapt to, and interact with the physical world. Interestingly, such tasks are often considered trivial for humans~\cite{cai2025has}. One of the key limitations lies in the scarcity and imbalance of spatially grounded data. While recent efforts have introduced a surge of large-scale datasets targeting various facets of spatial reasoning, these resources remain fragmented and heterogeneous in scope and quality. Consequently, the community is still in the early stages of understanding how multimodal foundation models acquire and develop spatial intelligence, and what strategies are effective in fostering this capability.

% Our approach
In this work, we aim to provide timely insights into cultivating spatial intelligence within state-of-the-art multimodal foundation models by leveraging their powerful generalist backbones and scaling up diverse data collections. Our study investigates the data scaling laws of spatial intelligence through extensive experiments on the widely adopted InternVL3 multimodal foundation model family~\cite{zhu2025internvl3}, and further extends the analysis to Qwen3-VL~\cite{Qwen3-VL} as well as Bagel~\cite{deng2025bagel}, a unified understanding and generation model.
We envision the resulting models, denoted by the \textbf{\name} prefix, as open research platforms to advance studies in spatial intelligence. To preserve compatibility with existing research pipelines, we deliberately avoid altering the original architectures of the base models. Instead, we adopt a data-centric approach, emphasizing the role of data scaling and training strategies as the primary drivers of spatial understanding capability.
Our systematic collection and synthesis of spatial data are guided by a principled taxonomy of fundamental spatial intelligence capabilities~\cite{cai2025has}, resulting in \textit{eight million} samples (named \name-8M) spanning five key domains: Metric Measurement (MM), Spatial Relations (SR), Mental Reconstruction (MR), Perspective-taking (PT), and Comprehensive Reasoning (CR). 
We analyze a diverse collection of public datasets for spatial intelligence, followed by strategic further scaling that places a special focus on perspective-taking, an underrepresented capability that is critical to spatial intelligence, while isolated from general multimodal capabilities~\cite{li2024core}.

% Results
We evaluate the \name foundation models across a broad suite of benchmarks, including VSI-Bench~\cite{vsi}, MMSI~\cite{yang2025mmsi}, MindCube~\cite{mindcube}, ViewSpatial~\cite{li2025viewspatial}, SITE~\cite{wang2025site}, BLINK~\cite{fu2024blink}, 3DSR~\cite{ma20243dsrbench}, and EmbSpatial~\cite{du2024embspatial}, following continued training on our comprehensive spatial intelligence data collection. The models achieve state-of-the-art performance among open-source models of comparable sizes, with the best performance achieving 68.8\% on VSI-Bench, 43.3\% on MMSI, 85.7\% on MindCube, 54.7\% on ViewSpatial, 47.7\% on SITE, 63.9\% on BLINK, 55.5\% on 3DSR, and 72.0\% on EmbSpatial, while retaining their original strengths on general multimodal understanding benchmarks such as MMBench-En (84.9\%). Our analysis reveals several key findings:
(1) Scaling law of spatial intelligence. We systematically investigate how spatial intelligence scales under mixed data regimes. Our analysis reveals distinct scaling behaviors across spatial capabilities and model sizes, and suggests that the observed saturation trends may signal that future advances require paradigm shifts built upon and beyond \name.
(2) Emergent generalization through diverse data. We  report surprising findings that point to early signs of emergent spatial intelligence: models trained on one set of spatial tasks exhibit nontrivial transfer to seemingly unrelated tasks, and demonstrate extrapolation to longer spatial contexts beyond the training distribution.
(3) Robustness against overfitting and shortcuts. Through controlled experiments and circular test designs, we rigorously validate that \name genuinely acquires spatial capabilities rather than exploiting memorization, annotation biases, or unintended shortcuts in the training data.
(4) Spatial chain-of-thought (CoT) may not be effective. We construct and evaluate three representative text CoT schemes and reinforcement learning, but find that they cannot reliably improve spatial reasoning beyond what is achieved through simple QA-style data scaling. These results suggest that extending text-based CoT paradigms to spatial intelligence is non-trivial and may require fundamentally different reasoning mechanisms. 
(5) Downstream task validation. To assess the practical utility of \name, we apply \name to robotic manipulation tasks without any finetuning, and achieve notable performance improvements on EmbodiedBench~\cite{yang2025embodiedbench}, demonstrating the potential of \name as a foundation for embodied AI.

% Summary
In summary, we introduce the \name series of multimodal foundation models, which achieve new state-of-the-art performance across major spatial intelligence benchmarks. Our study further validates that data scaling governs the progression of spatial intelligence. We envision \name as a strong, robust baseline that future research can build upon to drive deeper advances in this critical field.

%% file: figures/teaser.tex
% \twocolumn[{
%     \renewcommand\twocolumn[1][]{#1}
%     \maketitle
%     \vspace{-25pt}
%     \begin{center}
%         \includegraphics[width=1\textwidth]{figures/images/teaser_v5.pdf}
%         \vspace{-6mm}
%         \captionof{figure}{
%         Guided by taxonomy of spatial intelligence~\cite{cai2025has}, we scaled spatial data to construct \textbf{\name-8M}.
%         Leveraging \textbf{\name-8M}, we investigated the impact of data scaling on cultivating spatial capabilities in various MLLMs.
%         Across core spatial capabilities, \textbf{\name} achieved comprehensive improvements, surpassing open-source models and even outperforming GPT-5 in specific spatial abilities, such as Perspective-taking.
%         In terms of benchmarks, \textbf{\name} achieves SOTA on 5 recent spatial intelligence benchmarks while maintaining strong generalization.
%         }
%         \label{fig:teaser}
%     \end{center}
% }]

\begin{figure*}[t]
    \centering
    \setlength{\belowcaptionskip}{-2pt}
    \includegraphics[width=1\textwidth]{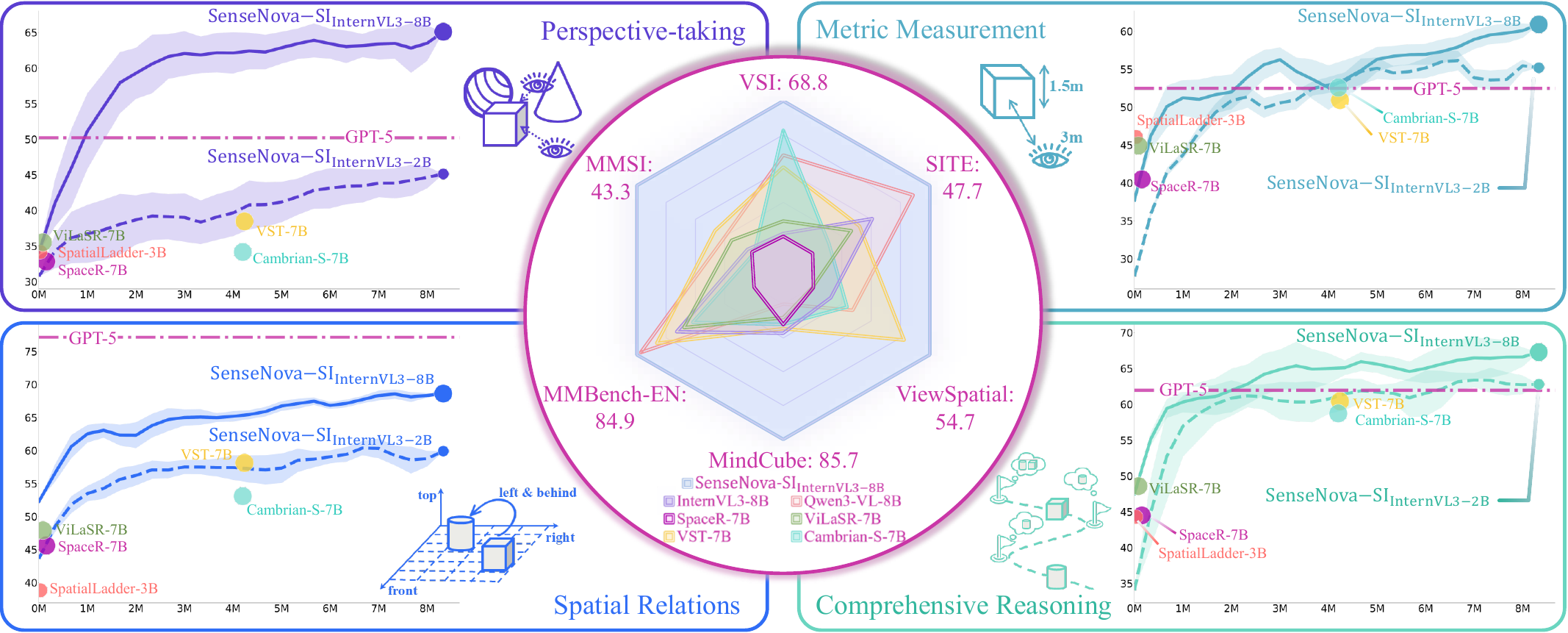}
    \caption{
        Guided by taxonomy of spatial intelligence~\cite{cai2025has}, we scaled spatial data to construct \textbf{\name-8M}, which we leverage to investigate the impact of data scaling on cultivating spatial capabilities in various MLLMs.
        The four subfigures at the corners elaborate \textbf{\name}'s performance on four core spatial capabilities (\ie, Perspective-taking, Spatial Relations, Metric Measurement, and Comprehensive Reasoning). Through data scaling, \name surpasses open-source models and even outperforms GPT-5 in specific spatial abilities, such as Perspective-taking.
        The lines denote the average performance across benchmark subtasks within each capability, while the shaded regions (confidence bands) represent $\pm0.5$ standard deviation. 
        At center, we show \textbf{\name} achieves state-of-the-art (SoTA) results on five recent spatial intelligence benchmarks (VSI, MMSI, MindCube, ViewSpatial, and SITE) while maintaining strong performance on a general multimodal benchmark (MMBench-En).
    }
    \label{fig:teaser}
\end{figure*}

%% file: sec/2_related_works.tex
\section{Related Works}
\label{sec:related_works}

\subsection{Multimodal Foundational Models}

Recent studies~\cite{cai2025has,yu2025far,li2024core} reveal that while models like GPT-5 demonstrate strong planar reasoning capabilities, they still lag significantly behind humans in Spatial Intelligence (SI).
Furthermore, \benchname~\cite{cai2025has} shows that the performance gap between open-source and closed-source models on SI tasks is relatively small.
These findings motivate us to enhance the spatial intelligence of widely used open-source models (\eg, QwenVL series~\cite{Qwen-VL,Qwen2-VL,Qwen2.5-VL,Qwen3-VL} and InternVL series~\cite{chen2024internvl,zhu2025internvl3,wang2025internvl3}).
This not only enables fairer comparisons among models of similar scale but also facilitates the community’s direct use of our models for downstream tasks, (\eg, VLA~\cite{kim2024openvla,zitkovich2023rt,yang2025embodiedbench}), with minimal substitution costs.

\subsection{Multimodal Models for Spatial Intelligence}

% # 3D expert encoders
% - SpatialMLLM
% - VLM-3R
% Very recent works: 3D Thinker
% # Data-centric
% - SpatialVLM (not compared, but first work)
% - SpatialLadder
% - SpaceR
% - ViLaSR
% - Very recent works: VST, Cabriam-S

Efforts to enhance spatial intelligence in multimodal models primarily follow two approaches: \textit{leveraging 3D experts} or \textit{curating spatial-specific datasets}.
As spatial intelligence is inherently linked to 3D vision, an intuition is to employ 3D expert encoders that infer key 3D attributes from images ~\cite{wu2025spatial,wang2025ross3d,chen2025think}.
Spatial-MLLM~\cite{wu2025spatial} incorporates VGGT~\cite{wang2025vggt} as an input-level encoder to capture 3D information, while VLM-3R~\cite{vlm3r} integrates 3D information using combined geometry and camera-view tokens.
Recently, 3DThinker~\cite{chen2025think} aligns model-generated 3D features with VGGT-derived supervision at the output level.
Conversely, some studies~\cite{chen2024spatialvlm,cheng2024spatialrgpt,xu2025multi,vilasr}
% argue that specialized 3D encoders are unnecessary for achieving strong spatial intelligence.
inject visual-spatial knowledge through dataset curation and training paradigm.
SpatialVLM~\cite{chen2024spatialvlm} pioneered this direction by synthesizing 2B VQA samples focused on two-object spatial relationships.
% TODO: maybe highlight MindCube a litte bit
% SpaceR~\cite{ouyang2025spacer} and MindCube~\cite{yin2025spatial(mindcube)} enhance spatial reasoning by incorporating textual cognition maps into reasoning traces.
SpaceR~\cite{ouyang2025spacer} uses RL for spatial reasoning, while MindCube~\cite{mindcube} explores SFT and RL using QA and two types of cognitive maps.
%
% ViLaSR~\cite{wu2025reinforcing(vilasr)} employs RL to cultivate tool-integrated reasoning, enabling iterative thinking and drawing operations.
%
SpatialLadder~\cite{li2025spatialladder} constructs a dataset with 26K samples and introduces a three-stage progressive training strategy.
%: object localization for spatial perception, multi-dimensional tasks for spatial understanding, and RL for reasoning.
%
% Concurrently, VST~\cite{vst2025} adopts a two-phase training approach, using the VST-P dataset (4.1M samples) for SFT on spatial perception and the VST-R dataset (135K samples) for RL on spatial reasoning.
Concurrently, VST~\cite{vst2025} adopts a two-phase training approach, using 4.1M samples for SFT on spatial perception and 135K samples for RL on spatial reasoning.
Cambrian-S~\cite{yang2025cambrian} develops VSI-590K dataset and employs a four-stage training framework to progressively enhance spatial video understanding.
In this work, we systematically scale datasets targeting core spatial capabilities~\cite{cai2025has}, addressing key gaps in existing datasets, particularly the previously overlooked perspective-taking tasks.
%
% In addition, we explore the impact of scaling spatial data on  multimodal models, aiming to establish a more comprehensive and robust foundation for spatial intelligence.

%% file: sec/3_data.tex
\section{Data}
\label{sec:data}

The limitations in spatial intelligence mainly stem from high-quality, diverse data scarcity. 
In this work, we strategically scale data to expand coverage toward holistic spatial intelligence, rather than merely increasing data volume.

\subsection{Task Taxonomy}
We adopt a principled approach, following the EASI~\cite{cai2025has} protocol to decompose spatial intelligence into fundamental capabilities. We focus on five key capabilities closely aligned with real-world scenarios. For each capability, we analyze the underlying cognitive operations and derive corresponding tasks to ensure comprehensive coverage. \Cref{fig:data} illustrates the dataset constructed under this taxonomy.

\input{figures/data}
\clearpage

\noindent \textbf{Metric Measurement (MM).}
MM involves a basic understanding of the physical scale and typical object sizes. We include distances estimation between the camera and objects and pairs of objects, and size estimation across scales from individual objects to entire scenes. 

\noindent \textbf{Spatial Relations (SR).}
We define SR as the ability to impose and reason within a 3D coordinate system. In egocentric, local level of view, it unfolds into front–back, left–right, and up–down relations between subjects. In global, scene level, these relations extend to near–far and relative scale (large–small) comparisons. 

\noindent \textbf{Mental Reconstruction (MR).}
MR focuses on inferring 3D object structure from limited 2D observations. 
We adopt a diagnostic task, which identifies which side of an object is visible. This requires the integration of sparse 2D cues to infer 3D geometry and align views in a canonical object-centric frame.

\noindent \textbf{Perspective-taking (PT).}
PT addresses reasoning with changing camera viewpoints. We construct PT tasks in a progressively more challenging hierarchy:

\begin{itemize}
    \item \textit{View Correspondence.} Establish correspondences of points or objects across views, recognizing entities under changes in viewpoint, scale, and occlusion.
    \item \textit{Camera Motion Reasoning.} Infer relative camera motion between views, linking appearance changes to 3D transformations.
    \item \textit{Allocentric Transformation.} Simulate viewpoint shifts and express spatial relations across coordinate systems, including camera, object-target, and self-oriented views.
\end{itemize}

This layered design ensures that PT goes beyond pattern matching across images, encouraging the model to build internal representations of how observations transform with viewpoint changes.

\noindent \textbf{Comprehensive Reasoning (CR).}
CR tasks involve coordinating multiple spatial capabilities with extended memory and multi-step reasoning. 
Such data is scarce and often limited to simple scenarios. 
As these tasks lie beyond our main goal of scaling spatial QA and core spatial capabilities, we reuse existing datasets as a lightweight complement.

\subsection{Data Sources.}

\noindent \textbf{General QA.}
We collect a set of open-source general-purpose QA datasets for 2D image understanding. Specifically, we use VSR~\cite{VSR}, SPEC~\cite{SPEC}, GQA~\cite{hudson2019gqa}, VQA~\cite{Antol2015vqa}, and IconQA~\cite{lu2021iconqa}, resulting in about 0.6M QA pairs.

\noindent \textbf{Community Datasets on Spatial Intelligence.}
Among existing open-source resources, we identify several datasets that focus on spatial reasoning, including Open3D-VQA~\cite{zhang2025open3dvqa}, CLEVR-series~\cite{johnson2017clevr}, REL3D~\cite{goyal2020rel3d}, SAT~\cite{ray2024sat}, GRiD-3D~\cite{grid-3d}, MultiSpa~\cite{xu2025multi}, MindCube~\cite{mindcube}, ViCA~\cite{feng2025towards}, VLM-3R~\cite{vlm3r}, and VSI-590K~\cite{yang2025cambrian}. We incorporate all of these datasets, yielding in total about 3.3M QA pairs.

\noindent \textbf{Further Scaling on Spatial Intelligence.}
Building on these open-source data, we find gaps in task coverage and data imbalance. MM and SR dominate the data, while PT and MR remain underrepresented. For point, object, scene level correspondence, only MultiSpa provides point level QAs. Camera motion is also mostly limited to MultiSpa. Allocentric viewpoint transformation, especially object-centric and hypothetical views, is largely unexplored, as real-world QA labels are scarce. Tasks such as object reconstruction remain unaddressed.

To address these gaps, we leverage richly annotated, scene-diverse 3D datasets, including MessyTable~\cite{cai2020messytable}, ScanNet~\cite{dai2017scannet}, ScanNet++~\cite{yeshwanth2023scannet++}, SUN RGB-D~\cite{song2015sun}, CA-1M~\cite{lazarow2025cubify}, Ego-Exo4D~\cite{grauman2024ego}, and Matterport3D~\cite{chang2017matterport3d}, to generate large-scale, accurate and task-balanced QA pairs. This scaling process contributes 4.5M data, increasing the overall corpus size to 8.5M QA pairs.

%% file: figures/data.tex
\begin{figure}[t]
    \centering
    \setlength{\abovecaptionskip}{-2pt} 
    \setlength{\belowcaptionskip}{-4pt}
    \includegraphics[width=1\linewidth]{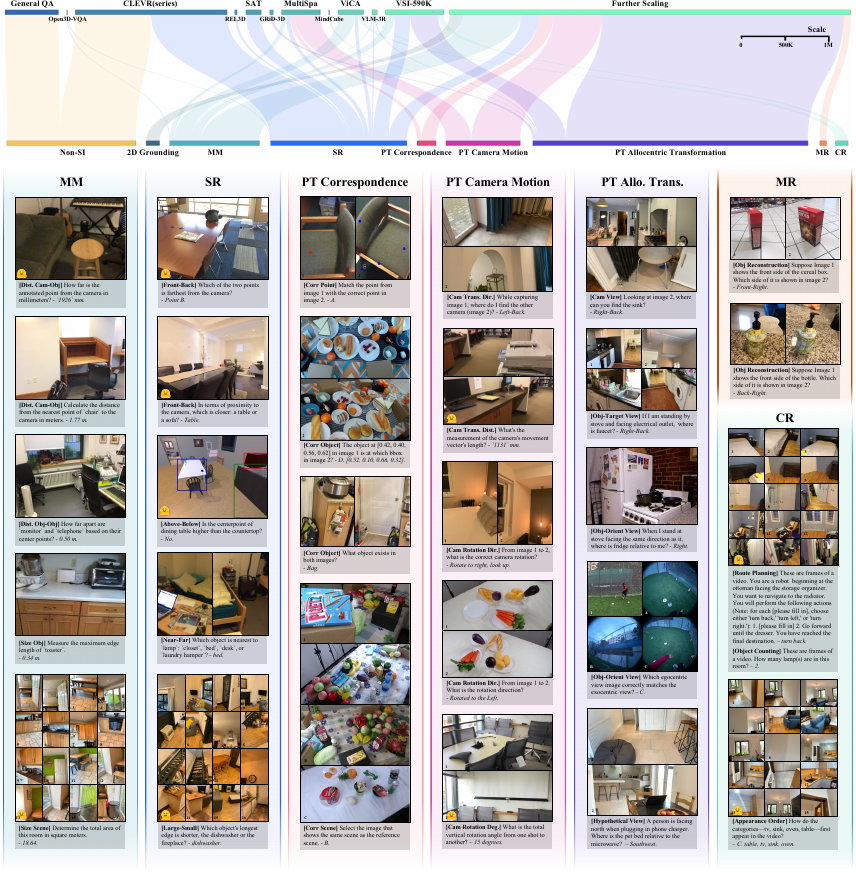}
    \caption{
        \textbf{\name-8M} reorganizes 4M open-source samples and scales up an additional 4.5M samples, following a taxonomy of fundamental spatial capabilities~\cite{cai2025has}.
        It covers general visual understanding (Non-SI), 2D grounding, and five core spatial capabilities: Metric Measurement (MM), Spatial Relationship (SR), Perspective-Taking (PT), Mental Reconstruction (MR), and Comprehensive Reasoning (CR).
        Notably, \name-8M explicitly incorporates Perspective-Taking (PT), which has been largely overlooked in prior datasets.
        The mapping from each data source to the corresponding spatial capabilities is illustrated at the top (with the scale in the upper-right corner indicating the number of QA pairs), while representative samples are organized by capability below.
        The ``Hugging Face'' icon denotes community datasets, whereas the remaining data sources are curated to support further scaling.
    }
    \label{fig:data}
\end{figure}

%% file: sec/4_training.tex
\section{Training}
\label{sec:training}

We adopt three multimodal foundation models in this study.

\noindent 
\textbf{Qwen3-VL~\cite{Qwen3-VL}} is the most capable multimodal model in the Qwen series to date. It adopts a strategy to scale from language foundation, that expand a strong LLM foundation to handle vision or audio modalities.

\noindent
\textbf{InternVL-3}~\cite{zhu2025internvl3} is natively multimodal, training vision and language jointly from scratch, thus enables stronger cross-modal alignment, more efficient scaling, and improved visual–language reasoning.

\noindent
\textbf{Bagel}~\cite{deng2025bagel} represents a new paradigm of unified understanding and generation. We include it in our study to examine whether such unified architectures can acquire strong spatial understanding capabilities.

\noindent
\textbf{Training Scheme.} Each foundation model is trained for one epoch on the same dataset using 128 GPUs with batch size 2048. Each training takes approximately three days. We employ AdamW~\cite{loshchilov2017decoupled} with a learning rate of 5$\times$10$^{-6}$ for all model-training runs. Maximum 16 frames are sampled for video data.

%% file: sec/5_results.tex
\input{tables/main_results_spatial}

\section{Experiments}
\label{sec:results}

\subsection{Evaluation Benchmarks.}

To assess \name under a broad range of scenarios, we select five newly released benchmarks for a complementary coverage of spatial intelligence.
%
% \noindent 
\textbf{VSI-Bench}~\cite{vsi} targets \textit{video-based} visual-spatial reasoning, evaluating a model’s ability to perceive and understand the 3D layout of real indoor scenes over extended temporal contexts. During evaluation, we uniformly sample 32 frames from each video.
%
% \noindent 
\textbf{MMSI-Bench}~\cite{yang2025mmsi} extends spatial reasoning to \textit{multi-image} settings, requiring models to integrate spatial cues across multiple views. MMSI is particularly challenging, as each question is manually crafted by experts rather than automatically generated from templates.
%
% \noindent 
\textbf{MindCube}~\cite{mindcube} targets \textit{mental modeling} of scenes from limited observations, probing the ability to reconstruct occluded spaces and simulate viewpoints. Following the official setup, we train and evaluate on the non-overlapping MindCube-10K and MindCube-Tiny respectively.
% 
% \noindent 
\textbf{ViewSpatial-Bench}~\cite{li2025viewspatial} focuses on \textit{multi-perspective} localization, evaluating a model’s perspective-taking ability to reason across egocentric (camera-centric) and allocentric (object- or human-centric) viewpoints.
%
% \noindent 
\textbf{SITE}~\cite{wang2025site} offers \textit{broad cognitive coverage} by unifying over thirty datasets spanning diverse aspects of spatial intelligence. We adopt SITE to evaluate the generalization capability of \name, as it contains highly abstract and diverse test cases.
%
% \noindent \textbf{BLINK}~\cite{fu2024blink} covers Perspective-taking tasks such as visual correspondence and multi-view reasoning. 
%
% \noindent \textbf{3DSR}~\cite{ma20243dsrbench} evaluates spatial reasoning on natural images. Its paired-view setup further tests model robustness.
%
% \noindent \textbf{EmbSpatial}~\cite{du2024embspatial} targets on egocentric views, assessing object relation reasoning in embodied scenes.
%
% \noindent 
\textbf{BLINK}~\cite{fu2024blink} targets \textit{core visual perception}, covering tasks such as relative depth estimation, visual correspondence, and multi-view reasoning. We include BLINK to probe whether \name possesses the low-level perceptual foundations required for higher-level spatial reasoning.
%
% \noindent 
\textbf{3DSR}~\cite{ma20243dsrbench} targets \textit{3D spatial reasoning} on natural images, evaluating object location, orientation, height, and multi-object relations under diverse camera viewpoints. Its paired-view setup further tests the robustness of \name to common and uncommon perspectives.
%
% \noindent 
\textbf{EmbSpatial}~\cite{du2024embspatial} targets \textit{embodied spatial understanding} from an egocentric perspective, assessing whether models can reason about spatial relations grounded in embodied scenes. We adopt EmbSpatial to examine \name in action-oriented settings where spatial understanding must be aligned with first-person perception.

\subsection{Main Results}

% \noindent \textbf{Spatial Intelligence Benchmarks.}
We compare \name against leading open-source and proprietary multimodal models. As shown in~\cref{tab:main_results_spatial}, we observe three key findings:
(1) \name outperforms all general open-source models by clear margins, and even surpasses strong proprietary ones such as GPT-5~\cite{openai_gpt5_systemcard}, revealing persistent knowledge gaps in existing foundation models.
(2) \name also achieves superior performance over all dedicated spatial-intelligence models, suggesting that algorithmic innovation alone may be premature when the benefits of large-scale spatial data have not yet been fully realized. Notably, \name surpasses two recent strong baselines (VST~\cite{vst2025} and Cambrian-S~\cite{yang2025cambrian}) even when using comparable amounts of training data (\cref{fig:teaser}) and a smaller model (2B parameters). We attribute these gains to the inclusion of extensive perspective-taking data, which is central to spatial intelligence.
(3) While InternVL3~\cite{zhu2025internvl3}, Qwen3-VL~\cite{Qwen3-VL}, and Bagel~\cite{deng2025bagel} exhibit slightly different behaviors, \name consistently improves upon all three families. This further validates the effectiveness of our scaling strategy across diverse architecture designs and pretraining paradigms.

Moreover, we include model performance on general undestanding benchmarks 
(MMBench-EN~\cite{liu2024mmbench}, MMStar~\cite{chen2024rightwayevaluatinglarge}, AI2D~\cite{Hiippala_2020}, 
OCRB~\cite{Liu_2024},  
DocVQA~\cite{mathew2021docvqa}, 
MMVP~\cite{tong2024mmvp}, 
V*~\cite{wu2024v}, 
MMMU~\cite{yue2024mmmu} and 
Vid-MME~\cite{fu2025videomme}) 
in~\cref{tab:main_results_general}, and find that data diversity is crucial: incorporating a wide coverage of multimodal data and varied general knowledge sources effectively mitigates catastrophic forgetting and preserves overall multimodal competence.
%
% Additional results on general benchmarks, including MMStar~\cite{chen2024rightwayevaluatinglarge}, AI2D~\cite{Hiippala_2020}, and OCRB~\cite{Liu_2024}, are provided in the \Supp.

\subsection{Scaling}
\label{sec:exp:scaling}

\subsubsection{Effectiveness.}
As shown in~\cref{fig:teaser}, scaling spatial intelligence data leads to steady improvements across all key capability dimensions. 
% Per-benchmark scaling curves are provided in the \Supp. 
We highlight three observations:
(1) Data mixing is highly effective. By aggregating a wide collection of public datasets and further enlarging the spatial intelligence corpus, \name surpasses existing 7B spatial-intelligence baselines with models one size smaller (2B) under comparable data budgets.
(2) Model size impacts capability trends. While InternVL3 2B and 8B variants exhibit similar performance trajectories on MM, SR, and CR tasks, their behaviors diverge sharply on PT tasks. We hypothesize that the 2B model lacks sufficient capacity to robustly learn viewpoint transformations: a challenging but essential component of spatial intelligence.
(3) Capability-wise differences reveal data-driven gains. Proprietary models such as GPT-5~\cite{openai_gpt5_systemcard} are notably strong on SR, yet show clear deficiencies in PT. In contrast, \name-InternVL3-8B convincingly outperforms GPT-5 on PT, benefiting from the large-scale, comprehensive perspective-taking data introduced during continued scaling. Interestingly, even though we include very limited CR data during training, \name still gradually surpasses GPT-5 in CR performance. This suggests the presence of capability synergy, where advances in fundamental spatial tasks (\eg, PT and SR) transfer to more complex reasoning skills. We discuss this further in~\cref{sec:exp:capability_emergence}.

\subsubsection{Saturation.}
As shown in~\cref{fig:teaser}, the performance gains gradually diminish as the amount of training data increases. While it remains unclear whether continued scaling will eventually reach a tipping point that triggers stronger emergent capabilities (though we note some early signs discussed in~\cref{sec:exp:capability_emergence}), we concur with the broader community that data scaling alone is unlikely to achieve human-level spatial intelligence~\cite{yang2025cambrian}.
Motivated by this, we commit to fully open-sourcing the weights of \name, allowing the community to bypass the costly scaling stage and instead focus on advancing algorithmic innovation on top of a strong, spatially capable foundation.

\clearpage

\input{figures/spillover}

\subsection{Capability Emergence}
\label{sec:exp:capability_emergence}

We present several interesting cases observed during scaling that may suggest early signs of emerging spatial intelligence.

\subsubsection{Spill-Over.}

\input{tables/5_4_frames_ablation}
Large-scale mixed-domain training inevitably exposes models to a broad distribution of scenarios, making it increasingly difficult to determine whether downstream improvements stem from genuine, generalizable spatial reasoning or from incidental overlap with training data. To more rigorously examine spatial capability spill-over, we therefore conduct controlled experiments in which models are trained on a single dataset and evaluated on tasks drawn from entirely different domains.
As shown in Fig.~\ref{fig:spillover}, we observe clear emergence and transfer of spatial understanding across tasks.
The view-transformation dataset, constructed from Ego-Exo4D~\cite{grauman2024ego}, requires models to translate between egocentric and exocentric viewpoints—forcing them to infer cross-view geometric relationships. This ability transfers strongly to downstream tasks such as Maze Pathfinding~\cite{wang2025site} and Pos-Cam-Cam~\cite{yang2025mmsi}, both of which depend on sequential viewpoint simulation and aggregating information across views.
Similarly, the dataset built from MessyTable~\cite{cai2020messytable} images requires models to identify shared objects and infer spatial relationships between two viewpoints. This yields notable gains on benchmark sub-categories such as MMSI~\cite{yang2025mmsi} Pos-Cam-Cam and Attr-Appr, both of which rely on robust spatial correspondence identification between paired images.

\subsubsection{Extrapolation.}

A surprising observation is that although \name is trained with at most 16 frames per sample, it generalizes effectively to sequences of 32 frames or more at inference time, as shown in~\cref{tab:vsi_frames_ablation_compact_no_multirow}. This suggests that \name learns to construct coherent spatial structure rather than merely repeating patterns confined to the supervised training window.
Interestingly, while \name does not continue to extrapolate beyond 64 frames, unlike Cambrian-S~\cite{yang2025cambrian}, which is explicitly trained with much longer context windows of 64 or 128 frames, \name nevertheless achieves performance comparable to Cambrian-S while using substantially fewer frames at inference. This indicates that \name possesses a stronger spatial understanding capability that enables it to form meaningful connections across larger temporal gaps, without relying on densely sampled frame sequences.

\subsection{Overfit and Shortcut Analysis}

\input{tables/5_mindcube_wo_image}

Recent studies~\cite{vsidebiased} have shown that multimodal models can exploit language shortcuts to answer questions without genuine visual reasoning. To ensure that the improvements of \name are not due to overfitting to QA text, we conduct targeted analyses on VSI~\cite{vsi} and MindCube~\cite{mindcube}.

The recently proposed VSI-Debiased~\cite{vsidebiased} is a specifically designed variant of VSI to eliminate text-only shortcuts by removing questions that can be answered correctly without visual understanding. As shown in~\cref{tab:vsi_frames_ablation_compact_no_multirow}, when evaluated on VSI-Debiased, \name exhibits a substantially smaller performance drop compared to Cambrian-S-7B~\cite{yang2025cambrian}, indicating that \name relies less on textual heuristics and more on spatially grounded understanding.

For MindCube, we follow the protocol in~\cite{wang2024picture} and evaluate models \textit{without visual inputs}.
Surprisingly, as shown in \cref{tab:5_mindcube_circular}, the previous open-source SoTA on MindCube, \textit{MindCube-RawQA-SFT}~\cite{mindcube} achieves a score of 50.7 without any images, which is almost identical to its performance with full visual inputs, revealing a heavy dependence on language priors rather than visual reasoning.

In contrast, \name drops from 85.6 to 52.5 in the no-vision setting, validating that it genuinely uses visual information rather than relying on language shortcuts. Notably, both models converge to a score around 50 in the absence of images, underscoring the importance of debiasing benchmarks, as argued in~\cite{vsidebiased}.

To further verify that \name does not overfit to text option ordering, we conduct circular tests~\cite{liu2024mmbench,li2024core,cai2025has}, which reorders the choices in the questions to eliminate dependency on certain text patterns. As reported in~\cref{tab:5_mindcube_circular}, \name exhibits minimal degradation under the Soft circular test~\cite{li2024core}.
Even in the Hard circular test~\cite{liu2024mmbench}, which requires robust handling of all rotations of answer choices, \name drops 10 points, whereas MindCube-RawQA-SFT drops nearly 30 points. This demonstrates that \name is far less sensitive to superficial text patterns and possesses more stable, input-grounded reasoning.

\subsection{Spatial Chain-of-Thought}
\label{sec:exp:spatial_cot}

\input{tables/5_6_cot}
Chain-of-Thought (CoT)~\cite{wei2022chain} has become the \textit{de facto} paradigm for complex reasoning tasks. However, despite numerous recent attempts~\cite{mindcube,yang2025mmsi,wang2025spatialviz,jia2025omnispatial}, incorporating CoT variants yields only marginal improvements (typically around $\sim$2\%), which are consistently overshadowed by gains from large-scale, carefully curated spatial datasets.

In~\cref{tab:5_6_cot}, we present a preliminary evaluation of different CoT styles. We examine three paradigms:
(1) CoT-GPT-5, which directly uses a large language model (GPT-5~\cite{openai_gpt5_systemcard}) to annotate CoT given the question and ground-truth answer;
(2) CoT-MindCube-Aug-CGMap, which follows MindCube~\cite{mindcube} and constructs a JSON-style cognition map (CogMap) within the CoT;
(3) CoT-\name-CGMap, our extended CogMap that provides step-by-step tracking of objects across frames, maps them to a world coordinate system with precise (rather than coarse-grid) coordinates, and reasons about relative spatial relationships more explicitly.
(4) CoT-\name-CGMap, followed by followed by reinforcement learning based on GRPO~\cite{shao2024deepseekmath}.

We train each variant on roughly 100K examples, reasonably large compared to typical CoT studies, and evaluate on VSI’s Object Relative Direction task, a challenging subset known to impede strong baselines such as InternVL3. We find that (1) our elaborated CoT achieves the highest improvement among the three, but (2) all CoT variants yield limited absolute gains, insufficient to justify their computational overhead, especially given the extra tokens required during both training and inference. (3) RL does not yield clear performance gains over a strong baseline. We hypothesize RL in LLM may not be readily helpful for spatial reasoning as long text may not be ideal for spatial reasoning: as discussed in~\cref{sec:case_study_text_cot}, we discover that long spatial CoT is prone to inconsistency and internal mistakes, that undermines the performance. 

Our findings suggest that while carefully engineered CoT can offer modest benefits, text-based reasoning alone may be neither the most efficient nor the most effective paradigm for spatial intelligence. Hence, multimodal RL for spatial reasoning remains largely underexplored, consistent with observations in SpatialReasoner~\cite{ma2025spatialreasoner}. This may signal the need for a broader paradigm shift beyond conventional CoT.

\subsection{Downstream Task}

\input{tables/5_7_embodied_bench}

To evaluate the practical utility of \name's enhanced spatial intelligence, we conduct downstream robot manipulation experiments on EmbodiedBench~\cite{yang2025embodiedbench}, focusing specifically on its spatial subset. In this setting, \name is instantiated as an embodied agent that controls a virtual Franka Panda robot to execute user instructions containing rich spatial language such as "left", "on top of", "rear", and "horizontal".
Importantly, \textit{no finetuning} of \name is performed for this evaluation. Quantitative results for the spatial subset are shown in~\cref{tab:embodied_bench}.

\input{figures/downstream_embodied_bench}

We report success rates under two prompting settings: the official prompt (OP) and a spatial-intelligence-oriented prompt (SIP). OP supplies bounding-box coordinates extracted from the input image, whereas SIP enriches OP with additional object-grounding cues to reduce ambiguity from object recognition and better isolate spatial-reasoning performance.

Across both OP and SIP, \name delivers substantial improvements, demonstrating that enhanced spatial intelligence directly benefits embodied manipulation: \name more reliably identifies key spatial cues, enabling more accurate reasoning and more consistent action planning.

Representative rollouts are visualized in~\cref{fig:downstream_embodied_bench}. These examples demonstrate that \name effectively integrates spatial information from both language instructions and visual observations, plans coherent motion trajectories, and generates action sequences that enable the robot to successfully complete the tasks.

% Additional implementation details and experimental configurations are provided in the \Supp.

%% file: tables/main_results_spatial.tex
\begin{table*}[t]
\centering
\small

\setlength{\tabcolsep}{1.2pt}
\resizebox{1.0\linewidth}{!}{

\begin{tabular}{lcccccccccc} % 11 columns (1 model name + avg + 8 benchmarks)
\toprule
\textbf{Models} &
\textbf{Avg.} &
\textbf{VSI-Bench~\cite{vsi}} &
\textbf{MMSI-Bench~\cite{yang2025mmsi}} &
\textbf{MindCube*~\cite{mindcube}} &
\textbf{ViewSpatial~\cite{li2025viewspatial}} &
\textbf{SITE~\cite{wang2025site}} &
\textbf{BLINK~\cite{fu2024blink}} &
\textbf{3DSR~\cite{ma20243dsrbench}} &
\textbf{EmbSpatial~\cite{du2024embspatial}}

\\

\midrule
\rowcolor{line-blue}\textbf{Metric} & & MRA, Acc & Acc & Acc & Acc & CAA & Acc & Acc & Acc\\

% \midrule
% \rowcolor{line-blue}\textbf{Human Evaluation} & & & & & & \\
% \footnotesize $\Delta$(Best Model,Human) &
% \colornum{-10.42} &
% \colornum{-5.62} &
% \colornum{-53.90} &
% \colornum{-8.88} &
% - & \colornum{-28.97} \\

\midrule
\textbf{Human} & - & \textbf{79.2} & \textbf{97.2}& \textbf{94.5} & - & \textbf{67.5} & \textbf{95.67} & \textbf{95.7} & \textbf{90.33} \\

% \midrule
\textbf{Random Choice} & - & 34.0 & 25.0 & 33.0 & 26.3 & 0.0 & 38.09 & 45.8 & 25.0 \\

\midrule
\rowcolor{line-blue}\textbf{Proprietary Models} & & & & & & & & & \\

Seed-1.6-2025-06-15~\cite{seed2025seed1_5vl}
& 54.2
& 49.9
& 38.3
& 48.8
& 43.9
& 54.6
& 65.9
& 56.9
& 75.4
\\

Gemini-2.5-Pro-2025-06~\cite{team2023gemini}
& 58.0
& \cellcolor{secondcolor}{53.6}
& 38.0
& 57.6
& \cellcolor{secondcolor}{46.1}
& 57.1
& \cellcolor{secondcolor}{73.5}
& 59.3
& 78.8
\\

Grok-4-2025-07-09~\cite{grok4_xai_2025}
& 53.3
& 47.9
& 37.8
& \cellcolor{secondcolor}{63.6}
& 43.2
& 47.0
& 56.4
& 54.9
& 75.5
\\

GPT-5-2025-08-07~\cite{openai_gpt5_systemcard}
& \cellcolor{secondcolor}{58.8}
& \cellcolor{bestcolor}{\textbf{55.0}}
& \cellcolor{secondcolor}{41.8}
& 56.3
& 45.6
& \cellcolor{secondcolor}{61.9}
& 68.0
& \cellcolor{secondcolor}{60.3}
& \cellcolor{secondcolor}{81.6}
\\

Gemini-3-Pro-Preview~\cite{gemini_3_pro_systemcard}
& \cellcolor{bestcolor}{\textbf{63.8}}
& 52.5
& \cellcolor{bestcolor}{\textbf{45.2}}
& \cellcolor{bestcolor}{\textbf{70.9}}
& \cellcolor{bestcolor}{\textbf{50.4}}
& \cellcolor{bestcolor}{\textbf{62.3}}
& \cellcolor{bestcolor}{\textbf{76.0}}
& \cellcolor{bestcolor}{\textbf{68.9}}
& \cellcolor{bestcolor}{\textbf{84.3}}
\\

\midrule
\rowcolor{line-blue}\textbf{Open-source General Models} & & & & & & & & & \\

Bagel-7B-MoT~\cite{deng2025bagel} &
45.3 &
31.4 &
31.0 &
34.7 &
41.3 &
37.0 &
63.6 &
50.2 &
73.1
\\

%% Qwen2.5-VL

Qwen2.5-VL-3B-Instruct~\cite{Qwen2.5-VL} & 39.1 & 27.0 & 28.6 & 37.6 & 32.0 & 33.1 & 48.7 & 43.5 & 62.3 \\

Qwen2.5-VL-7B-Instruct~\cite{Qwen2.5-VL} & 43.1 & 32.3 & 26.8 & 36.0 & 36.9 & 37.6 & 55.9 & 47.5 & 71.8 \\

% Qwen2.5-VL-72B-Instruct~\cite{Qwen2.5-VL} & 35.77 & 47.41 & 32.50 & 42.21 & 60.97 & 86.00 & 57.27 & 40.82 \\

%% Qwen3-VL

Qwen3-VL-2B-Instruct~\cite{Qwen3-VL} & 44.6 & 50.4 & 28.9 & 34.5 & 37.0 & 35.7 & 53.2 & 47.5 & 70.1 \\

Qwen3-VL-8B-Instruct~\cite{Qwen3-VL} & 50.6 & 57.9 &
31.1 &
29.4 & 42.2 &
45.8 &
\cellcolor{bestcolor}{\textbf{66.7}} &
\cellcolor{secondcolor}{53.9} &
\cellcolor{bestcolor}{\textbf{77.7}} \\

%% InternVL3 & 3.5

InternVL3-2B~\cite{zhu2025internvl3} & 39.8 & 33.0 & 26.5 & 37.5 & 32.6 & 30.0 & 50.8 & 47.7 & 60.1\\

InternVL3-8B~\cite{zhu2025internvl3} & 45.7 & 42.1 & 28.0 & 41.5 & 38.7 & 41.1 & 53.5 & 44.3 & \cellcolor{secondcolor}{76.3} \\

% InternVL3-78B~\cite{zhu2025internvl3} & 47.55 & 52.72 & 30.50 & 49.52 & xx.xx & xx.xx & 60.80 & 42 \\

% InternVL3.5-8B~\cite{wang2025internvl3} &  &  &  &  &  &  &  &  \\

\rowcolor{line-blue}\textbf{Open-source Spatial Intelligence Models} & & & & & & & & & \\

MindCube-3B-RawQA-SFT~\cite{mindcube} & 22.0 & 17.2 & 1.7 &
51.7 &
24.1 &
6.3 &
35.1 &
2.8 &
37.0 \\

% MindCube-3B-Aug-CGMap-FFR-Out-SFT~\cite{yin2025spatial(mindcube)} & 29.6 & 29.1 & 41.0 & 30.9 & 28.2 \\

% MindCube-3B-Plain-CGMap-FFR-Out-SFT~\cite{yin2025spatial(mindcube)} & 29.9 & 30.4 & 39.9 & 31.2 & 28.7 \\

SpatialLadder-3B~\cite{li2025spatialladder} & 40.9 & 44.9 & 27.4 &
43.5 & 39.9 & 28.0 & 43.0 & 42.8 & 58.2 \\

Spatial-MLLM-4B~\cite{wu2025spatial} & 35.6 & 46.3 & 26.1 & 33.5 & 34.7 & 18.0 & 40.5 & 36.2 & 50.0\\

SpaceR-7B~\cite{ouyang2025spacer} & 41.8 & 41.6 & 27.4 & 38.0 & 35.9 & 34.3 & 49.6 & 40.5 & 66.9 \\

ViLaSR-7B~\cite{vilasr} & 43.7 & 44.6 & 30.2 & 35.1 & 35.7 & 38.7 & 51.4 & 46.6 & 67.3 \\

VST-3B-SFT~\cite{vst2025} & 47.7 & 51.4 & 28.8 & 36.0 & \cellcolor{secondcolor}{52.9} & 35.9 & 58.8 & 48.7 & 69.0\\

% VST-3B-RL$^{\dagger}$~\cite{vst2025} & 57.7 & - & 31.3 & - & - & 57.2\\

VST-7B-SFT~\cite{vst2025} &
50.8 &
55.5 &
32.5 &
39.7 &
50.5 &
39.7 &
61.9 &
53.1 &
73.7 \\

% VST-7B-RL$^{\dagger}$~\cite{vst2025} & 61.2 & - & 34.8 & - & - & 62.6\\

Cambrian-S-3B~\cite{yang2025cambrian} &
42.0 &
56.1 & 27.0 & 38.4 & 41.0 & 31.0 & 37.7 & 41.4 & 63.5\\

Cambrian-S-7B~\cite{yang2025cambrian} &
45.1 &
62.9 & 27.1 & 37.9 & 41.3 & 36.1 & 37.9 & 45.0 & 72.8\\

\rowcolor{line-blue}\textbf{Ours} & & & & & & & & & \\

\textbf{\name\textsubscript{Bagel-7B-MoT}}  &
48.6\df{(+3.3)} &
41.5\df{(+10.1)} &
34.5\df{(+3.5)} &
46.8\df{(+12.1)} &
46.9\df{(+5.6)} &
42.0\df{(+5.0)} &
\cellcolor{secondcolor}{65.4\df{(+1.8)}} &
42.4\ddf{(-7.8)} &
69.0\ddf{(-4.1)} \\
% \textbf{\name\textsubscript{Bagel-7B-MoT}}  & 
% 41.6\df{(+10.2)} & 
% 36.2\df{(+5.2)} & 
% 50.8\df{(+16.1)} & 
% 50.3\df{(+9.0)} & 
% 41.6\df{(+4.6)} \\

% \footnotesize $\Delta$ (\name\textsubscript{Bagel-7B-MoT}, Bagel-7B-MoT) &
% \colornum{10.2} &
% \colornum{5.2} &
% \colornum{16.1} & 
% \colornum{9.0} & 
% \colornum{4.6} \\

% \textbf{SenseSI\textsubscript{Qwen3-VL-2B}} & 50.74& 36.90 & 32.30& 35.87 & 37.89 & 52.29 \\

\textbf{\name\textsubscript{Qwen3-VL-8B}} &
\cellcolor{secondcolor}{58.1\df{(+7.5)}} &
\cellcolor{secondcolor}{64.8\df{(+6.9)}} &
\cellcolor{secondcolor}{38.1\df{(+7.0)}} &
\cellcolor{secondcolor}{73.8\df{(+44.4)}} &
51.2\df{(+9.0)} &
\cellcolor{bestcolor}{\textbf{49.6}\df{(+3.8)}} &
61.9\ddf{(-4.8)} &
53.2\ddf{(-0.7)} &
72.5\ddf{(-5.2)} \\
% \textbf{\name\textsubscript{Qwen3-VL-8B}} &
% 62.9\df{(+5.0)} &
% 37.5\df{(+6.4)} &
% 74.8\df{(+45.4)} &
% 48.4\df{(+6.2)} &
% \cellcolor{bestcolor}{\textbf{51.5}\df{(+5.7)}} \\

% \footnotesize $\Delta$ (\name\textsubscript{Qwen3-VL-8B}, Qwen3-VL-8B) &
% \colornum{5.0} &
% \colornum{6.4} &
% \colornum{45.4} & 
% \colornum{6.2} & 
% \colornum{5.7} \\

% 1.0
% \textbf{SenseSI-1.0\textsubscript{InternVL3-2B}} & \textbf{58.47}& \textbf{34.59} & \textbf{35.50} & \textbf{71.35} & \textbf{40.62} & \textbf{46.24} \\

% 1.1
\textbf{\name\textsubscript{InternVL3-2B}} &
49.4\df{(+9.6)} &
63.7\df{(+30.7)} &
34.2\df{(+7.7)} &
41.8\df{(+4.3)} &
52.7\df{(+20.1)} &
36.8\df{(+6.8)} &
52.4\df{(+1.6)} &
50.5\df{(+2.8)} &
62.8\df{(+2.7)} \\
% \textbf{\name\textsubscript{InternVL3-2B}} &
% 63.7\df{(+30.8)} &
% 34.2\df{(+7.7)} &
% 41.8\df{(+4.3)} &
% 52.6\df{(+20.1)} &
% 36.7\df{(+6.7)} \\

% \footnotesize $\Delta$ (\name\textsubscript{InternVL3-2B}, InternVL3-2B) &
% \colornum{30.8} &
% \colornum{7.7} &
% \colornum{4.3} & 
% \colornum{20.1} & 
% \colornum{6.7} \\

% \footnotesize $\Delta$ Base\textsubscript{InternVL3-2B} &
% \colornum{25.49} &
% \colornum{4.56} &
% \colornum{9.00} &
% \colornum{33.85} &
% \colornum{8.06} &
% \colornum{2.32}
% \\

% 1.0 - comment
% \textbf{SenseSI-1.0\textsubscript{InternVL3-8B}} & \textbf{62.80}& \textbf{44.51} & \textbf{37.90} & \textbf{89.33} & \textbf{53.92}  &  \textbf{54.71} \\

% 1.1
\textbf{\name\textsubscript{InternVL3-8B}} &
\cellcolor{bestcolor}{\textbf{61.5}\df{(+15.8)}} &
\cellcolor{bestcolor}{\textbf{68.8}\df{(+26.7)}} &
\cellcolor{bestcolor}{\textbf{43.3}\df{(+15.3)}} &
\cellcolor{bestcolor}{\textbf{85.7}\df{(+44.2)}} &
\cellcolor{bestcolor}{\textbf{54.7}\df{(+16.0)}} &
\cellcolor{secondcolor}{47.7\df{(+6.6)}} &
63.9\df{(+10.4)} &
\cellcolor{bestcolor}{\textbf{55.5}}\df{(+11.2)} &
72.0\ddf{(-4.3)} \\
% \textbf{\name\textsubscript{InternVL3-8B}} &
% \cellcolor{bestcolor}{\textbf{68.7}\df{(+26.6)}} &
% \cellcolor{bestcolor}{\textbf{43.3}\df{(+15.3)}} &
% \cellcolor{bestcolor}{\textbf{85.6}\df{(+44.1)}} &
% \cellcolor{bestcolor}{\textbf{54.6}\df{(+16.0)}} &
% \cellcolor{secondcolor}{47.7\df{(+6.6)}} \\

% \footnotesize $\Delta$ (\name\textsubscript{InternVL3-8B}, InternVL3-8B) &
% \colornum{26.6} &
% \colornum{15.3} &
% \colornum{44.1} & 
% \colornum{16.0} & 
% \colornum{6.6} \\

% \footnotesize $\Delta$ Base\textsubscript{InternVL3-8B} &
% \colornum{20.66} &
% \colornum{3.36} &
% \colornum{9.9} &
% \colornum{47.79} &
% \colornum{15.26} & 
% \colornum{1.05} 
% \\

% \footnotesize $\Delta$ (Best \name, Best Open-source) &
% \colornum{1.28} &
% \colornum{11.30} &
% \colornum{33.94} &
% \colornum{1.78} & 
% \colornum{6.47}  \\

% \footnotesize $\Delta$ (Best \name, Best Proprietary) &
% \colornum{13.75} &
% \colornum{1.50} &
% \colornum{22.11} &
% \colornum{8.61} & 
% \colornum{-9.58}  \\

\bottomrule
\end{tabular}
}
\vspace{-2pt}
\caption{\textbf{Evaluation on key spatial intelligence and general benchmarks.}
All results are evaluated on EASI~\cite{cai2025has}, using the official EASI-8 protocol.
MindCube$^*$ denotes MindCube-Tiny.
\textbf{\raisebox{0pt}[0pt][0pt]{\colorbox{bestcolor}{Dark purple}}} highlights the best result and \raisebox{0pt}[0pt][0pt]{\colorbox{secondcolor}{light purple}} indicates the second-best result within Proprietary and Open-source models, respectively.
% \TODO{explain why some numbers are quoted, some are by EASI. }
}
\label{tab:main_results_spatial}
\end{table*}

%% file: figures/spillover.tex
\begin{figure*}[t]
    \centering
    \setlength{\belowcaptionskip}{-2pt}
    \includegraphics[width=\linewidth]{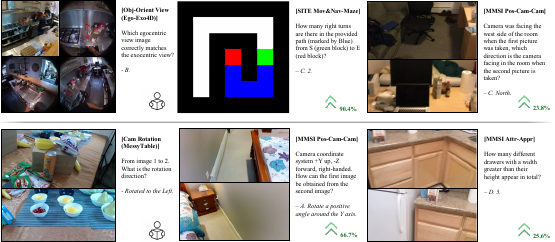}
    \caption{
    % Spill-Over of Spatial Capabilities from single task training and single data source.
    Observations on \textbf{generalization ability} from a single data source and single task.
    The upper example demonstrates how training on ego-exo association task enhance performance on task required imagined first-person perspectives.
    The lower example demonstrates how a camera rotation task, based on cross-view visual correspondence, generalizes to tasks with distinct questions and visual appearances.
    These findings suggest the potential existence of \textit{meta-tasks} in PT, which may enable related spatial capabilities.
    }
    \label{fig:spillover}
\end{figure*}

%% file: tables/5_4_frames_ablation.tex
\begin{wraptable}{r}{0.4\textwidth}
% \begin{table}[t]
\vspace{-4mm}
\centering
\small
\setlength{\tabcolsep}{1.5mm}
\resizebox{1\linewidth}{!}{
\begin{tabular}{llcccc}
\toprule
\multirow{2}{*}{\textbf{Model}} & \multirow{2}{*}{\textbf{Benchmark}} & \multicolumn{4}{c}{\textbf{\# Frames}} \\
\cmidrule(lr){3-6}
 &  & \textbf{16} & \textbf{32} & \textbf{64} & \textbf{128} \\

\midrule
\multirow{2}{*}{Cambrian-S-7B~\cite{yang2025cambrian}}
& VSI~\cite{vsi} & 58.6 & 63.6 & 66.4 & \textbf{67.5} \\
& VSI-Debiased~\cite{vsidebiased} & 49.7 & 55.6 & 59.1 & \textbf{59.9} \\
 
% \midrule
% \multirow{2}{*}{InternVL3-8B~\cite{zhu2025internvl3}}
% & VSI & 41.6 & 42.1 & \textbf{42.8} & 40.6 \\
% & VSI-Debiased & 37.1 & 38.5 & 38.0 & \textbf{40.6} \\
 
\midrule
\multirow{2}{*}{\name\textsubscript{InternVL3-8B}}
& VSI~\cite{vsi} & 64.6 & 68.7 & \textbf{68.8} & 66.3 \\
& VSI-Debiased~\cite{vsidebiased} & 58.9 & \textbf{62.8} & 62.4 & 59.7 \\
 
\bottomrule
\end{tabular}}
\vspace{-2pt}
\caption{Ablation on inference frames.
Our model is trained on maximum 16 frames per sample, while Cambrian-S-7B~\cite{yang2025cambrian} is trained on 64/128 frames.
\name demonstrates strong extrapolation capabilities beyond the training number of frames.
Interestingly, \name shows a clear lead over Cambrian-S-7B~\cite{yang2025cambrian} on two benchmarks, even with fewer frames at inference.
}
%for VSI and VSI-Debiased~\cite{brown2025benchmark(vsidebiased)}.}
\label{tab:vsi_frames_ablation_compact_no_multirow}
\vspace{-4pt}
% \end{table}
\end{wraptable}

%% file: tables/5_mindcube_wo_image.tex
\begin{wraptable}{r}{0.5\textwidth}
% \begin{table}[t]
\centering
\small
\setlength{\tabcolsep}{1.0mm}
\vspace{-4mm}
\resizebox{1\linewidth}{!}{
\begin{tabular}{lcccc}
\toprule
\textbf{Models} &
\textbf{Standard} &
\textbf{Soft cir.} &
\textbf{Hard cir.} &
\textbf{w/o. Vis.} \\

\midrule

Gemini-3-Pro-Preview~\cite{gemini_3_pro_systemcard} & 70.9 & 75.4 & 59.6 & 39.7 \\

MindCube-SFT-RawQA~\cite{yang2025cambrian} & 51.7 & 45.8 & 23.1 & 50.7 \\

\name\textsubscript{InternVL3-8B} & 85.6 & 84.0 & 75.6 & 52.5 \\

\bottomrule
\end{tabular}}
\vspace{-2pt}
\caption{Analysis on MindCube~\cite{mindcube}.
\textit{Soft cir.} and \textit{Hard cir.} stands for Soft circular and Hard circular following~\cite{cai2025has}.
\textit{w/o. Vis.} indicates testing without visual as input, following~\cite{wang2024picture}.
}
\label{tab:5_mindcube_circular}
\vspace{-4pt}
% \end{table}
\end{wraptable}

%% file: tables/5_6_cot.tex
% \begin{wraptable}{r}{0.45\textwidth}
% % \begin{table}[h]
% \vspace{-1mm}
% \centering
% \small
% \setlength{\tabcolsep}{1.0mm}
% \resizebox{1\linewidth}{!}{
% \begin{tabular}{lccccc}
% \toprule
% \multirow{2}{*}{\textbf{CoT Style}} & 
% \multirow{2}{*}{\textbf{\makecell{Average \# \\ Output Token}}} &
% \multicolumn{4}{c}{\textbf{VSI-Bench: Obj. Rel. Direction}} \\
% \cmidrule(lr){3-6}
% & & 
% \multicolumn{1}{c}{\textbf{Overall}} & 
% \textbf{Easy} & 
% \textbf{Medium} & 
% \textbf{Hard} \\

% \midrule
% \rowcolor{line-blue}\textbf{InternVL3-8B} & 1        & 39.3 & 48.8 & 47.0 & 21.9 \\
% No CoT                                    & 3.4       & 54.9   & 62.2   & 55.8   & 46.6 \\
% CoT-GPT-5                                 & 1070.7   & 40.0 & 41.4 & 43.1 & 36.1 \\
% CoT-MindCube-Aug-CGMap                    & 1490.6       & 39.9 & 45.9 & 42.7 & 33.7 \\
% CoT-\name-CGMap                           & 2262.8       & 47.9 & 53.9 & 51.3 & 41.0 \\
% \bottomrule
% \end{tabular}}
% \vspace{-2pt}
% \caption{Impact of Chain-of-Thought (CoT) formats on the Object Relative Direction subset of VSI-Bench.}
% \label{tab:5_6_cot}
% \vspace{-4pt}
% % \end{table}
% \end{wraptable}

\begin{wraptable}{r}{0.4\textwidth}
\vspace{-4mm}
\centering
\small
\setlength{\tabcolsep}{1.0mm}
\resizebox{1\linewidth}{!}{
\begin{tabular}{lcc}
\toprule
Model & \#Token & VSI-Bench \\
\midrule
InternVL3-8B            & 1.0    & 42.1 \\ 
\midrule
\rowcolor{line-blue} \multicolumn{3}{l}{\textbf{Train set: Rel. Dir. Subset}} \\
No CoT                  & 3.4    & 40.6 \\
CoT-GPT-5               & 1175.5 & 26.5 \\
CoT-MindCube-Aug-CGMap  & 3940.7 & 17.0 \\
CoT-\name-CGMap         & 2534.5 & 31.8 \\
\midrule
\rowcolor{line-blue} \multicolumn{3}{l}{\textbf{Train set: Full set (QA + CoT)}} \\
CoT-\name-CGMap         & 190.8  & 49.2 \\
+ RL (GRPO)             & 1299.2 & 43.1 \\ 
\bottomrule
\end{tabular}}
\vspace{-2pt}
\caption{Impact of Chain-of-Thought (CoT) formats on the Object Relative Direction subset or the full set of VSI-Bench.}
\label{tab:5_6_cot}
\vspace{-4pt}
\end{wraptable}

%% file: tables/5_7_embodied_bench.tex
\begin{wraptable}{r}{0.45\textwidth}
% \begin{table}[t]
\centering
\small
\setlength{\tabcolsep}{1.0mm}
\captionsetup{skip=2pt}
\vspace{-4mm}
\begin{tabular}{cccccc}
\toprule
\multicolumn{2}{c}{\textbf{GPT-4o}} &
\multicolumn{2}{c}{\textbf{InternVL3-8B}} &
\multicolumn{2}{c}{\textbf{\name\textsubscript{InternVL3-8B}}} \\
\cmidrule(lr){1-2} \cmidrule(lr){3-4} \cmidrule(lr){5-6}
OP & SIP & OP & SIP & OP & SIP \\
\midrule
37.5 & 45.8 & 10.4 & 20.8 & \textbf{16.6} (+59.6\%) & \textbf{33.3} (+60.0\%) \\
\bottomrule
\end{tabular}
\caption{
Success rate on \textbf{Spatial} subset of EmbodiedBench~\cite{yang2025embodiedbench}.
\textbf{OP}: Official Prompt; \textbf{SIP}: Spatial-Intelligence-oriented Prompt.
}
\label{tab:embodied_bench}
% \end{table}
\end{wraptable}

%% file: figures/downstream_embodied_bench.tex
\begin{wrapfigure}{r}{0.45\textwidth}
% \begin{figure}[t]
    \centering
    \setlength{\abovecaptionskip}{-2pt} 
    \setlength{\belowcaptionskip}{-2pt}
    \vspace{-4mm}
    \includegraphics[width=\linewidth]{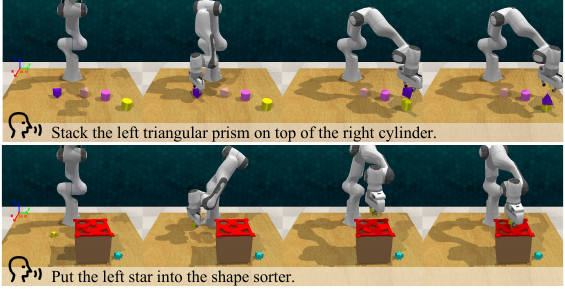}
    \caption{
        Visualization of the manipulation task rollout in EmbodiedBench~\cite{yang2025embodiedbench}, performed by the embodied agent powered by \name.
    }
    \label{fig:downstream_embodied_bench}
    % \vspace{mm}
% \end{figure}
\end{wrapfigure}

%% file: sec/6_conclusion.tex
\section{Conclusion}
\label{sec:conclusion}
In this work, we validate the effectiveness of scaling spatial intelligence across multiple multimodal foundation models, achieving consistent performance gains across a wide range of benchmarks. We further show that the enhanced models retain their general capabilities while exhibiting improved generalization, including abilities that do not emerge without large-scale, diverse training data. We hope this study provides a strong foundation for future research on developing spatial intelligence in multimodal foundation models.

%% file: sec/x_suppl.tex
\section{Details of Fig.\ref{fig:teaser}}

\subsection{Four Subfigures at the Corners}
The four subfigures at the corners elaborate \textbf{\name}'s performance on four core spatial capabilities (\ie, Perspective-taking, Spatial Relations, Metric Measurement, and Comprehensive Reasoning). Through data scaling, \name surpassing open-source models and even outperforms GPT-5 in specific spatial abilities, such as Perspective-taking. The lines denote the average performance across benchmark subtasks within each capability, while the shaded regions (confidence bands) represent $\pm0.5$ standard deviation. The detailed benchmark sub-tasks associated with each spatial capability are listed below.

\begin{itemize}[leftmargin=*]
\item\textbf{Perspective-taking}. VSI-Bench: Obj. Rel. Direction; MMSI-Bench: Positional Relationship subtasks (Cam-Cam, Obj-Obj, Reg-Reg, Cam-Obj, Obj-Reg, Cam-Reg), Motion subtasks (Motion-Cam, Motion-Obj); SITE: multi-view \& cross-image reasoning.
\item\textbf{Spatial Relations}. VSI-Bench: Obj. Rel. Distance; SITE: 3d information understanding, spatial relationship reasoning.
\item\textbf{Metric Measurement}. VSI-Bench: Obj. Size, Room Size, Obj. Abs. Distance; MMSI-Bench: Attribute Meas..
\item\textbf{Comprehensive Reasoning}. VSI-Bench: Obj. Cnt., Obj. Appear Order, Route Plan; MMSI-Bench: MSR.    
\end{itemize}

\subsection{Normalization for Radar Chart Visualization}
For the radar charts in \cref{fig:teaser}, we apply normalization to enable a fair and intuitive comparison of relative performance. Specifically, for each metric, we first scale all values by dividing them by the maximum value observed across models. We normalize all metrics so that the best score among the models is mapped to 1.0 and the worst score is mapped to 0.2. The radar chart axes have a range of 0.0 to 1.0.

\section{Additional Details in Data Curation}
Our unified data pipeline collects data from diverse sources and efficiently converts them into reliable, high-quality QA and Chain-of-Thought (CoT) labels.

\subsection{Data Processing}

\subsubsection{Unified Annotation.} We standardize heterogeneous raw data from source datasets into a unified set of spatial and multi-view annotations.
Specifically, we convert existing formats and augment the data with additional labels to obtain: 3D camera poses, 3D object poses including bounding boxes and orientations, 2D point and object visibility, and rich semantic labels of object and human–object interaction descriptions.

\subsubsection{Dataset-specific Processing.}
\begin{itemize}[leftmargin=*]
    \item \textbf{ScanNet~\cite{dai2017scannet}, ScanNet++~\cite{yeshwanth2023scannet++}.}
    These datasets provide 3D camera poses, 3D object bounding boxes, and 3D point clouds with object IDs. For each camera view, we project the 3D point cloud onto the image plane to establish correspondences between 2D pixels and 3D points, and to derive per-object projected and visible 2D bounding boxes.

    \item \textbf{SUN RGB-D~\cite{song2015sun}, CA-1M~\cite{lazarow2025cubify}.}  
    These datasets provide 3D camera poses, 3D object poses, and 2D object bounding boxes. Building on this, we refine and standardize the 3D object orientations, discard object categories whose orientations are inconsistent across scenes, and, using the accurate orientation labels, further annotate possible human–object interactions with hypothetical 3D poses and rich textual descriptions.
    
    \item \textbf{MessyTable~\cite{cai2020messytable}.}  
    This dataset provides 3D camera poses, 2D object bounding boxes, and cross-view instance association labels (the same object instance is assigned the same instance ID in different viewpoints). We further employ a vision–language model (VLM) to annotate fine-grained textual descriptions of object appearance details.
\end{itemize}

\subsection{Object Selection}
We adopt a unified object selection pipeline to retain only recognizable objects with sufficient informative details captured within each frame. The resulting per-frame object sets provide a clean and reliable basis for cross-frame association analysis and QA construction.

\subsubsection{Semantic Filtering.}
We first filter out object categories with weak geometric structure and ambiguous 3D position, such as \textit{floor}, \textit{ceiling}, and \textit{wall}, as well as objects with unclear or undefined semantic labels. 

\subsubsection{Visibility-based Filtering.}
\begin{itemize}[leftmargin=*]
    \item \textbf{Minimum Visible Size.}  
     We keep only the objects whose visible 2D bounding box (\ie, the portion not occluded by other objects and lying within the camera view) occupies at least a certain fraction of the image area.
    
    \item \textbf{Visibility Ratio Threshold.}  
    We further discard objects whose visible 2D bounding box area falls below a fixed ratio of their total projected 2D bounding box area.
\end{itemize}

\input{figures/case_messytable}

\subsection{Image Selection}
To derive multi-view image sets that are well-posed, visually associated, and sufficiently challenging, we adopt a three-stage image selection pipeline.

\subsubsection{Basic Pose Filtering.}
We first discard views with extreme camera poses. In particular, we remove images whose camera pitch (severely top-down or bottom-up) or yaw deviates excessively from the typical viewing direction. This step eliminates degenerate or highly uninformative viewpoints.

\subsubsection{Cross-view Association Control.}
\begin{itemize}[leftmargin=*]
    \item \textbf{Connectivity.}  
    We select images into sets in a manner such that, for any pair of images in the same set, there exists at least one connecting path, along which every pair of adjacent images satisfies a minimum association score. The calculation of the association score depends on the available annotation. For datasets with 2D point visibility, we compute the score based on the number of overlapping visible points. Otherwise, we compute the number of shared visually valid objects.

    \item \textbf{Difficulty.}  
    To avoid trivial cross-view associations, we enforce that the association score between any pair of images does not exceed a specified maximum. We further exploit dataset-specific properties to design richer forms of difficulty control. For example, as shown in~\cref{fig:case_messytable}, we emphasize hard cases in MessyTable where multiple visually similar or identical objects exist in the same scene. In such cases, establishing cross-view associations requires fully understanding of the 3D spatial layout, while appearance-based shortcuts are impossible.
\end{itemize}

\subsubsection{Full-scene Coverage Selection.}
For scan-based datasets with point-level annotations, we further extend our selection to image sets with broader coverage of the scenes. Leveraging the temporal continuity of the scan videos, we design a time-efficient greedy algorithm that iteratively adds views to maximize point coverage, while maintaining the cross-view connectivity and difficulty constraints above. The resulting procedure is summarized in Algorithm~\ref{alg:frame-selection}.

\begin{algorithm}[h]
\caption{Frame Selection with Overlap Control}
\label{alg:frame-selection}
\begin{algorithmic}[1]
\REQUIRE Video frames $\{f_1, \dots, f_T\}$, visible points $V_{f_k}$, minimum frames $N_{\text{min}}=16$
\ENSURE Selected frame set $S$
\STATE $C \gets \emptyset$ \COMMENT{Initialize covered point set}
\STATE $S \gets \{f_1\}$ \COMMENT{Start with first frame}
\STATE $C \gets V_{f_1}$

\FOR{$k = 2$ \TO $T$}
    \STATE $\rho_k \gets |V_{f_k} \cap C| / |V_{f_k}|$ \COMMENT{Compute overlap with coverage}
    \IF{$0.03 \le \rho_k \le 0.20$}
        \STATE $S \gets S \cup \{f_k\}$
        \STATE $C \gets C \cup V_{f_k}$ \COMMENT{Update covered points}
    \ENDIF
\ENDFOR

\IF{$|S| < N_{\text{min}}$}
    \STATE Insert additional frames uniformly in temporal gaps until $|S| = N_{\text{min}}$
\ENDIF
\RETURN $S$
\end{algorithmic}
\end{algorithm}

\subsection{QA Selection}
We apply quality control and quantity balancing to ensure reliable QA generation.

\subsubsection{Ambiguity Reduction.} 
\begin{itemize}[leftmargin=*]
    \item To avoid ambiguous references when questions involve object names, we require that only a single instance of queried semantic object category is present within the image set.
    \item We discard cases in which the angular range of a referenced direction cannot be clearly mapped to a unique spatial sector (\eg, front/left/right). Such ambiguous geometric configurations are removed to avoid confusion in answer interpretation.
\end{itemize}

\subsubsection{Balanced Sampling.}  
We encourage both textual and visual diversity while maintaining balanced sampling.
\begin{itemize}[leftmargin=*]
    \item For questions with the same underlying intent, we randomly vary the textual descriptions (\eg, paraphrased phrasings and directional expressions), while capping the total number of samples to avoid redundancy.
    \item Within each image set, we select diverse combinations of objects to construct QAs, while limiting the number of QAs per set to maintain a balanced distribution across different scenes.
\end{itemize}

\subsection{Chain-of-Thought (CoT) Strategies}

We explore three Chain-of-Thought (CoT) strategies for multi-frame reasoning.

\noindent\textbf{VLM-generated CoT.} 
We provide QA pairs and step-wise instructions to GPT-5 for CoT annotations.

\noindent\textbf{MindCube Aug-CGMap CoT.} 
MindCube~\cite{mindcube} uses discrete 2D cognitive map (CogMap) to describe top-down view of the scenes, projecting objects and cameras onto a 2D grid. The designed CoT contains two steps:
\begin{itemize}
    \item \textbf{CGMap Inference.} Directly generate a JSON-formatted CogMap with a discretized grid (\eg, $10 \times 10$), encoding approximate positions and four-direction orientations of objects and camera views.
    \item \textbf{Free-Form Reasoning.} Perform free-form reasoning on camera changes between consecutive frames, relate observations across views, and finally derive the answer from the aggregated observations.
\end{itemize}

\noindent\textbf{Our Procedural CGMap CoT.}
We also adopt a top-down CogMap representation, but use continuous (non-gridded) coordinates and construct the CogMap in a procedural, step-by-step manner interleaved with textual reasoning. This CoT design exhibits improved geometric accuracy and more coherent reasoning, particularly in scenes with complex object layouts and diverse viewpoints. Experimental results can be found in~\cref{sec:exp:spatial_cot}. The detailed procedure is as follows:
\begin{itemize}
  \item \textbf{Keyframe Localization.} Identify the keyframe set \(A\) in which the queried objects appear. These frames are emphasized during subsequent reasoning.
  \item \textbf{Incremental Relative Camera Estimation.} Traverse frames in temporal order. For each adjacent pair of frames, describe the shared objects and estimate coarse camera pose changes.
  \item \textbf{CogMap Construction.} Construct the CogMap along the keyframe set \(A\), following an efficient path inferred from the previous step. We build a global CogMap and fix its origin and positive \(y\)-axis with the first keyframe as reference.  For each new frame, we perform metric 3D grounding of newly observed objects, estimate the camera rotation and translation relative to the reference frame, and then transform the placement of the new camera view and objects into the global CogMap.
  \item \textbf{Answer Derivation.} We may flexibly rotate the CogMap coordinate system according to any desired allocentric transformation, and reason about geometric relations (\eg, distances, angles, relative ordering) to produce the final answer.
\end{itemize}

% \subsection{Capability Intersecting.}
% \input{tables/5_4_mixed_training}
% We observe in our training data mixing data from diverse sources shows synergetic effects. In~\cref{tab:5_4_mixed_training}

% Set A primarily captures pairwise geometric transformations, serving as a basis for understanding how objects and cameras move relative to one another. 

% Set B expands this paradigm by incorporating multi-image samples, which capture holistic scene-level reasoning. 

% This dataset includes questions involving spatial relationships among multiple objects distributed across distinct views, and thus demands the model to aggregate visual information globally. 

% Set C consists of single-image data, focusing on object-level semantic orientation—such as describing the facing direction of a sofa or chair.

% Individually, none of these datasets is sufficient for robust multi-view generalization. 

% However, combining Set A and Set B yields a marked improvement, as Set A supplies essential geometric priors while Set B promotes complex contextual reasoning. 

% The best and most consistent results arise when all three datasets (A+B+C) are trained jointly. 

% This full configuration unifies geometric, relational, and semantic cues, allowing the model to reason more holistically across diverse visual domains. The improvement in MMSI metrics demonstrates the complementary of mixed training.

% \TODO{\begin{CJK}{UTF8}{gbsn}C任务 = A任务 + B任务: C在训练中没见过. @ruisi\end{CJK}}

\section{Full Results of Single Dataset Training}
For each dataset, we train a model on its training set and evaluate its performance on five key spatial intelligence benchmarks: VSI, MMSI, MindCube, ViewSpatial, and SITE.
As shown in~\cref{tab:single_dataset}, training on a single dataset often yields strong performance on a few benchmarks while sacrificing performance on others. For example, a model trained solely on VSI-590K achieves the best MRA accuracy on VSI (64.0), but its results on tasks like MMSI, MindCube, and SITE drop noticeably.

\input{tables/supp_tables/single_dataset}

This pattern highlights that no single dataset provides comprehensive spatial intelligence coverage, and therefore mixed-data training is crucial for building more balanced models. As different datasets tend to bias the model toward a particular subset of spatial reasoning skills, determining how to effectively balance datasets during training remains an open challenge. This table also provides the community with a useful reference for dataset selection, helping researchers understand which datasets contribute to which aspects of spatial reasoning and how to design more complementary training mixtures.

\section{Capability Transfer Matrix}

\input{tables/supp_tables/capability_transfer}

We study the capability transfer matrix by training \name-8B on MM, SR, and PT with sufficient training data ($>$1M), and observe that there exists cross-capability synergy in~\cref{tab:capability_transfer}. For example, PT may act as a more comprehensive capability benefiting others.

% \subsection{The importance of general capability} 
% Grounding etc.

\section{Impact of Scaling on Benchmarks} 
In~\cref{fig:teaser}, we illustrate the effect of data scaling on model performance, grouped by core spatial capabilities. In contrast,~\cref{tab:5_3_data_scaling} presents the scaling effects at the benchmark level. Across both views, we observe a consistent trend: model performance improves steadily as more data is introduced during training, validating that high-quality, diverse spatial data is effective in addressing key knowledge deficiencies in spatial intelligence.
\input{tables/5_3_data_scaling}

\section{Retention of General Capabilities}
To evaluate whether \name retains its general understanding capabilities after continued training on spatial intelligence data (\ie, \name-8M), we evaluate its performance on additional multimodal benchmarks:
\textbf{MMBench-En}~\cite{liu2024mmbench} and \textbf{MMStar}~\cite{chen2024rightwayevaluatinglarge} for holistic multimodal understanding,
\textbf{AI2D}~\cite{Hiippala_2020} for scientific diagram reasoning,
\textbf{OCRB}~\cite{Liu_2024} and \textbf{DocVQA}~\cite{mathew2021docvqa} for text-rich image and document understanding,
\textbf{MMVP}~\cite{tong2024mmvp} and \textbf{V*}~\cite{wu2024v} for fine-grained visual perception and grounding,
\textbf{MMMU}~\cite{yue2024mmmu} for multidisciplinary multimodal reasoning, and
\textbf{Vid-MME}~\cite{fu2025videomme} for video understanding.

As shown in~\cref{tab:main_results_general}, \name\textsubscript{Qwen3-VL-8B} and \name\textsubscript{InternVL3-2B} exhibit minimal performance drops on MMBench-En relative to their respective base models, while \name\textsubscript{Bagel-7B-MoT} and \name\textsubscript{InternVL3-8B} even show slight improvements. Across the remaining three benchmarks, only marginal declines are observed.

Notably, compared with other open-source spatial-intelligence models, \name maintains competitive general visual understanding.
Furthermore, prior studies (such as VST~\cite{vst2025} and Cambrian-S~\cite{yang2025cambrian}) suggest that incorporating additional general visual understanding data can further preserve or enhance this capability, a direction we plan to explore in future work.

\input{tables/main_results_general}

\section{Text-only Training}

To further investigate the potential impact of language shortcuts, we conduct an experiment in which models are trained on text-only data, with all visual inputs removed (\cref{tab:text_only_training}). The results show that text-only training yields substantially smaller gains, indicating that visual input is critical. Moreover, we observe that certain benchmarks are highly resistant to language shortcuts. For example, on MMSI-Bench, text-only training brings only minimal improvements.

\input{tables/supp_tables/text_only_training}

\section{Downstream Task}

\subsection{\name as an Embodied Agent for Manipulation.}
Following EmbodiedBench~\cite{yang2025embodiedbench}, we implement \name as an embodied agent in the downstream manipulation task to demonstrate its application. In this setting, the \name model controls a simulated Franka Panda robot with a parallel gripper. Conditioned on a language instruction and the visual state of the scene, the agent receives a symbolic description of the environment, where each object is represented by a discrete 3D position in a table-top coordinate frame. The model is required to output a sequence of low-level gripper actions in a structured action space, where each action specifies the target end-effector position, orientation, and a binary gripper state, all expressed in the same discretized coordinate system. 
This formulation enables direct execution of the predicted actions in the simulator without additional post-processing. 
This demonstrates the model’s ability to ground language and perception into coherent, executable manipulation trajectories that require spatial reasoning. 
% The detailed definitions of the input space and output action space are provided in the task prompt for the agent shown below.

\subsection{Task Prompt for the Embodied Agent.}
The official task prompt (OP) includes the role description, the definitions of the input space and output action space, the color space, example task conversations, and the instructions for the output JSON format. This design enables the \name model to perform reasoning and action planning while generating executable actions in the required format. 
For the spatial-intelligence-oriented prompt (SIP), instead of providing object bounding box coordinates with generic indices such as "object 1" and "object 2" as in OP, we provide the specific name of the objects, such as "first cylinder” or "second container". This removes interference from object recognition and allows the model to focus on spatial reasoning.
The official task prompt (OP) is shown below.

\subsection{Case Study.}
\cref{fig:demo_chat} illustrates how the \name\textsubscript{InternVL3-8B} model behaves as an embodied agent in the manipulation task under the official task prompt. For each task instance, we show the task instruction, the scene observation provided to the model, the model output, and the resulting execution rollout in the simulator. Cases \cref{fig:demo_chat}(a) and \cref{fig:demo_chat}(b) demonstrate successful executions, where the model produces correct object recognition and accurate action sequences that lead to successful task completion. In contrast, cases \cref{fig:demo_chat}(c) and \cref{fig:demo_chat}(d) illustrate typical failure modes, including incorrect object recognition in the visual state description that leads to erroneous reasoning and execution planning, as well as limited manipulation precision that causes task failure even when the executable action plan is correct. Incorrect information in the output is highlighted in red. Please refer to the video demonstration in the \Supp for the full rollout, more task instances and performance comparison with the base InternVL3-8B model.

\begin{figure*}[t]   % 星号：跨两栏
\centering
\fbox{%
\ttfamily\footnotesize
    \begin{minipage}{1\textwidth}  % 黑框内部的正文区域
        \input{figures/embodied_prompt_part1}
    \end{minipage}%
}
\end{figure*}

\clearpage
\begin{figure*}[t]   % 星号：跨两栏
\centering
\fbox{%
\ttfamily\footnotesize
    \begin{minipage}{1\textwidth}  % 黑框内部的正文区域
        \input{figures/embodied_prompt_part2}
    \end{minipage}%
}
\end{figure*}

\clearpage

\input{figures/demo_chat}
\clearpage

\section{Detail Results on Key Benchmarks}
% vsi, mmsi, mindcube, viewspatial, site, mmbench
In this section, we provide detailed per-benchmark results that complement the aggregated scores reported in the main text. 
The following tables report per-benchmark results for VSI-Bench (\cref{tab:vsi_oo}), MMSI-Bench (\cref{tab:mmsi_oo}), MindCubeBench-Tiny (\cref{tab:mindcube_oo}), ViewSpatial (\cref{tab:viewspatial_oo}), SITE (\cref{tab:site_oo}), BLINK (\cref{tab:blink_oo}), 3DSR (\cref{tab:3dsrbench_oo}), and Embspatial (\cref{tab:embspatial_oo}), respectively. 
For each benchmark, we break down performance over all relevant subsets and question types, enabling a more fine-grained analysis of model strengths and failure modes than is possible from the single aggregated metrics in the main tables. 
On most subsets, our model attains the best or near-best accuracy among open-source models, and on several challenging subsets (\eg, Rel.Dir in VSI), its performance is comparable to, or even surpasses, that of proprietary models.

\input{tables/supp_tables/vsi}
% \input{tables/supp_tables/vsi_4_tech}

\input{tables/supp_tables/mmsi}

% \input{tables/supp_tables/mmsi_4_tech}

\input{tables/supp_tables/mindcube}
% \input{tables/supp_tables/mindcube_4_tech}

\input{tables/supp_tables/viewspatial}
% \input{tables/supp_tables/viewspatial_4_tech}

\input{tables/supp_tables/site}

\input{tables/supp_tables/blink}

\input{tables/supp_tables/3dsr}

\input{tables/supp_tables/embspatial}

% \input{tables/main_results_spatial_4_tech}

% \section{in the wild case}
% @??
% image level 

% \section{performance of more foundation model}
% @fanyi

% \section{Rationale}
% \label{sec:rationale}
% % 
% Having the supplementary compiled together with the main paper means that:
% % 
% \begin{itemize}
% \item The supplementary can back-reference sections of the main paper, for example, we can refer to \cref{sec:intro};
% \item The main paper can forward reference sub-sections within the supplementary explicitly (\eg referring to a particular experiment); 
% \item When submitted to arXiv, the supplementary will already included at the end of the paper.
% \end{itemize}
% % 
% To split the supplementary pages from the main paper, you can use \href{https://support.apple.com/en-ca/guide/preview/prvw11793/mac#:~:text=Delete%20a%20page%20from%20a,or%20choose%20Edit%20%3E%20Delete).}{Preview (on macOS)}, \href{https://www.adobe.com/acrobat/how-to/delete-pages-from-pdf.html#:~:text=Choose%20%E2%80%9CTools%E2%80%9D%20%3E%20%E2%80%9COrganize,or%20pages%20from%20the%20file.}{Adobe Acrobat} (on all OSs), as well as \href{https://superuser.com/questions/517986/is-it-possible-to-delete-some-pages-of-a-pdf-document}{command line tools}.

\clearpage

\section{Case Study on Text-based CoT}
\label{sec:case_study_text_cot}
% @chenyang successful vs failure cases

As discussed in~\cref{sec:exp:spatial_cot}, we observe that textual CoT does not appear to yield genuinely stronger 3D reasoning capabilities. With supporting case studies in \cref{fig:cot_case1} and \cref{fig:cot_case2}, we tentatively consider several hypotheses rather than definitive explanations, and we leave a more thorough investigation to future work. Note that \cotsuchl{green} indicates correct reasoning steps, while \cotfailhl{red} indicates wrong reasoning steps.

\noindent\textbf{Train–test mismatch and error accumulation.}
Textual CoT enforces a long, linear per-token reasoning process. During training, tokens are learned with ground-truth context, whereas at inference time the model conditions on its own generated tokens. Early local errors in the reasoning (as in \cref{fig:cot_case2}) may therefore accumulate along the sequence and become increasingly difficult to correct; tasks that require a globally coherent 3D structure may be particularly sensitive to this discrepancy.

\noindent\textbf{Bias toward human-preferred explanation paths.}
Compared to direct QA supervision, textual CoT shifts the optimization focus from task accuracy towards reproducing specific intermediate verbal rationalizations, which may not align with the geometry-dominant nature of multi-view 3D reasoning. Supervision along a single, human-preferred explanatory path may effectively shrink the model's reasoning space.

% \noindent\textbf{Structural mismatch with latent 3D representations.}
% Textual CoT impose a discrete, linear serialization of inherently continuous, graph-structured 3D relations, potentially misaligning with the model’s latent geometric representations.

\clearpage

\input{figures/cot_case1}

\begin{figure*}[t]   % 星号：跨两栏
\centering
\noindent
\fbox{%
\ttfamily\footnotesize
    \begin{minipage}{1\textwidth}  % 黑框内部的正文区域
        \input{figures/cot_case1_text_part1}
      \end{minipage}%
}
\end{figure*}
\begin{figure*}[t]   % 星号：跨两栏
\centering
\noindent
\fbox{%
\ttfamily\footnotesize
    \begin{minipage}{1\textwidth}  % 黑框内部的正文区域
        \input{figures/cot_case1_text_part2}
    \end{minipage}%
}
\end{figure*}

\input{figures/cot_case2}

\begin{figure*}[t]   % 星号：跨两栏
\centering
\noindent
\fbox{%
\ttfamily\footnotesize
    \begin{minipage}{1\textwidth}  % 黑框内部的正文区域
        \input{figures/cot_case2_text_part1}
      \end{minipage}%
}
\end{figure*}
\begin{figure*}[t]   % 星号：跨两栏
\centering
\noindent
\fbox{%
\ttfamily\footnotesize
    \begin{minipage}{1\textwidth}  % 黑框内部的正文区域
        \input{figures/cot_case2_text_part2}
    \end{minipage}%
}
\end{figure*}
\begin{figure*}[t]   % 星号：跨两栏
\centering
\noindent
\fbox{%
\ttfamily\footnotesize
    \begin{minipage}{1\textwidth}  % 黑框内部的正文区域
        \input{figures/cot_case2_text_part3}
    \end{minipage}%
}
\end{figure*}

%% file: figures/case_messytable.tex
\begin{figure}[t]
    \centering
    \setlength{\belowcaptionskip}{-2pt}
    \includegraphics[width=0.6\linewidth]{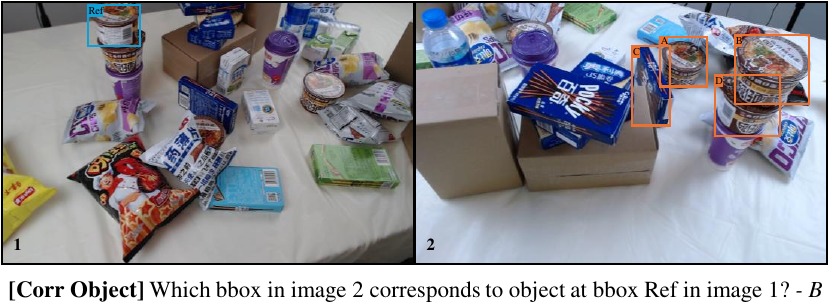}
    \caption{
        Hard cases in MessyTable~\cite{cai2020messytable}, where multiple instances of the same object class are present in the same scene.
    }
    \label{fig:case_messytable}
\end{figure}

%% file: tables/supp_tables/single_dataset.tex
\begin{table*}[h] \centering \small \setlength{\tabcolsep}{4pt} 
% \resizebox{1\linewidth}{!}{

\begin{tabular}{lcccccccccc|c}
\toprule

& \multicolumn{2}{c}{\textbf{VSI-Bench}} 
& \multicolumn{2}{c}{\textbf{MMSI-Bench}} 
& \multicolumn{2}{c}{\textbf{MindCube}} 
& \multicolumn{2}{c}{\textbf{ViewSpatial}} 
& \multicolumn{2}{c}{\textbf{SITE}}
& \textbf{Avg.} \\[-1pt]

\cmidrule(lr){2-3}
\cmidrule(lr){4-5}
\cmidrule(lr){6-7}
\cmidrule(lr){8-9}
\cmidrule(lr){10-11}

\textbf{Dataset} 
& MRA, Acc & $\star$ 
& Acc & $\star$
& Acc & $\star$
& Acc & $\star$
& CAA & $\star$
& -- \\
\midrule

% \midrule
% \rowcolor{line-blue}\textbf{Human Evaluation} & & & & & & \\
% \footnotesize $\Delta$(Best Model,Human) &
% \colornum{-10.42} &
% \colornum{-5.62} &
% \colornum{-53.90} &
% \colornum{-8.88} &
% - & \colornum{-28.97} \\

% \midrule
% \textbf{Human} & \textbf{79.2} & \textbf{97.2}& \textbf{94.5} & - & \textbf{67.5} & - \\

% \midrule

\textbf{Random Choice} & 34.0 & - & 25.0 & - & 33.0 & - & 26.3 & - & 0.0 & - & 23.7 \\ 
InternVL3-8B~\cite{zhu2025internvl3} & 42.1 & - & 28.0 & - & 41.5 & - & 38.6 & - & 41.1 & - & 38.3 \\ 
\midrule 
VLM-3R-DATA~\cite{vlm3r} & \cellcolor{secondcolor}{53.9} & 2 & \cellcolor{secondcolor}{28.5} & 2 & 34.8 & 3 & 54.2 & 4 & 36.7 & 3 & \cellcolor{bestcolor}{\textbf{41.6}} \\
VSR~\cite{VSR} & 41.1 & 3 & 28.3 & 3 & \cellcolor{secondcolor}{37.9} & 2 & \cellcolor{secondcolor}{55.9} & 2 & \cellcolor{bestcolor}{\textbf{40.9}} & 1 & \cellcolor{secondcolor}{40.8} \\
Rel3D~\cite{goyal2020rel3d} & 39.3 & 4 & 26.9 & 7 & \cellcolor{bestcolor}{\textbf{39.8}} & 1 & \cellcolor{bestcolor}{\textbf{57.7}} & 1 & \cellcolor{secondcolor}{39.5} & 2 & 40.6 \\
VSI590K~\cite{yang2025cambrian} & \cellcolor{bestcolor}{\textbf{64.0}} & 1 & \cellcolor{bestcolor}{\textbf{29.0}} & 1 & 26.7 & 7 & 48.1 & 5 & 34.7 & 5 & 40.5 \\
SPEC~\cite{SPEC} & 38.7 & 5 & 27.6 & 5 & 33.6 & 4 & 54.7 & 3 & 34.8 & 4 & 37.9 \\
SAT~\cite{ray2024sat} & 30.5 & 6 & 26.8 & 9 & 29.1 & 6 & 42.8 & 6 & 21.3 & 8 & 30.1 \\
GQA~\cite{hudson2019gqa} & 26.6 & 9 & 27.4 & 6 & 24.7 & 8 & 42.2 & 7 & 23.4 & 7 & 28.9 \\
MultiSPA~\cite{xu2025multi} & 21.8 & 10 & 27.7 & 4 & 22.8 & 9 & 34.4 & 9 & 32.2 & 6 & 27.8 \\
CLEVR~\cite{johnson2017clevr} & 29.7 & 8 & 25.6 & 10 & 30.0 & 5 & 33.1 & 10 & 18.6 & 10 & 27.4 \\
VQA~\cite{Antol2015vqa} & 30.0 & 7 & 26.9 & 7 & 15.3 & 10 & 39.4 & 8 & 20.2 & 9 & 26.4 \\

\bottomrule
\end{tabular}
% }
\vspace{-2pt}
\caption{\textbf{Evaluation on key spatial intelligence benchmarks using InternVL3-8B trained on each single dataset.}
$\star$: ranking on each benchmark. \textbf{\raisebox{0pt}[0pt][0pt]{\colorbox{bestcolor}{Dark purple}}} highlights the best result and \raisebox{0pt}[0pt][0pt]{\colorbox{secondcolor}{light purple}} indicates the second-best result within models trained on different single datasets, respectively.}

\label{tab:single_dataset}
\end{table*}

%% file: tables/supp_tables/capability_transfer.tex
\begin{table}[t]
\centering
\small
\setlength{\tabcolsep}{1.2mm}
\begin{tabular}{lcccccc}
\toprule
Train & MM & MR & SR & PT & CR  \\
\midrule
% SSI\textsubscript{InternVL3-8B} & +61.2 & +107.6 & +36.4 & +78.5 & +51.5 \\
% \hline
MM & +52.5\% & -7.6\% & +18.4\% & +6.4\% & +8.4\% \\
SR & +10.6\% & -61.5\% & +18.7\% & -19.2\% & -6.0\% \\
PT & +3.0\% & +46.1\% & -0.2\% & +83.8\% & +10.9\%  \\
% CR & -19.9 & -15.3 & 0.0 & -5.9 & +34.7 \\
\bottomrule
\end{tabular}
\caption{\textbf{Capability transfer matrix.} Each row shows the percentage change in model performance on other key spatial capabilities after training on data corresponding to a specific capability.}
\label{tab:capability_transfer}
\end{table}

% \begin{table}[t]
% \centering
% \small
% \setlength{\tabcolsep}{1.0mm}
% \captionsetup{skip=2pt}
% \begin{tabular}{cccccc}
% \toprule
% \multicolumn{2}{c}{\textbf{GPT-4o}} &
% \multicolumn{2}{c}{\textbf{InternVL3-8B}} &
% \multicolumn{2}{c}{\textbf{\name\textsubscript{InternVL3-8B}}} \\
% \cmidrule(lr){1-2} \cmidrule(lr){3-4} \cmidrule(lr){5-6}
% OP & SIP & OP & SIP & OP & SIP \\
% \midrule
% 37.5 & 45.8 & 10.4 & 20.8 & \textbf{16.6} (+59.6\%) & \textbf{33.3} (+60.0\%) \\
% \bottomrule
% \end{tabular}
% \caption{
% Success rate on \textbf{Spatial} subset of EmbodiedBench~\cite{yang2025embodiedbench}.
% \textbf{OP}: Official Prompt; \textbf{SIP}: Spatial-Intelligence-oriented Prompt.
% }
% \label{tab:embodied_bench}
% \end{table}

%% file: tables/5_3_data_scaling.tex
\begin{table}[h]
\vspace{-1mm}
\centering
\small
\setlength{\tabcolsep}{1.0mm}
% \resizebox{1\linewidth}{!}{
{
\begin{tabular}{lccccc} % 9 columns (1 model name + 4 original + 4 new)
\toprule
\textbf{\# Data} &
% \textbf{Avg.} &
% \textbf{Rank} &
\textbf{VSI-Bench} &
\textbf{MMSI-Bench} &
\textbf{MindCube$^*$} &
\textbf{ViewSpatial} &
\textbf{SITE} \\

\midrule
\rowcolor{line-blue}\textbf{Metric} & MRA, Acc & Acc & Acc & Acc & CAA\\
\midrule
% \rowcolor{line-blue}0M(\textbf{InternVL3-8B}) & 42.14 & 58.39 & 28.00 & 41.54 & 42.52 \\
0M & 42.1 & 28.0 & 41.5 & 52.0 & 42.1 \\
1M & 56.3 & 36.0 & 58.8 & 55.2 & 44.5 \\
2M & 60.3 & 39.6 & 76.2 & 56.7 & 47.1 \\
3M & 64.4 & 41.9 & 81.1 & 56.2 & 46.7 \\
4M & 62.7 & 41.9 & 83.9 & 55.6 & 45.2 \\
5M & 65.9 & 40.8 & 81.7 & 53.7 & 46.0 \\
6M & 66.3 & 41.8 & 85.0 & 55.2 & 47.0 \\
7M & 67.9 & 42.3 & 85.7 & 54.7 & 46.5 \\
% 8M(\textbf{SenseSI\textsubscript{InternVL3-8B}}) & \textbf{68.78} & \textbf{63.05} & \textbf{43.30} & \textbf{85.67} & \textbf{65.20} \\
8M & \textbf{68.7} & \textbf{43.3} & \textbf{85.6} & \textbf{54.6} & \textbf{47.7} \\

\bottomrule
\end{tabular}}
% }
\vspace{-2pt}
\caption{\textbf{Impact of scaling on key spatial intelligence benchmarks}. MindCube$^*$ denotes MindCube-Tiny. 0M indicates base model (InternVL3-8B) whereas 8M indicates \name\textsubscript{InternVL3-8B}.}
\label{tab:5_3_data_scaling}
\end{table}

%% file: tables/main_results_general.tex
\begin{table}[h] 
% \vspace{-1mm}
\centering
% \small
\setlength{\tabcolsep}{1.2pt}
\resizebox{1\linewidth}{!}{
\begin{tabular}{lccccccccc}
\toprule
\textbf{Models} & \textbf{MMB-EN~\cite{liu2024mmbench}} & \textbf{MMStar~\cite{chen2024rightwayevaluatinglarge}} & \textbf{AI2D~\cite{Hiippala_2020}} & \textbf{OCRB~\cite{Liu_2024}} & \textbf{DocVQA~\cite{mathew2021docvqa}} & \textbf{MMVP~\cite{tong2024mmvp}} & \textbf{V*~\cite{wu2024v}} & \textbf{MMMU~\cite{yue2024mmmu}} & \textbf{Vid-MME~\cite{fu2025videomme}} \\
\midrule

\rowcolor{line-blue}\textbf{Open-source General Models} & & & & & & & & & \\

Bagel-7B-MoT~\cite{deng2025bagel} & \cellcolor{secondcolor}{82.8} & 67.3 & \cellcolor{bestcolor}{\textbf{89.5}} & 811 & \cellcolor{secondcolor}{94.1} & \cellcolor{bestcolor}{\textbf{68.7}} & \cellcolor{secondcolor}{71.7} & 50.4 & 68.9 \\

% Qwen2.5-VL-3B-Instruct~\cite{Qwen2.5-VL} & 77.4 & 55.9 & 81.6 & 797 & - & - & - & - & - \\

% Qwen2.5-VL-7B-Instruct~\cite{Qwen2.5-VL} & 82.6 & 63.9 & 83.9 & 864 & 94.6 & 54.7 & 75.4 & 49.9 & 71.1 \\

Qwen3-VL-8B~\cite{Qwen3-VL} & \cellcolor{bestcolor}{\textbf{84.6}} & \cellcolor{bestcolor}{\textbf{68.5}} & 81.3 & \cellcolor{bestcolor}{\textbf{906}} & \cellcolor{bestcolor}{\textbf{95.8}} & 58.7 & \cellcolor{bestcolor}{\textbf{82.7}} & \cellcolor{bestcolor}{\textbf{60.9}} & \cellcolor{bestcolor}{\textbf{78.1}} \\

InternVL3-2B~\cite{zhu2025internvl3} & 79.7 & 56.9 & 77.8 & 853 & 87.3 & 56.7 & 68.6 & 43.2 & 69.9 \\

InternVL3-8B~\cite{zhu2025internvl3} & 81.7 & \cellcolor{secondcolor}{68.2} & \cellcolor{secondcolor}{85.2} & \cellcolor{secondcolor}{880} & 92.1 & \cellcolor{secondcolor}{59.3} & 57.1 & \cellcolor{secondcolor}{55.6} & \cellcolor{secondcolor}{76.1} \\

\rowcolor{line-blue}\textbf{Open-source SI Models} & & & & & & & & & \\

% MindCube-3B-SFT-RawQA~\cite{mindcube} & 32.3 & 35.9 & 65.8 & 392 & - & - & - & - & - \\

% SpatialLadder-3B~\cite{li2025spatialladder} & 72.5 & 48.7 & 76.1 & 468 & - & - & - & - & - \\

% SpaceR-7B~\cite{ouyang2025spacer} & 75.4 & 58.6 & 84.3 & 873 & - & - & - & - & - \\

% ViLaSR-7B~\cite{wu2025reinforcing(vilasr)} & 81.1 & 62.9 & 84.7 & 865 \\

VST-7B-SFT~\cite{vst2025} &
83.3$^\dagger$ &
63.1$^\dagger$ &
\cellcolor{secondcolor}{84.9$^\dagger$} &
\cellcolor{secondcolor}{855$^\dagger$} & 91.7 & 54.7 & \cellcolor{bestcolor}{\textbf{81.2}} & \cellcolor{secondcolor}{50.9} & 72.1 \\

Cambrian-S-7B~\cite{yang2025cambrian} &
80.4$^\dagger$ &
43.9 &
76.9$^\dagger$ &
648$^\dagger$ & 83.7 & 54.0 & 70.7 & 47.1 & 70.6 \\

\rowcolor{line-blue}\textbf{Ours} & & & & & & & & & \\

\textbf{\name\textsubscript{Bagel-7B-MoT}} & 83.4 & \cellcolor{bestcolor}{\textbf{67.8}} & \cellcolor{bestcolor}{\textbf{88.8}} & 797 & \cellcolor{secondcolor}{93.9} & \cellcolor{bestcolor}{\textbf{65.3}} & 69.6 & 50.2 & 64.6 \\

\textbf{\name\textsubscript{Qwen3-VL-8B}} & \cellcolor{secondcolor}{83.5} & \cellcolor{secondcolor}{65.7} & 84.2 & \cellcolor{bestcolor}{\textbf{863}} & \cellcolor{bestcolor}{\textbf{95.4}} & \cellcolor{secondcolor}{57.3} & 81.1 & \cellcolor{bestcolor}{\textbf{57.6}} & \cellcolor{bestcolor}{\textbf{76.4}} \\

\textbf{\name\textsubscript{InternVL3-2B}} & 78.9 & 57.0 & 76.8 & 817 & 85.2 & 47.3 & \cellcolor{bestcolor}{\textbf{81.2}} & 44.4 & 66.6 \\

\textbf{\name\textsubscript{InternVL3-8B}} & \cellcolor{bestcolor}{\textbf{84.9}} & 65.4 & 79.0 & 815 & 84.9 & 56.0 & 71.2 & 49.4 & \cellcolor{secondcolor}{72.2} \\

\bottomrule
\end{tabular}
}

\vspace{-2pt}
\caption{\textbf{Evaluation on general undestanding benchmarks.}
\emph{$\dagger$ denotes benchmark results directly cited from their papers.}
\textbf{\raisebox{0pt}[0pt][0pt]{\colorbox{bestcolor}{Dark purple}}} highlights the best result and \raisebox{0pt}[0pt][0pt]{\colorbox{secondcolor}{light purple}} indicates the second-best result within Open-source and \name models, respectively.
MMB-EN: MMBench-En. OCRB: OCRBench.
}
\label{tab:main_results_general}
\vspace{-6pt}
\end{table}

% %% Original data refers to https://pgzodr3heu.feishu.cn/wiki/SvClwK3uxi8ZaFkkh4qcK7x5nbd?from=from_copylink

%% file: tables/supp_tables/text_only_training.tex
\begin{table}[h]
\centering
\small
\caption{Training with text-only SSI-800K (10\% of SSI-8M).}
\label{tab:text_only_training}
% \vspace{-4mm}
\setlength{\tabcolsep}{1.4mm}
{
\resizebox{\linewidth}{!}{
\begin{tabular}{lcccccc}
\hline
Model & VSI-Debiased & VSI-Bench & MMSI-Bench & MindCube-Tiny & ViewSpatial & SITE \\
\hline
InternVL3-8B          & 38.5 & 42.1 & 28.0 & 41.5 & 38.6 & 41.1 \\
+SSI-800K             & 53.8 & 60.9 & 36.4 & 56.9 & 52.5 & 47.7 \\
+SSI-800K (Text-Only) & 42.4 & 50.2 & 28.2 & 44.2 & 44.1 & 42.5 \\
\hline
\end{tabular}}
}
\end{table}

%% file: figures/embodied_prompt_part1.tex
\textbf{The official task prompt (OP) for the embodied agent:} 

You are a Franka Panda robot with a parallel gripper. You can perform various tasks and output a sequence of gripper actions to accomplish a given task with images of your status. The input space, output action space and color space are defined as follows.\\

Input space

- Each input object is represented as a 3D discrete position in the following format: [X, Y, Z].

- There is a red XYZ coordinate frame located in the top-left corner of the table. The X-Y plane is the table surface.

- The allowed range of X, Y, Z is [0, 100].

- Objects are ordered by Y in ascending order.\\

Output action space

- Each output action is represented as a 7D discrete gripper action in the following format: [X, Y, Z, Roll, Pitch, Yaw, Gripper state].

- X, Y, Z are the 3D discrete position of the gripper in the environment. They follow the same coordinate system as the input object coordinates.

- The allowed range of X, Y, Z is [0, 100].

- Roll, Pitch, Yaw are the 3D discrete orientation of the gripper in the environment, represented as discrete Euler angles.

- The allowed range of Roll, Pitch, Yaw is [0, 120] and each unit represents 3 degrees.

- Gripper state is 0 for close and 1 for open.\\

Color space

- Each object can be described using one of the colors below:
  ["red", "maroon", "lime", "green", "blue", "navy", "yellow", "cyan", "magenta", "silver", "gray", "olive", "purple", "teal", "azure", "violet", "rose", "black", "white"].\\

Below are some examples to guide you in completing the task.\\

Example 0:
Human Instruction: Pick up the star and place it into the yellow container.\\
Input: \{"object 1": [45, 13, 18], "object 2": [72, 20, 18], "object 3": [50, 42, 17], "object 4": [36, 42, 18], "object 5": [70, 43, 17], "object 6": [50, 77, 17]\}

Output: \{
    "visual\_state\_description": "From left to right, I can see a purple cylinder at [45, 13, 18], a blue moon at [72, 20, 18], a green star at [50, 42, 17], a purple cube at [36, 42, 18], a yellow container at [70, 43, 17] and a purple container at [50, 77, 17].",
    "reasoning\_and\_reflection": "The task is to pick up the star located at [50, 42, 17] and place it into the yellow container located at [70, 43, 17]. The plan involves moving the gripper to the star's position, closing the gripper to pick it up, moving to the yellow container's position, and opening the gripper to release the object into the container.",
    "executable\_plan": "[[50, 42, 17, 6, 61, 36, 1], [50, 42, 17, 6, 61, 36, 0], [47, 41, 26, 6, 61, 36, 0], [69, 39, 24, 88, 32, 77, 0], [69, 39, 24, 87, 32, 77, 1]]"
\}\\

Example 1:
Human Instruction: Pick up the star and place it into the rose container.\\
Input: \{"object 1": [78, 20, 17], "object 2": [28, 30, 17], "object 3": [39, 47, 17],
        "object 4": [50, 72, 17], "object 5": [39, 75, 18]\}
        
Output: \{
    "visual\_state\_description": "From left to right, I can see a rose container at [78, 20, 17], a black star at [28, 30, 17], a green container at [39, 47, 17], a green triangular prism at [50, 72, 17] and a grey cylinder at [39, 75, 18].",
    "reasoning\_and\_reflection": "The task is to pick up the star located at [28, 30, 17] and place it into the rose container located at [78, 20, 17]. The plan involves moving the gripper to the star's position, closing the gripper to pick it up, moving to the rose container's position, and opening the gripper to release the object into the container.",
    "executable\_plan": "[[28, 32, 26, 0, 60, 94, 1], [27, 32, 18, 0, 60, 94, 0], [27, 32, 28, 0, 60, 94, 0], [76, 19, 28, 6, 66, 109, 1]]"
\}\\

Now you are supposed to follow the above examples to generate a sequence of discrete gripper actions that completes the human instruction below.\\

Human Instruction: <Task instruction>\\
Input: <Dict of object positions>
    
The output JSON format should be \{"visual\_state\_description": str, "reasoning\_and\_reflection": str, "language\_plan": str, "executable\_plan": str\}.\\

%% file: figures/embodied_prompt_part2.tex
The fields in the above JSON follow the purposes below:\\

1. visual\_state\_description: Describe the color and shape of each object in the detection box in the numerical order in the image. Then provide the 3D coordinates of the objects chosen from the input.

2. reasoning\_and\_reflection: Reason about the overall plan that needs to be taken on the target objects, and reflect on the previous actions taken if available.

3. language\_plan: A list of natural language actions to achieve the user instruction. Each language action is started by the step number and the language action name.

4. executable\_plan: A list of discrete actions needed to achieve the user instruction, with each discrete action being a 7-dimensional discrete action.

5. Keep your plan efficient and concise.

When generating content for JSON strings, avoid using any contractions or abbreviated forms (like 's, 're, 've, 'll, 'd, n't) that use apostrophes. Instead, write out full forms (is, are, have, will, would, not) to prevent parsing errors in JSON.

Please do not output anything other than the above-mentioned JSON.

%% file: figures/demo_chat.tex
\begin{figure*}[t]
    \centering
    \setlength{\belowcaptionskip}{-2pt}
    \includegraphics[width=0.854\textwidth]
    {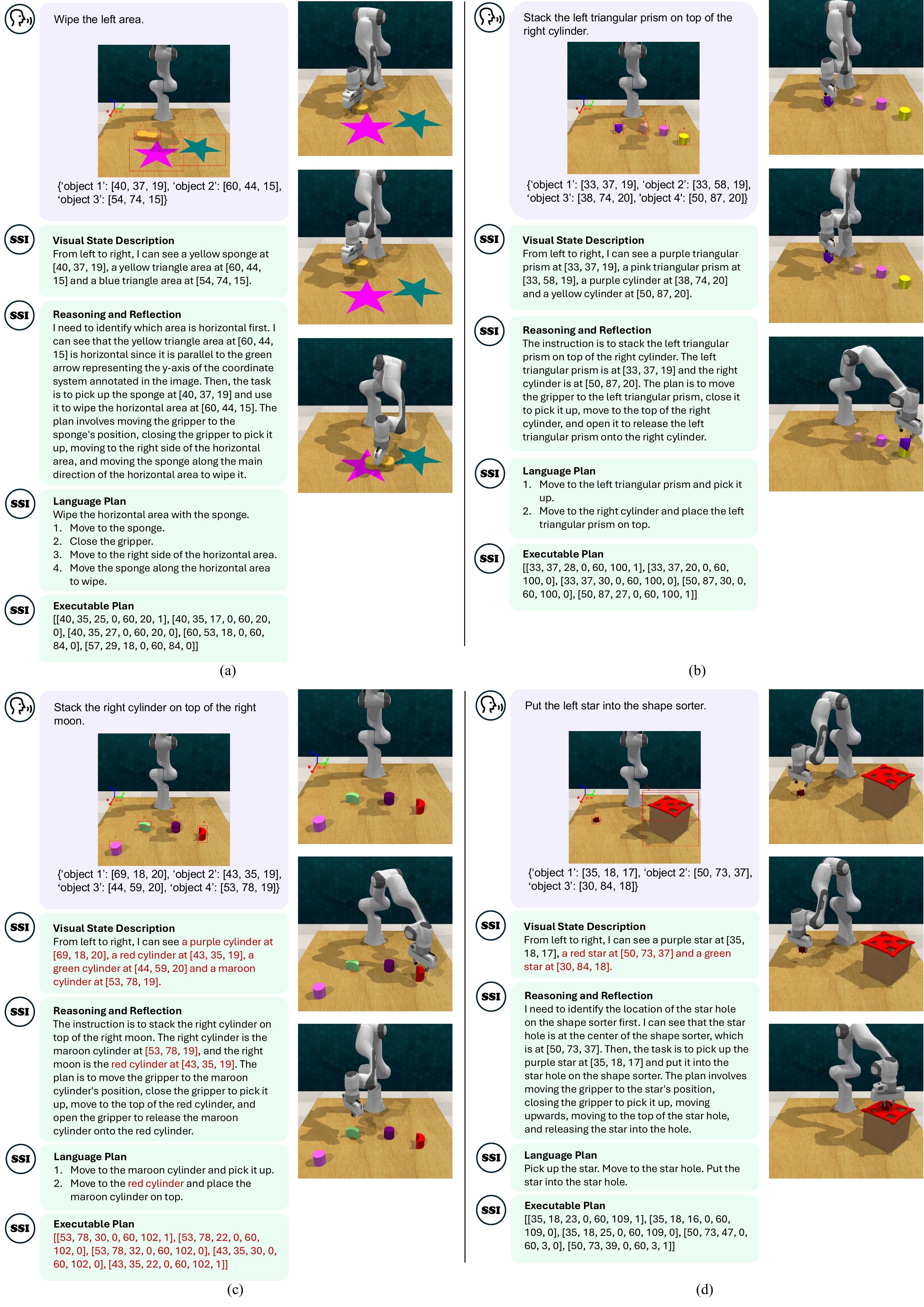}
    \caption{
    Demonstration of manipulation with \name\textsubscript{InternVL3-8B} as an embodied agent. We present the task instruction, scene observation input, model output, and task execution rollout. Cases (a) and (b) are successful executions, while cases (c) and (d) are failure cases. Incorrect information in the output is highlighted in red. Please see the video in the Supplementary Material for the full rollout.
    }
    \label{fig:demo_chat}
\end{figure*}

%% file: tables/supp_tables/vsi.tex
% =================== VSI-Bench (from Excel auto-generated) ===================
\begin{table*}[t]          % 1. 一定是 table*，不要嵌在别的环境里
\centering
\small
\setlength{\tabcolsep}{2pt}
\resizebox{1.0\linewidth}{!}{% 2. 先不用 adjustbox，和上面那张表保持一致
\begin{tabular}{l*{9}{c}}
\toprule
\multirow{2}{*}{\textbf{Models}} & \multirow{2}{*}{\textbf{Avg.}} &
\multicolumn{4}{c}{\textbf{Numerical Answer}} &
\multicolumn{4}{c}{\textbf{Multiple-Choice Answer}} \\
\cmidrule(lr){3-6}\cmidrule(lr){7-10}
& \multicolumn{1}{c}{} &
\makecell{\textbf{Obj. Count}\\\textcolor{sensepurple}{\scriptsize CR}} &
\makecell{\textbf{Abs. Dist}\\\textcolor{sensepurple}{\scriptsize MM}} &
\makecell{\textbf{Obj. Size}\\\textcolor{sensepurple}{\scriptsize MM}} &
\makecell{\textbf{Room Size}\\\textcolor{sensepurple}{\scriptsize MM}} &
\makecell{\textbf{Rel. Dis}\\\textcolor{sensepurple}{\scriptsize SR,MM}} &
\makecell{\textbf{Rel. Dir}\\\textcolor{sensepurple}{\scriptsize PT}} &
\makecell{\textbf{Route Plan}\\\textcolor{sensepurple}{\scriptsize CR}} &
\makecell{\textbf{Appr. Order}\\\textcolor{sensepurple}{\scriptsize CR}} \\

\midrule
\textbf{Human} & \textbf{79.2} & \textbf{94.3} & \textbf{47.0} & \textbf{60.4} & \textbf{45.9} & \textbf{94.7} & \textbf{95.8} & \textbf{95.8} & \textbf{100.0} \\

% \midrule
Random Choice(Frequency) & 34.0 & 62.1 & 32.0 & 29.9 & 33.1 & 25.1 & 47.9 & 28.4 & 25.2 \\

\midrule
\rowcolor{line-blue}\multicolumn{10}{l}{\textbf{Proprietary Models}} \\

Seed-1.6-2025-06-15~\cite{seed2025seed1_5vl}
& 49.9 & 43.5 & 34.4 & 66.1 & \cellcolor{secondcolor}{52.8} & 55.1 & 35.7 & 44.3 & 68.0 \\

Gemini-2.5-pro-2025-06~\cite{team2023gemini}
& \cellcolor{secondcolor}{53.6} & \cellcolor{secondcolor}{46.0} & \cellcolor{bestcolor}{\textbf{37.4}} & \cellcolor{secondcolor}{68.7} & \cellcolor{bestcolor}{\textbf{54.4}} & \cellcolor{secondcolor}{62.0} & \cellcolor{secondcolor}{43.9} & \cellcolor{secondcolor}{47.4} & \cellcolor{secondcolor}{68.8} \\

Grok-4-2025-07-09~\cite{grok4_xai_2025}
& 47.9 & 37.2 & 33.0 & 60.8 & 45.4 & 53.1 & 39.7 & 47.4 & 66.8 \\

GPT-5-2025-08-07~\cite{openai_gpt5_systemcard}
& \cellcolor{bestcolor}{\textbf{55.0}} & \cellcolor{bestcolor}{\textbf{53.3}} & \cellcolor{secondcolor}{34.5} & \cellcolor{bestcolor}{\textbf{73.3}} & 47.5 & \cellcolor{bestcolor}{\textbf{63.7}} & \cellcolor{bestcolor}{\textbf{48.7}} & \cellcolor{bestcolor}{\textbf{50.3}} & \cellcolor{bestcolor}{\textbf{68.9}} \\

\midrule
\rowcolor{line-blue}\multicolumn{10}{l}{\textbf{Open-source General Models}} \\

Bagel-7B-MoT~\cite{deng2025bagel}
& 31.4 & 30.1 & 29.2 & 35.5 & 25.8 & 34.9 & 41.4 & 30.4 & 24.1 \\

Qwen2.5-VL-3B-Instruct~\cite{Qwen2.5-VL}
& 27.0 & 19.2 & 21.2 & 24.3 & 27.3 & 33.8 & 42.1 & 27.3 & 20.9 \\

Qwen2.5-VL-7B-Instruct~\cite{Qwen2.5-VL}
& 32.3 & 32.9 & 18.2 & 43.9 & 31.7 & 38.0 & 37.4 & 28.4 & 28.0 \\

Qwen3-VL-2B-Instruct~\cite{Qwen3-VL}
& 50.4 & 62.2 & 40.3 & 71.5 & 49.8 & 52.3 & 42.0 & 30.4 & 54.5 \\

Qwen3-VL-8B-Instruct~\cite{Qwen3-VL}
& 57.9 & 67.6 & 47.0 & \cellcolor{secondcolor}{76.3} & 61.9 & 58.0 & 51.0 & 35.1 & 66.3 \\

InternVL3-2B~\cite{zhu2025internvl3}
& 33.0 & 64.8 & 30.9 & 32.4 & 23.0 & 32.3 & 34.9 & 33.0 & 12.6 \\

InternVL3-8B~\cite{zhu2025internvl3}
& 42.1 & 66.1 & 34.9 & 43.6 & 47.5 & 48.0 & 39.3 & 26.3 & 31.4 \\

\rowcolor{line-blue}\multicolumn{10}{l}{\textbf{Open-source Spatial Intelligence Models}} \\

MindCube-3B-RawQA-SFT~\cite{mindcube}
& 17.2 & 12.9 & 22.8 & 4.3 & 23.5 & 20.3 & 15.7 & 16.0 & 22.5 \\

SpatialLadder-3B~\cite{li2025spatialladder}
& 44.9 & 62.2 & 35.4 & 62.0 & 41.4 & 45.6 & 46.5 & 27.3 & 38.5 \\

Spatial-MLLM-4B~\cite{wu2025spatial}
& 46.3 & 66.7 & 38.1 & 63.7 & 35.5 & 40.4 & 48.2 & 33.0 & 44.3 \\

SpaceR-7B~\cite{ouyang2025spacer}
& 41.6 & 30.0 & 25.2 & 47.0 & 29.6 & 40.3 & 46.5 & 32.5 & 39.3 \\

ViLaSR-7B~\cite{vilasr}
& 44.6 & 58.1 & 33.9 & 61.4 & 28.9 & 45.1 & 46.5 & 29.9 & 53.2 \\

VST-3B-SFT~\cite{vst2025}
& 51.4 & 60.7 & 37.5 & 72.7 & 45.9 & 51.3 & 45.9 & 40.2 & 56.8 \\

VST-7B-SFT~\cite{vst2025}
& 55.5 & 68.9 & 37.3 & 74.5 & 62.2 & 55.2 & 48.7 & 41.8 & 55.5 \\

Cambrian-S-3B~\cite{yang2025cambrian}
& 56.1 & 69.4 & 38.7 & 66.3 & 52.7 & 61.8 & 58.3 & 28.4 & 73.1 \\

Cambrian-S-7B~\cite{yang2025cambrian}
& 62.9 & 68.2 & 45.8 & 72.5 & 67.6 & \cellcolor{secondcolor}{66.8} & 69.6 & 39.2 & 73.8 \\

\rowcolor{line-blue}\multicolumn{10}{l}{\textbf{Ours}} \\
\textbf{\name\textsubscript{Bagel-7B-MoT}} & 41.5 & 42.2 & 33.5 & 57.2 & 22.7 & 44.9 & 46.4 & 33.5 & 51.9 \\

\textbf{\name\textsubscript{Qwen3-VL-8B}} & \cellcolor{secondcolor}{64.8} & \cellcolor{secondcolor}{71.4} & \cellcolor{secondcolor}{48.7} & 76.0 & \cellcolor{secondcolor}{69.6} & 65.5 & 72.2 & \cellcolor{secondcolor}{43.8} & 71.5 \\

\textbf{\name\textsubscript{InternVL3-2B}} & 63.7 & 70.1 & 47.2 & 74.5 & 67.1 & 61.0 & \cellcolor{secondcolor}{73.1} & 41.2 & \cellcolor{secondcolor}{75.6} \\

\textbf{\name\textsubscript{InternVL3-8B}} & \cellcolor{bestcolor}{\textbf{68.8}} & \cellcolor{bestcolor}{\textbf{72.0}} & \cellcolor{bestcolor}{\textbf{53.5}} & \cellcolor{bestcolor}{\textbf{76.8}} & \cellcolor{bestcolor}{\textbf{72.8}} & \cellcolor{bestcolor}{\textbf{69.6}} & \cellcolor{bestcolor}{\textbf{80.8}} & \cellcolor{bestcolor}{\textbf{48.5}} & \cellcolor{bestcolor}{\textbf{76.4}} \\

% \midrule
% \rowcolor{line-blue}\multicolumn{10}{l}{\textbf{Human Evaluation}} \\
% \footnotesize $\Delta$(Best Model,Human) & 
% \colornum{-} &
% \colornum{-} &
% \colornum{} &
% \colornum{} &
% \colornum{} &
% \colornum{} &
% \colornum{} &
% \colornum{-} &
% \colornum{-} \\

% Human & \textbf{79.2} & \textbf{94.3} & \textbf{47.0} & \textbf{60.4} & \textbf{45.9} & \textbf{94.7} & \textbf{95.8} & \textbf{95.8} & \textbf{100.0} \\

\bottomrule
\end{tabular}
}

\vspace{-2pt}
\caption{\textbf{Evaluation on VSI-Bench~\cite{vsi}.}
Numerical Answer uses \textbf{MRA} score, MCA uses \textbf{Acc} score, \textbf{Avg.} is the simple average across these metrics, following the original paper.
All results are evaluated on EASI~\cite{cai2025has}.
}
\label{tab:vsi_oo}

\end{table*}

%% file: tables/supp_tables/mmsi.tex
% =================== MMSI (auto from Excel) ===================
\begin{table*}[ht!]
\centering
\small                      % 或 \footnotesize / \scriptsize
\setlength{\tabcolsep}{1.2pt}
\renewcommand{\arraystretch}{1}
\begin{adjustbox}{max width=\textwidth}
\begin{tabular}{l*{12}{c}}
\toprule
\multirow{2}{*}{\textbf{Models}} &
\multirow{2}{*}{\textbf{Avg.}} &
\multicolumn{6}{c}{\textbf{Positional Relationship}} &
\multicolumn{2}{c}{\textbf{Attribute}} &
\multicolumn{2}{c}{\textbf{Motion}} &
\multicolumn{1}{c}{\textbf{MSR}} \\
\cmidrule(lr){3-8}\cmidrule(lr){9-10}\cmidrule(lr){11-12}\cmidrule(lr){13-13}
& \multicolumn{1}{c}{} &
\makecell{\textbf{C--C}\\\textcolor{sensepurple}{\scriptsize PT}} &
\makecell{\textbf{O--O}\\\textcolor{sensepurple}{\scriptsize PT}} &
\makecell{\textbf{R--R}\\\textcolor{sensepurple}{\scriptsize PT}} &
\makecell{\textbf{C--O}\\\textcolor{sensepurple}{\scriptsize PT}} &
\makecell{\textbf{O--R}\\\textcolor{sensepurple}{\scriptsize PT}} &
\makecell{\textbf{C--R}\\\textcolor{sensepurple}{\scriptsize PT}} &
\makecell{\textbf{Meas.}\\\textcolor{sensepurple}{\scriptsize MM}} &
\makecell{\textbf{Appr.}\\\textcolor{sensepurple}{\scriptsize MR}} &
\makecell{\textbf{Cam.}\\\textcolor{sensepurple}{\scriptsize PT}} &
\makecell{\textbf{Obj.}\\\textcolor{sensepurple}{\scriptsize PT}} &
\makecell{\textbf{--}\\\textcolor{sensepurple}{\scriptsize CR}} \\

\midrule

\textbf{Human} & \textbf{97.2} & \textbf{95.7} & \textbf{98.9} & \textbf{97.5} & \textbf{94.2} & \textbf{98.8} & \textbf{96.4} & \textbf{95.3} & \textbf{98.5} & \textbf{98.6} & \textbf{98.7} & \textbf{97.0} \\

Random Choice & 25.0 & 25.0 & 25.0 & 25.0 & 25.0 & 25.0 & 25.0 & 25.0 & 25.0 & 25.0 & 25.0 & 25.0 \\

\midrule
\rowcolor{line-blue}\multicolumn{13}{l}{\textbf{Proprietary Models}} \\

Seed-1.6-2025-06-15~\cite{seed2025seed1_5vl} & \cellcolor{secondcolor}{38.3} & 36.6 & \cellcolor{bestcolor}{\textbf{36.2}} & 32.1 & 32.6 & \cellcolor{secondcolor}{42.4} & 46.9 & 48.4 & \cellcolor{secondcolor}{33.0} & 31.1 & \cellcolor{secondcolor}{42.1} & \cellcolor{bestcolor}{\textbf{40.4}} \\

Gemini-2.5-pro-2025-06~\cite{team2023gemini} & 38.0 & \cellcolor{secondcolor}{38.7} & 34.0 & \cellcolor{bestcolor}{\textbf{40.7}} & \cellcolor{secondcolor}{44.2} & 38.8 & 41.0 & \cellcolor{bestcolor}{\textbf{62.5}} & 30.3 & \cellcolor{secondcolor}{39.2} & 25.0 & 33.3 \\

Grok-4-2025-07-09~\cite{grok4_xai_2025} & 37.8 & 36.6 & \cellcolor{secondcolor}{35.1} & \cellcolor{secondcolor}{39.5} & 34.9 & \cellcolor{bestcolor}{\textbf{45.9}} & \cellcolor{secondcolor}{50.6} & 21.9 & 22.7 & \cellcolor{bestcolor}{\textbf{40.5}} & \cellcolor{bestcolor}{\textbf{43.4}} & \cellcolor{secondcolor}{38.4} \\

GPT-5-2025-08-07~\cite{openai_gpt5_systemcard} & \cellcolor{bestcolor}{\textbf{41.8}} & \cellcolor{bestcolor}{\textbf{41.9}} & 33.0 & 35.8 & \cellcolor{bestcolor}{\textbf{49.8}} & 42.4 & \cellcolor{bestcolor}{\textbf{68.7}} & \cellcolor{secondcolor}{54.7} & \cellcolor{bestcolor}{\textbf{37.4}} & 28.3 & 40.8 & 36.4 \\

\midrule
\rowcolor{line-blue}\multicolumn{13}{l}{\textbf{Open-source General Models}} \\

Bagel-7B-MoT~\cite{deng2025bagel} & 31.0 & 34.4 & 35.1 & 29.6 & 32.6 & \cellcolor{secondcolor}{42.4} & 31.3 & 34.4 & 21.2 & 18.9 & 27.6 & \cellcolor{bestcolor}{\textbf{30.3}} \\

Qwen2.5-VL-3B-Instruct~\cite{Qwen2.5-VL} & 28.6 & 36.6 & 30.9 & 28.4 & 26.7 & 28.2 & 31.3 & 31.2 & 16.7 & 16.2 & \cellcolor{bestcolor}{\textbf{35.5}} & 28.8 \\

Qwen2.5-VL-7B-Instruct~\cite{Qwen2.5-VL} & 26.8 & 28.0 & 26.6 & 19.8 & 32.6 & 38.8 & 28.9 & 23.4 & 21.2 & 20.3 & 30.3 & 24.8 \\

Qwen3-VL-2B-Instruct~\cite{Qwen3-VL} & 28.9 & 26.9 & 29.8 & 30.9 & 38.4 & 35.3 & 33.7 & 23.4 & 28.8 & 29.7 & 28.9 & 21.2 \\

Qwen3-VL-8B-Instruct~\cite{Qwen3-VL} & 31.1 & 28.0 & 37.2 & 32.1 & 31.4 & 35.3 & 38.5 & 37.5 & 15.2 & 27.0 & 28.9 & \cellcolor{secondcolor}{29.8} \\

InternVL3-2B~\cite{zhu2025internvl3} & 26.5 & 31.2 & 22.3 & 28.4 & 30.2 & 28.2 & 28.9 & 25.0 & 22.7 & 16.2 & 28.9 & 26.8 \\

InternVL3-8B~\cite{zhu2025internvl3} & 28.0 & 22.6 & 22.3 & 34.6 & 31.4 & \cellcolor{secondcolor}{42.4} & 33.7 & 25.0 & 19.7 & 20.3 & \cellcolor{secondcolor}{34.2} & 24.8 \\

\rowcolor{line-blue}\multicolumn{13}{l}{\textbf{Open-source Spatial Intelligence Models}} \\

MindCube-3B-RawQA-SFT~\cite{mindcube} & 1.7 & 0.0 & 2.1 & 2.5 & 2.3 & 2.4 & 3.6 & 3.1 & 0.0 & 0.0 & 0.0 & 2.0 \\

SpatialLadder-3B~\cite{li2025spatialladder} & 27.4 & 36.6 & 29.8 & 29.6 & 32.6 & 30.6 & 24.1 & 18.8 & 31.8 & 23.0 & 23.7 & 23.2 \\

Spatial-MLLM-4B~\cite{wu2025spatial} & 26.1 & 24.7 & 21.3 & 28.4 & 30.2 & 29.4 & 28.9 & 18.8 & 34.9 & 10.8 & 23.7 & \cellcolor{secondcolor}{29.8} \\

SpaceR-7B~\cite{ouyang2025spacer} & 27.4 & 25.8 & 31.9 & 29.6 & 25.6 & 31.8 & 22.9 & 26.6 & 28.8 & 16.2 & \cellcolor{secondcolor}{34.2} & 27.3 \\

ViLaSR-7B~\cite{vilasr} & 30.2 & 29.0 & 35.1 & 28.4 & 39.5 & 40.0 & 44.6 & 31.2 & 16.7 & 17.6 & 31.6 & 23.2 \\

VST-3B-SFT~\cite{vst2025} & 28.8 & 32.3 & 31.9 & 28.4 & 27.9 & 23.5 & 36.1 & 32.8 & 34.9 & 27.0 & 28.9 & 22.7 \\

VST-7B-SFT~\cite{vst2025} & 32.5 & 39.8 & 36.2 & \cellcolor{secondcolor}{35.8} & 37.2 & 29.4 & 33.7 & 29.7 & \cellcolor{bestcolor}{\textbf{47.0}} & \cellcolor{bestcolor}{\textbf{36.5}} & \cellcolor{bestcolor}{\textbf{35.5}} & 18.2 \\

Cambrian-S-3B~\cite{yang2025cambrian} & 27.0 & 25.8 & 28.7 & 24.7 & 48.8 & 24.7 & 33.7 & 29.7 & 22.7 & 20.3 & 28.9 & 18.7 \\

Cambrian-S-7B~\cite{yang2025cambrian} & 27.1 & 24.7 & 26.6 & 24.7 & 47.7 & 22.4 & 31.3 & 32.8 & 24.2 & 12.2 & 30.3 & 24.2 \\

\rowcolor{line-blue}\multicolumn{13}{l}{\textbf{Ours}} \\

\textbf{\name\textsubscript{Bagel-7B-MoT}} & 34.5 & \cellcolor{secondcolor}{48.4} & 34.0 & 23.5 & 46.5 & 34.1 & 41.0 & 34.4 & 33.3 & \cellcolor{secondcolor}{32.4} & 32.9 & 26.8 \\

\textbf{\name\textsubscript{Qwen3-VL-8B}} & \cellcolor{secondcolor}{38.1} & 44.1 & 38.3 & 33.3 & \cellcolor{bestcolor}{\textbf{65.1}} & 38.8 & \cellcolor{secondcolor}{59.0} & \cellcolor{bestcolor}{\textbf{48.4}} & 24.2 & 29.7 & \cellcolor{secondcolor}{34.2} & 22.2 \\

\textbf{\name\textsubscript{InternVL3-2B}} & 34.2 & 39.8 & \cellcolor{secondcolor}{45.7} & 33.3 & 46.5 & 30.6 & 39.8 & 31.2 & 30.3 & 29.7 & 32.9 & 24.8 \\

\textbf{\name\textsubscript{InternVL3-8B}}& \cellcolor{bestcolor}{\textbf{43.3}} & \cellcolor{bestcolor}{\textbf{50.5}} & \cellcolor{bestcolor}{\textbf{47.9}} & \cellcolor{bestcolor}{\textbf{42.0}} & \cellcolor{secondcolor}{62.8} & \cellcolor{bestcolor}{\textbf{44.7}} & \cellcolor{bestcolor}{\textbf{69.9}} & \cellcolor{secondcolor}{40.6} & \cellcolor{secondcolor}{40.9} & \cellcolor{secondcolor}{32.4} & 32.9 & 27.8 \\

% \midrule
% \rowcolor{line-blue}\multicolumn{13}{l}{\textbf{Human Evaluation}} \\

% \footnotesize $\Delta$(Best Model,Human) & \colornum{-53.90} & \colornum{-45.16} & \colornum{-51.03} & \colornum{-55.52} & \colornum{-31.41} & \colornum{-52.92} & \colornum{-26.52} & \colornum{-32.80} & \colornum{-53.05} & \colornum{-58.06} & \colornum{-55.28} & \colornum{-56.60} \\

\bottomrule
\end{tabular}
\end{adjustbox}

\vspace{-2pt}
\caption{\textbf{Evaluation on \textbf{MMSI-Bench}~\cite{yang2025mmsi}}. Scores are \textbf{Acc} as in the original paper. Under Positional Relationship, C: Camera; O: Object; R: Region. All results are evaluated on EASI~\cite{cai2025has}.
}
\label{tab:mmsi_oo}
\end{table*}

%% file: tables/supp_tables/mindcube.tex
% =================== MindCube (auto from Excel) ===================
\begin{table*}[ht!]
\centering
\setlength{\tabcolsep}{1.2pt}
\begin{tabular}{l *{4}{c}}
\toprule
\textbf{Models} & \textbf{Avg.} & \makecell{\textbf{Rotation}\\\textcolor{sensepurple}{\scriptsize PT}} & \makecell{\textbf{Among}\\\textcolor{sensepurple}{\scriptsize PT}} & \makecell{\textbf{Around}\\\textcolor{sensepurple}{\scriptsize PT}} \\

\midrule
\textbf{Human} & 94.5 & - & - & - \\
Random Choice & 33.0 & 33.3 & 31.8 & 35.7 \\

\midrule
\rowcolor{line-blue}\textbf{Proprietary Models} & & & & \\

Seed-1.6-2025-06-15~\cite{seed2025seed1_5vl} & 48.8 & 89.0 & 36.4 & 45.6 \\
Gemini-2.5-pro-2025-06~\cite{team2023gemini} & \cellcolor{secondcolor}{57.6} & 88.0 & \cellcolor{secondcolor}{44.9} & \cellcolor{secondcolor}{63.2} \\
Grok-4-2025-07-09~\cite{grok4_xai_2025} & \cellcolor{bestcolor}{\textbf{63.6}} & \cellcolor{secondcolor}{93.0} & \cellcolor{bestcolor}{\textbf{54.4}} & 61.6 \\
GPT-5-2025-08-07~\cite{openai_gpt5_systemcard} & 56.3 & \cellcolor{bestcolor}{\textbf{94.5}} & 38.2 & \cellcolor{bestcolor}{\textbf{68.4}} \\

\midrule
\rowcolor{line-blue}\textbf{Open-source General Models} & & & & \\

Bagel-7B-MoT~\cite{deng2025bagel} & 34.7 & 34.5 & 31.4 & 42.8 \\
Qwen2.5-VL-3B-Instruct~\cite{Qwen2.5-VL} & 37.6 & 33.5 & 35.9 & 44.8 \\
Qwen2.5-VL-7B-Instruct~\cite{Qwen2.5-VL} & 36.0 & 37.0 & 32.4 & 44.0 \\
Qwen3-VL-2B-Instruct~\cite{Qwen3-VL} & 34.5 & 32.5 & 31.7 & 42.8 \\
Qwen3-VL-8B-Instruct~\cite{Qwen3-VL} & 29.4 & 29.5 & 28.6 & 31.2 \\
InternVL3-2B~\cite{zhu2025internvl3} & 37.5 & 29.0 & 37.0 & 45.6 \\
InternVL3-8B~\cite{zhu2025internvl3} & 41.5 & 36.5 & 38.1 & 53.6 \\

\rowcolor{line-blue}\textbf{Open-source Spatial Intelligence Models} & & & & \\

MindCube-3B-RawQA-SFT~\cite{mindcube} & 51.7 & 34.0 & 51.0 & 67.6 \\

SpatialLadder-3B~\cite{li2025spatialladder} & 43.5 & 35.0 & 43.2 & 50.8 \\

Spatial-MLLM-4B~\cite{wu2025spatial} & 33.5 & 39.0 & 30.5 & 36.0 \\
SpaceR-7B~\cite{ouyang2025spacer} & 38.0 & 35.0 & 34.2 & 49.2 \\
ViLaSR-7B~\cite{vilasr} & 35.1 & 35.5 & 31.0 & 44.4 \\
VST-3B-SFT~\cite{vst2025} & 36.0 & 32.0 & 34.9 & 41.6 \\
VST-7B-SFT~\cite{vst2025} & 39.7 & 37.0 & 35.9 & 50.8 \\
Cambrian-S-3B~\cite{yang2025cambrian} & 38.4 & 28.0 & 40.0 & 42.8 \\
Cambrian-S-7B~\cite{yang2025cambrian} & 37.9 & 33.0 & 39.0 & 39.2 \\

\rowcolor{line-blue}\textbf{Ours} & & & & \\

\textbf{\name\textsubscript{Bagel-7B-MoT}} & 46.8 & 33.0 & 50.7 & 48.8 \\

\textbf{\name\textsubscript{Qwen3-VL-8B}} & \cellcolor{secondcolor}{73.8} & \cellcolor{secondcolor}{79.5} & \cellcolor{secondcolor}{73.2} & \cellcolor{secondcolor}{70.4} \\

\textbf{\name\textsubscript{InternVL3-2B}} & 41.8 & 30.5 & 46.4 & 40.0 \\

\textbf{\name\textsubscript{InternVL3-8B}} & \cellcolor{bestcolor}{\textbf{85.7}} & \cellcolor{bestcolor}{\textbf{82.0}} & \cellcolor{bestcolor}{\textbf{84.9}} & \cellcolor{bestcolor}{\textbf{90.4}} \\

% \midrule
% \rowcolor{line-blue}\textbf{Human Evaluation} & & & & \\%
% \footnotesize $\Delta$(Best Model,Human) & \colornum{-8.88} & - & - & - \\
% Human & 94.5 & - & - & - \\

\bottomrule
\end{tabular}
\vspace{-2pt}
\caption{
\textbf{Evaluation on \textbf{MindCube-Tiny}~\cite{mindcube}}.
All scores are \textbf{Acc}.
}
\label{tab:mindcube_oo}
\end{table*}

%% file: tables/supp_tables/viewspatial.tex
% =================== ViewSpatial (auto from Excel) ===================
\begin{table*}[ht!]
\centering\small
\setlength{\tabcolsep}{1.2pt}
\begin{adjustbox}{max width=\linewidth}
\begin{tabular}{l*{6}{c}}
\toprule
\multirow{2}{*}{\textbf{Models}} &
\multirow{2}{*}{\textbf{Overall}} &
\multicolumn{2}{c}{\textbf{Camera-based Tasks}} &
\multicolumn{3}{c}{\textbf{Person-based Tasks}} \\
\cmidrule(lr){3-4}\cmidrule(lr){5-7}
& \multicolumn{1}{c}{} &
\makecell{\textbf{Rel. Dir.}\\\textcolor{sensepurple}{\scriptsize PT}} &
\makecell{\textbf{Obj. Ori.}\\\textcolor{sensepurple}{\scriptsize PT}} &
\makecell{\textbf{Obj. Ori.}\\\textcolor{sensepurple}{\scriptsize PT}} &
\makecell{\textbf{Rel. Dir.}\\\textcolor{sensepurple}{\scriptsize PT}} &
\makecell{\textbf{Sec. Sim.}\\\textcolor{sensepurple}{\scriptsize PT}} \\

\midrule
% \textbf{Human} & \textbf{79.2} & \textbf{94.3} & \textbf{47.0} & \textbf{60.4} & \textbf{45.9} & \textbf{95.8}\\

% \midrule
Random Choice & 26.3 & 25.2 & 26.1 & 24.6 & 31.1 & 26.3 \\

\midrule
\rowcolor{line-blue}\multicolumn{7}{l}{\textbf{Proprietary Models}} \\

Seed-1.6-2025-06-15~\cite{seed2025seed1_5vl} & 43.9 & 55.8 & 26.9 & \cellcolor{bestcolor}{\textbf{54.8}} & \cellcolor{secondcolor}{48.5} & 26.6 \\

Gemini-2.5-pro-2025-06~\cite{team2023gemini} & \cellcolor{bestcolor}{\textbf{46.1}} & \cellcolor{secondcolor}{59.1} & \cellcolor{bestcolor}{\textbf{33.0}} & \cellcolor{secondcolor}{51.0} & 45.8 & \cellcolor{secondcolor}{32.6} \\

Grok-4-2025-07-09~\cite{grok4_xai_2025} & 43.2 & 57.1 & 23.9 & 47.6 & \cellcolor{bestcolor}{\textbf{51.7}} & 24.9 \\

GPT-5-2025-08-07~\cite{openai_gpt5_systemcard} & \cellcolor{secondcolor}{45.6} & \cellcolor{bestcolor}{\textbf{60.2}} & \cellcolor{secondcolor}{27.9} & 41.0 & 48.5 & \cellcolor{bestcolor}{\textbf{40.1}} \\

\midrule
\rowcolor{line-blue}\multicolumn{7}{l}{\textbf{Open-source General Models}} \\
Bagel-7B-MoT~\cite{deng2025bagel} & 41.3 & 48.3 & \cellcolor{secondcolor}{38.6} & 47.0 & 42.5 & 26.5 \\
Qwen2.5-VL-3B-Instruct~\cite{Qwen2.5-VL} & 32.0 & 40.8 & 28.7 & 30.1 & 29.2 & 24.4 \\
Qwen2.5-VL-7B-Instruct~\cite{Qwen2.5-VL} & 36.9 & 46.8 & 31.2 & 40.0 & 32.4 & 26.6 \\
Qwen3-VL-2B-Instruct~\cite{Qwen3-VL} & 37.0 & 49.6 & 23.8 & 35.3 & 32.7 & 33.3 \\
Qwen3-VL-8B-Instruct~\cite{Qwen3-VL} & 42.2 & 54.2 & 29.7 & 47.3 & 40.3 & 31.1 \\
InternVL3-2B~\cite{zhu2025internvl3} & 32.6 & 42.0 & 17.6 & 38.9 & 34.6 & 23.7 \\
InternVL3-8B~\cite{zhu2025internvl3} & 38.7 & 50.3 & 27.5 & 42.6 & 37.5 & 27.3 \\

\rowcolor{line-blue}\multicolumn{7}{l}{\textbf{Open-source Spatial Intelligence Models}} \\
MindCube-3B-RawQA-SFT~\cite{mindcube} & 24.1 & 30.9 & 22.1 & 23.7 & 22.4 & 16.9 \\

SpatialLadder-3B~\cite{li2025spatialladder} & 39.9 & 46.2 & 25.7 & 56.2 & 31.9 & 33.6 \\

Spatial-MLLM-4B~\cite{wu2025spatial} & 34.7 & 35.0 & 23.4 & 40.4 & 40.4 & 34.6 \\

SpaceR-7B~\cite{ouyang2025spacer} & 35.9 & 43.2 & 28.9 & 37.5 & 34.1 & 30.2 \\

ViLaSR-7B~\cite{vilasr} & 35.7 & 46.8 & 25.3 & 39.1 & 32.7 & 26.6 \\

VST-3B-SFT~\cite{vst2025} & \cellcolor{secondcolor}{52.9} & 46.9 & 35.4 & \cellcolor{bestcolor}{\textbf{70.3}} & \cellcolor{bestcolor}{\textbf{52.6}} & 62.8 \\

VST-7B-SFT~\cite{vst2025} & 50.5 & 52.7 & 29.6 & 51.9 & \cellcolor{secondcolor}{50.7} & \cellcolor{secondcolor}{64.5} \\

Cambrian-S-3B~\cite{yang2025cambrian} & 41.0 & 47.0 & 21.0 & 50.1 & 39.8 & 42.0 \\

Cambrian-S-7B~\cite{yang2025cambrian} & 41.3 & 50.4 & 22.7 & 45.0 & 38.8 & 41.9 \\

\rowcolor{line-blue}\multicolumn{7}{l}{\textbf{Ours}} \\

\textbf{\name\textsubscript{Bagel-7B-MoT}} & 46.9 & 54.7 & 33.5 & 45.9 & 43.9 & 49.5 \\

\textbf{\name\textsubscript{Qwen3-VL-8B}} & 51.2 & 60.3 & 22.0 & 67.8 & 41.5 & 55.6 \\

\textbf{\name\textsubscript{InternVL3-2B}} & 52.7 & \cellcolor{secondcolor}{65.7} & 19.5 & \cellcolor{secondcolor}{70.2} & 39.4 & 55.9 \\

\textbf{\name\textsubscript{InternVL3-8B}} & \cellcolor{bestcolor}{\textbf{54.7}} & \cellcolor{bestcolor}{\textbf{66.3}} & \cellcolor{bestcolor}{\textbf{43.2}} & 38.4 & 43.1 & \cellcolor{bestcolor}{\textbf{70.0}} \\

\bottomrule
\end{tabular}
\end{adjustbox}
\vspace{-2pt}
\caption{\textbf{Evaluation on \textbf{ViewSpatial-Bench}~\cite{li2025viewspatial}.} All scores are \textbf{Acc}.}
\label{tab:viewspatial_oo}
\end{table*}

%% file: tables/supp_tables/site.tex
% =================== SITE (CAA; auto from Excel) ===================
\begin{table*}[ht!]
\centering\small
\setlength{\tabcolsep}{1.2pt}
\begin{adjustbox}{max width=\linewidth}
\begin{tabular}{l *{7}{c}}
\toprule
\textbf{Models} &\textbf{Overall} &\makecell{\textbf{Count}\\\textcolor{sensepurple}{\scriptsize -}} &\makecell{\textbf{Loc}\\\textcolor{sensepurple}{\scriptsize -}} &\makecell{\textbf{3D Inf}\\\textcolor{sensepurple}{\scriptsize MM,SR}} &\makecell{\textbf{MultiV}\\\textcolor{sensepurple}{\scriptsize PT}} &\makecell{\textbf{Rel}\\\textcolor{sensepurple}{\scriptsize SR}} &\makecell{\textbf{Mov}\\\textcolor{sensepurple}{\scriptsize CR}} \\

\midrule
\textbf{Human} & \textbf{67.5} & \textbf{66.0} & \textbf{83.3} & \textbf{54.7} & \textbf{87.5} & \textbf{73.0} & \textbf{52.5} \\
Random Choice & 0.0 & 0.0 & 0.0 & 0.0 & 0.0 & 0.0 & 0.0 \\

\midrule
\rowcolor{line-blue}\multicolumn{8}{l}{\textbf{Proprietary Models}} \\
Seed-1.6-2025-06-15~\cite{seed2025seed1_5vl} & 54.6 & \cellcolor{secondcolor}{62.0} & \cellcolor{secondcolor}{66.5} & \cellcolor{secondcolor}{60.4} & 37.1 & 70.6 & 32.4 \\

Gemini-2.5-pro-2025-06~\cite{team2023gemini} & \cellcolor{secondcolor}{57.1} & 61.3 & \cellcolor{bestcolor}{\textbf{69.2}} & 55.2 & \cellcolor{secondcolor}{38.5} & \cellcolor{secondcolor}{71.5} & \cellcolor{secondcolor}{48.6} \\

Grok-4-2025-07-09~\cite{grok4_xai_2025} & 47.0 & 50.4 & 60.3 & 51.6 & 26.2 & 61.2 & 37.4 \\

GPT-5-2025-08-07~\cite{openai_gpt5_systemcard} & \cellcolor{bestcolor}{\textbf{61.9}} & \cellcolor{bestcolor}{\textbf{63.1}} & 57.0 & \cellcolor{bestcolor}{\textbf{73.1}} & \cellcolor{bestcolor}{\textbf{49.9}} & \cellcolor{bestcolor}{\textbf{72.2}} & \cellcolor{bestcolor}{\textbf{60.7}} \\

\midrule
\rowcolor{line-blue}\multicolumn{8}{l}{\textbf{Open-source General Models}} \\
Bagel-7B-MoT~\cite{deng2025bagel} & 37.0 & 50.0 & \cellcolor{bestcolor}{\textbf{64.2}} & 22.3 & 4.3 & 63.2 & 10.8 \\

Qwen2.5-VL-3B-Instruct~\cite{Qwen2.5-VL} & 33.1 & 48.4 & 43.6 & 16.1 & 8.5 & 56.3 & 12.4 \\

Qwen2.5-VL-7B-Instruct~\cite{Qwen2.5-VL} & 37.6 & 54.3 & 45.6 & 19.1 & 12.0 & 63.8 & 15.7 \\

Qwen3-VL-2B-Instruct~\cite{Qwen3-VL} & 35.7 & 43.6 & 45.8 & 26.1 & 14.4 & 57.0 & 18.9 \\

Qwen3-VL-8B-Instruct~\cite{Qwen3-VL} & 45.8 & 55.8 & 60.9 & 28.5 & 22.5 & \cellcolor{bestcolor}{\textbf{67.8}} & \cellcolor{secondcolor}{30.0} \\

InternVL3-2B~\cite{zhu2025internvl3} & 30.0 & 46.5 & 40.4 & 13.8 & 5.7 & 50.9 & 9.8 \\

InternVL3-8B~\cite{zhu2025internvl3} & 41.1 & \cellcolor{bestcolor}{\textbf{57.9}} & 52.4 & 30.3 & 9.9 & 61.9 & 26.1 \\

\rowcolor{line-blue}\multicolumn{8}{l}{\textbf{Open-source Spatial Intelligence Models}} \\

MindCube-3B-RawQA-SFT~\cite{mindcube} & 6.3 & 20.8 & 10.6 & -6.9 & -6.8 & 15.9 & -8.9 \\

SpatialLadder-3B~\cite{li2025spatialladder} & 28.0 & 45.0 & 32.0 & 17.4 & 2.9 & 47.3 & 11.3 \\

Spatial-MLLM-4B~\cite{wu2025spatial} & 18.0 & 30.3 & 24.2 & 7.0 & 7.3 & 20.0 & 10.9 \\

SpaceR-7B~\cite{ouyang2025spacer} & 34.3 & 49.6 & 39.0 & 16.4 & 9.1 & 60.2 & 16.0 \\

ViLaSR-7B~\cite{vilasr} & 38.7 & 54.9 & 43.2 & 19.4 & 12.4 & 64.4 & 23.0 \\

VST-3B-SFT~\cite{vst2025} & 35.9 & 47.4 & 35.7 & 27.7 & 16.2 & 57.8 & 18.6 \\

VST-7B-SFT~\cite{vst2025} & 39.7 & 54.7 & 37.1 & 33.1 & 14.0 & 63.9 & 23.2 \\

Cambrian-S-3B~\cite{yang2025cambrian} & 31.0 & 47.1 & 46.2 & 16.4 & 1.4 & 44.0 & 24.6 \\

Cambrian-S-7B~\cite{yang2025cambrian} & 36.1 & 50.0 & 50.3 & 24.1 & 8.7 & 51.2 & 26.6 \\

\rowcolor{line-blue}\multicolumn{8}{l}{\textbf{Ours}} \\

\textbf{\name\textsubscript{Bagel-7B-MoT}} & 42.0 & 47.4 & \cellcolor{secondcolor}{62.9} & 30.6 & 27.8 & 59.2 & 17.4 \\

\textbf{\name\textsubscript{Qwen3-VL-8B}} & \cellcolor{bestcolor}{\textbf{49.6}} & 53.1 & 56.1 & \cellcolor{secondcolor}{39.1} & \cellcolor{bestcolor}{\textbf{40.3}} & \cellcolor{secondcolor}{64.9} & 24.8 \\

\textbf{\name\textsubscript{InternVL3-2B}} & 36.8 & 51.0 & 45.8 & 23.1 & 16.1 & 48.9 & 26.8 \\

\textbf{\name\textsubscript{InternVL3-8B}} & \cellcolor{secondcolor}{47.7} & \cellcolor{secondcolor}{56.2} & 49.2 & \cellcolor{bestcolor}{\textbf{40.6}} & \cellcolor{secondcolor}{35.1} & 57.6 & \cellcolor{bestcolor}{\textbf{40.5}} \\

% \midrule
% \rowcolor{line-blue}\multicolumn{8}{l}{\textbf{Human Evaluation}} \\
% \footnotesize $\Delta$(Best Model,Human) & \colornum{-5.6} & \colornum{-2.9} & \colornum{-14.1} & \colornum{5.7} & \colornum{-36.0} & \colornum{-0.5} & \colornum{6.7} \\

% Human & \textbf{67.5} & \textbf{66.0} & \textbf{83.3} & \textbf{54.7} & \textbf{87.5} & \textbf{73.0} & \textbf{52.5} \\

\bottomrule
\end{tabular}
\end{adjustbox}
\vspace{-2pt}
\caption{\textbf{Evaluation on \textbf{SITE}~\cite{wang2025site}}. All scores are \textbf{CAA}. We follow the \textbf{SITE} paper’s original evaluation protocol, where MCQ are answered by \emph{direct QA}.}
\label{tab:site_oo}
\end{table*}

%% file: tables/supp_tables/blink.tex
\begin{table*}[ht!]
\centering
\setlength{\tabcolsep}{1.2pt}
\resizebox{1.0\linewidth}{!}{
\begin{tabular}{l *{15}{c}}
\toprule
\textbf{Models} & \textbf{Avg.} & \makecell{\textbf{Art}\\\textcolor{sensepurple}{\scriptsize -}} & \makecell{\textbf{Count}\\\textcolor{sensepurple}{\scriptsize -}} & \makecell{\textbf{Forens.}\\\textcolor{sensepurple}{\scriptsize -}} & \makecell{\textbf{FuncCr}\\\textcolor{sensepurple}{\scriptsize -}} & \makecell{\textbf{IQ}\\\textcolor{sensepurple}{\scriptsize CR}} & \makecell{\textbf{Jigsaw}\\\textcolor{sensepurple}{\scriptsize -}} & \makecell{\textbf{MV-R}\\\textcolor{sensepurple}{\scriptsize PT}} & \makecell{\textbf{ObjLoc}\\\textcolor{sensepurple}{\scriptsize -}} & \makecell{\textbf{Depth}\\\textcolor{sensepurple}{\scriptsize SR}} & \makecell{\textbf{Refl.}\\\textcolor{sensepurple}{\scriptsize -}} & \makecell{\textbf{SemCr}\\\textcolor{sensepurple}{\scriptsize -}} & \makecell{\textbf{SpatR}\\\textcolor{sensepurple}{\scriptsize SR}} & \makecell{\textbf{VisCr}\\\textcolor{sensepurple}{\scriptsize PT}} & \makecell{\textbf{VisSm}\\\textcolor{sensepurple}{\scriptsize -}} \\

\midrule
Random Choice & 37.7 & 50.0 & 25.0 & 25.0 & 25.0 & 25.0 & 50.0 & 50.0 & 50.0 & 50.0 & 33.3 & 25.0 & 50.0 & 25.0 & 50.0 \\

\midrule
\rowcolor{line-blue}\textbf{Proprietary Models} & & & & & & & & & & & & & & & \\

Seed-1.6-2025-06-15~\cite{seed2025seed1_5vl} & 65.9 & 80.3 & 74.2 & 62.9 & 53.1 & 28.7 & 59.3 & \cellcolor{secondcolor}{55.6} & 64.8 & 85.5 & 32.1 & 64.8 & 86.7 & 90.7 & 84.4 \\
Gemini-2.5-pro-2025-06~\cite{team2023gemini} & \cellcolor{secondcolor}{73.5} & \cellcolor{secondcolor}{86.0} & \cellcolor{bestcolor}{\textbf{78.0}} & \cellcolor{secondcolor}{90.0} & \cellcolor{secondcolor}{62.0} & \cellcolor{bestcolor}{\textbf{44.0}} & \cellcolor{bestcolor}{\textbf{85.0}} & 51.0 & \cellcolor{secondcolor}{76.0} & \cellcolor{secondcolor}{86.0} & \cellcolor{secondcolor}{34.0} & \cellcolor{secondcolor}{68.0} & \cellcolor{secondcolor}{90.0} & 91.0 & 86.0 \\
Grok-4-2025-07-09~\cite{grok4_xai_2025} & 56.4 & 82.0 & 65.8 & 78.8 & 47.7 & 8.0 & 60.0 & 42.1 & 54.1 & 68.5 & \cellcolor{bestcolor}{\textbf{40.3}} & 38.1 & 83.2 & 47.7 & 85.2 \\
GPT-5-2025-08-07~\cite{openai_gpt5_systemcard} & 68.0 & 85.5 & 69.2 & 79.5 & 60.0 & 32.7 & 70.0 & 40.6 & 61.5 & 81.5 & 32.8 & 66.2 & 88.8 & \cellcolor{secondcolor}{94.2} & \cellcolor{secondcolor}{87.4} \\
Gemini-3-Pro-Preview~\cite{gemini_3_pro_systemcard} & \cellcolor{bestcolor}{\textbf{76.0}} & \cellcolor{bestcolor}{\textbf{86.3}} & \cellcolor{secondcolor}{77.5} & \cellcolor{bestcolor}{\textbf{98.5}} & \cellcolor{bestcolor}{\textbf{66.9}} & \cellcolor{secondcolor}{42.0} & \cellcolor{secondcolor}{77.3} & \cellcolor{bestcolor}{\textbf{57.9}} & \cellcolor{bestcolor}{\textbf{85.2}} & \cellcolor{bestcolor}{\textbf{90.3}} & 32.1 & \cellcolor{bestcolor}{\textbf{72.7}} & \cellcolor{bestcolor}{\textbf{93.0}} & \cellcolor{bestcolor}{\textbf{95.9}} & \cellcolor{bestcolor}{\textbf{88.9}} \\

\midrule
\rowcolor{line-blue}\textbf{Open-source General Models} & & & & & & & & & & & & & & & \\

Bagel-7B-MoT~\cite{deng2025bagel} & 63.6 & \cellcolor{secondcolor}{78.6} & \cellcolor{secondcolor}{75.0} & 37.9 & 40.0 & \cellcolor{secondcolor}{32.0} & \cellcolor{bestcolor}{\textbf{85.3}} & 45.9 & \cellcolor{secondcolor}{72.1} & 89.5 & 31.3 & 46.0 & 82.5 & 82.0 & \cellcolor{bestcolor}{\textbf{92.6}} \\
Qwen2.5-VL-3B-Instruct~\cite{Qwen2.5-VL} & 48.7 & 57.3 & 68.3 & 32.6 & 22.3 & 20.0 & 46.0 & 43.6 & 55.7 & 74.2 & \cellcolor{bestcolor}{\textbf{38.8}} & 36.7 & 82.5 & 45.4 & 65.2 \\
Qwen2.5-VL-7B-Instruct~\cite{Qwen2.5-VL} & 55.9 & 71.8 & 71.7 & 47.7 & 23.9 & 26.7 & 56.0 & 55.6 & 52.5 & 79.0 & \cellcolor{secondcolor}{38.1} & 36.7 & \cellcolor{bestcolor}{\textbf{88.8}} & 52.3 & 88.2 \\
Qwen3-VL-2B-Instruct~\cite{Qwen3-VL} & 53.2 & 65.8 & 62.5 & 34.1 & 33.1 & 1.3 & 54.0 & 55.6 & 59.8 & 71.0 & 30.6 & 59.0 & 81.1 & 63.4 & 77.8 \\
Qwen3-VL-8B-Instruct~\cite{Qwen3-VL} & \cellcolor{bestcolor}{\textbf{66.7}} & 77.8 & 67.5 & \cellcolor{bestcolor}{\textbf{87.9}} & \cellcolor{bestcolor}{\textbf{53.9}} & 12.0 & 67.3 & 54.1 & 69.7 & 88.7 & 32.1 & \cellcolor{secondcolor}{59.7} & \cellcolor{secondcolor}{87.4} & 89.5 & 88.2 \\
InternVL3-2B~\cite{zhu2025internvl3} & 50.8 & 56.4 & 73.3 & 18.9 & 26.9 & 24.0 & 75.3 & 59.4 & 62.3 & 71.0 & 34.3 & 23.7 & 79.0 & 34.3 & 80.0 \\
InternVL3-8B~\cite{zhu2025internvl3} & 53.5 & 67.5 & 71.7 & 34.9 & 28.5 & 23.3 & 78.0 & 44.4 & 53.3 & 69.3 & 34.3 & 28.1 & 86.7 & 44.2 & 89.6 \\

\rowcolor{line-blue}\textbf{Open-source Spatial Intelligence Models} & & & & & & & & & & & & & & & \\

MindCube-3B-RawQA-SFT~\cite{mindcube} & 35.1 & 53.0 & 36.7 & 26.5 & 6.9 & 26.0 & 47.3 & 55.6 & 56.6 & 41.1 & 31.3 & 0.0 & 67.1 & 3.5 & 51.1 \\
SpatialLadder-3B~\cite{li2025spatialladder} & 43.0 & 53.9 & 53.3 & 22.7 & 23.9 & 26.7 & 49.3 & 44.4 & 50.0 & 58.1 & 31.3 & 32.4 & 75.5 & 29.1 & 57.8 \\
Spatial-MLLM-4B~\cite{wu2025spatial} & 40.5 & 47.0 & 40.8 & 22.0 & 20.8 & 23.3 & 54.0 & 43.6 & 54.9 & 63.7 & 29.1 & 34.5 & 54.5 & 35.5 & 46.7 \\
SpaceR-7B~\cite{ouyang2025spacer} & 49.6 & 76.1 & 65.0 & 32.6 & 20.0 & 4.0 & 60.0 & 43.6 & 52.5 & 66.1 & 31.3 & 35.2 & 80.4 & 51.2 & 83.7 \\
ViLaSR-7B~\cite{vilasr} & 51.4 & 68.4 & 70.0 & 55.3 & 26.1 & 18.0 & 63.3 & 48.1 & 40.2 & 74.2 & 12.7 & 19.4 & 86.7 & 55.2 & 85.9 \\
VST-3B-SFT~\cite{vst2025} & 58.8 & 65.8 & 69.2 & 53.0 & 23.1 & \cellcolor{bestcolor}{\textbf{33.3}} & 68.0 & 51.1 & 54.1 & 86.3 & 35.8 & 30.9 & 74.1 & 89.5 & 84.4 \\
VST-7B-SFT~\cite{vst2025} & 61.9 & 67.5 & 70.8 & 55.3 & 26.9 & 23.3 & \cellcolor{secondcolor}{80.0} & \cellcolor{secondcolor}{63.2} & 54.1 & \cellcolor{secondcolor}{91.9} & 34.3 & 33.1 & 81.1 & 94.2 & 85.9 \\
Cambrian-S-3B~\cite{yang2025cambrian} & 37.7 & 32.5 & 63.3 & 15.2 & 20.0 & 27.3 & 50.0 & 6.8 & 50.8 & 54.0 & 31.3 & 31.6 & 66.4 & 34.3 & 46.7 \\
Cambrian-S-7B~\cite{yang2025cambrian} & 37.9 & 42.7 & 63.3 & 0.0 & 20.8 & 24.0 & 37.3 & 44.4 & 51.6 & 69.3 & 28.4 & 28.8 & 77.6 & 28.5 & 21.5 \\

\rowcolor{line-blue}\textbf{Ours} & & & & & & & & & & & & & & & \\

\textbf{\name\textsubscript{Bagel-7B-MoT}} & \cellcolor{secondcolor}{65.4} & \cellcolor{bestcolor}{\textbf{81.2}} & 71.7 & 33.3 & 49.2 & 28.7 & 77.3 & 44.4 & \cellcolor{bestcolor}{\textbf{77.0}} & 87.1 & 33.6 & \cellcolor{bestcolor}{\textbf{66.9}} & 76.2 & \cellcolor{secondcolor}{94.8} & \cellcolor{secondcolor}{91.8} \\
\textbf{\name\textsubscript{Qwen3-VL-8B}} & 61.9 & 74.4 & 61.7 & \cellcolor{secondcolor}{69.7} & \cellcolor{secondcolor}{50.8} & 13.3 & 64.0 & 32.3 & 65.6 & 88.7 & 30.6 & 52.5 & 79.7 & 93.6 & 88.2 \\
\textbf{\name\textsubscript{InternVL3-2B}} & 52.4 & 61.5 & 65.8 & 12.9 & 35.4 & 21.3 & 63.3 & 6.0 & 58.2 & 80.7 & 35.1 & 36.0 & 74.1 & 93.0 & 83.7 \\
\textbf{\name\textsubscript{InternVL3-8B}} & 63.9 & 68.4 & \cellcolor{bestcolor}{\textbf{76.7}} & 32.6 & 33.9 & 21.3 & \cellcolor{secondcolor}{80.0} & \cellcolor{bestcolor}{\textbf{90.2}} & 54.9 & \cellcolor{bestcolor}{\textbf{93.5}} & 32.8 & 46.0 & 86.0 & \cellcolor{bestcolor}{\textbf{95.9}} & 77.8 \\

\bottomrule
\end{tabular}
}
\vspace{-2pt}
\caption{
\textbf{Evaluation on \textbf{BLINK}~\cite{fu2024blink}}.
All scores are \textbf{Acc}.
}
\label{tab:blink_oo}
\end{table*}

%% file: tables/supp_tables/3dsr.tex
% =================== 3DSRBench (auto-generated) ===================
\begin{table*}[ht!]
\centering\small
\setlength{\tabcolsep}{1.2pt}
\begin{adjustbox}{max width=\linewidth}
\begin{tabular}{l*{13}{c}}
\toprule
\multirow{2}{*}{\textbf{Models}} &
\multirow{2}{*}{\textbf{Overall}} &
\textbf{Height} &
\multicolumn{3}{c}{\textbf{Location}} &
\multicolumn{5}{c}{\textbf{Multi-Object}} &
\multicolumn{3}{c}{\textbf{Orientation}} \\
\cmidrule(lr){3-3}\cmidrule(lr){4-6}\cmidrule(lr){7-11}\cmidrule(lr){12-14}
& \multicolumn{1}{c}{} &
\makecell{\textbf{Higher}\\\textcolor{sensepurple}{\scriptsize SR}} &
\makecell{\textbf{Above}\\\textcolor{sensepurple}{\scriptsize SR}} &
\makecell{\textbf{Closer}\\\textcolor{sensepurple}{\scriptsize SR}} &
\makecell{\textbf{Next To}\\\textcolor{sensepurple}{\scriptsize SR}} &
\makecell{\textbf{Closer}\\\textcolor{sensepurple}{\scriptsize PT,SR}} &
\makecell{\textbf{Facing}\\\textcolor{sensepurple}{\scriptsize PT,SR}} &
\makecell{\textbf{Parall.}\\\textcolor{sensepurple}{\scriptsize PT,SR}} &
\makecell{\textbf{Same D.}\\\textcolor{sensepurple}{\scriptsize PT,SR}} &
\makecell{\textbf{ViewPt}\\\textcolor{sensepurple}{\scriptsize PT,SR}} &
\makecell{\textbf{Front}\\\textcolor{sensepurple}{\scriptsize PT}} &
\makecell{\textbf{Left}\\\textcolor{sensepurple}{\scriptsize PT}} &
\makecell{\textbf{ViewPt}\\\textcolor{sensepurple}{\scriptsize PT}} \\

\midrule
Random Choice & 45.8 & 50.0 & 50.0 & 50.0 & 50.0 & 50.0 & 50.0 & 50.0 & 50.0 & 25.0 & 50.0 & 50.0 & 25.0 \\

\midrule
\rowcolor{line-blue}\multicolumn{14}{l}{\textbf{Proprietary Models}} \\

Seed-1.6-2025-06-15~\cite{seed2025seed1_5vl} & 56.9 & 67.7 & 64.3 & 74.9 & 67.8 & \cellcolor{secondcolor}{67.1} & \cellcolor{secondcolor}{64.5} & 56.6 & \cellcolor{secondcolor}{48.8} & 17.5 & 61.9 & 32.7 & 23.6 \\
Gemini-2.5-pro-2025-06~\cite{team2023gemini} & 59.3 & \cellcolor{secondcolor}{68.1} & 66.3 & 78.8 & \cellcolor{bestcolor}{\textbf{81.1}} & 66.9 & 63.3 & \cellcolor{secondcolor}{56.9} & 45.4 & 15.2 & 67.2 & 31.2 & \cellcolor{secondcolor}{37.3} \\
Grok-4-2025-07-09~\cite{grok4_xai_2025} & 54.9 & 63.5 & \cellcolor{secondcolor}{68.6} & 71.5 & 66.1 & 52.6 & 61.6 & 46.3 & 46.5 & 21.9 & 62.8 & 36.4 & 23.0 \\
GPT-5-2025-08-07~\cite{openai_gpt5_systemcard} & \cellcolor{secondcolor}{60.3} & 66.1 & 67.8 & \cellcolor{bestcolor}{\textbf{82.7}} & \cellcolor{secondcolor}{72.9} & 64.9 & 63.9 & 54.6 & 45.1 & \cellcolor{secondcolor}{22.2} & \cellcolor{secondcolor}{74.4} & \cellcolor{secondcolor}{40.1} & 33.8 \\
Gemini-3-Pro-Preview~\cite{gemini_3_pro_systemcard} & \cellcolor{bestcolor}{\textbf{68.9}} & \cellcolor{bestcolor}{\textbf{69.0}} & \cellcolor{bestcolor}{\textbf{73.1}} & \cellcolor{secondcolor}{80.1} & 72.9 & \cellcolor{bestcolor}{\textbf{74.9}} & \cellcolor{bestcolor}{\textbf{73.4}} & \cellcolor{bestcolor}{\textbf{66.1}} & \cellcolor{bestcolor}{\textbf{63.4}} & \cellcolor{bestcolor}{\textbf{35.3}} & \cellcolor{bestcolor}{\textbf{82.3}} & \cellcolor{bestcolor}{\textbf{72.8}} & \cellcolor{bestcolor}{\textbf{48.4}} \\

\midrule
\rowcolor{line-blue}\multicolumn{14}{l}{\textbf{Open-source General Models}} \\

Bagel-7B-MoT~\cite{deng2025bagel} & 50.2 & 47.0 & \cellcolor{secondcolor}{64.0} & 75.5 & 49.6 & 56.6 & 58.1 & \cellcolor{secondcolor}{49.0} & 49.1 & 5.0 & \cellcolor{secondcolor}{58.1} & 29.2 & 26.5 \\
Qwen2.5-VL-3B-Instruct~\cite{Qwen2.5-VL} & 43.5 & 45.8 & 54.5 & 69.6 & 47.2 & 44.0 & 44.5 & 23.6 & 39.8 & 14.6 & 43.9 & \cellcolor{secondcolor}{34.4} & 20.4 \\
Qwen2.5-VL-7B-Instruct~\cite{Qwen2.5-VL} & 47.5 & 44.9 & 50.9 & 74.3 & 56.0 & 48.3 & 51.2 & 37.8 & 47.7 & 9.9 & 56.1 & \cellcolor{secondcolor}{34.4} & 31.8 \\
Qwen3-VL-2B-Instruct~\cite{Qwen3-VL} & 47.5 & 50.0 & 57.8 & 74.9 & 57.8 & 56.6 & 50.0 & 48.7 & 45.4 & 4.4 & 47.1 & 26.1 & 11.1 \\
Qwen3-VL-8B-Instruct~\cite{Qwen3-VL} & \cellcolor{secondcolor}{53.9} & 47.8 & 62.6 & \cellcolor{bestcolor}{\textbf{87.6}} & 57.8 & 62.0 & \cellcolor{secondcolor}{65.6} & 46.0 & 47.1 & 13.1 & 56.7 & 30.4 & \cellcolor{bestcolor}{\textbf{35.0}} \\
InternVL3-2B~\cite{zhu2025internvl3} & 47.7 & 50.0 & 54.2 & 61.8 & 55.8 & 53.4 & 45.4 & \cellcolor{bestcolor}{\textbf{50.4}} & \cellcolor{secondcolor}{52.6} & 12.5 & 53.2 & 34.1 & 26.2 \\
InternVL3-8B~\cite{zhu2025internvl3} & 44.2 & 49.4 & 58.2 & 76.1 & 59.3 & 46.3 & 37.6 & 17.4 & 34.0 & 13.7 & 43.6 & 21.8 & 23.3 \\

\rowcolor{line-blue}\multicolumn{14}{l}{\textbf{Open-source Spatial Intelligence Models}} \\

MindCube-3B-RawQA-SFT~\cite{mindcube} & 2.8 & 0.0 & 12.3 & 0.0 & 5.3 & 6.9 & 4.3 & 0.0 & 0.0 & 0.0 & 0.0 & 0.0 & 0.0 \\
SpatialLadder-3B~\cite{li2025spatialladder} & 42.8 & 45.1 & 39.9 & 65.3 & 47.5 & 46.9 & 59.0 & 45.1 & 48.8 & 8.8 & 39.8 & \cellcolor{secondcolor}{34.4} & 11.1 \\
Spatial-MLLM-4B~\cite{wu2025spatial} & 36.2 & 42.5 & 41.3 & 45.7 & 46.0 & 40.6 & 61.0 & 36.9 & 47.7 & 5.2 & 41.3 & 3.7 & 1.8 \\
SpaceR-7B~\cite{ouyang2025spacer} & 40.5 & 36.4 & 47.2 & 63.1 & 44.5 & 42.9 & 52.3 & 24.2 & 48.0 & 3.8 & 54.4 & 27.2 & 16.3 \\
ViLaSR-7B~\cite{vilasr} & 46.6 & 38.3 & \cellcolor{bestcolor}{\textbf{65.8}} & 66.4 & 50.4 & 45.4 & 59.5 & 47.8 & 45.4 & 3.5 & 52.3 & 33.5 & 20.1 \\
VST-3B-SFT~\cite{vst2025} & 48.7 & 47.7 & 55.4 & \cellcolor{secondcolor}{86.3} & 46.3 & 62.0 & 57.8 & 39.2 & 39.2 & 16.6 & 41.0 & 32.1 & 18.4 \\
VST-7B-SFT~\cite{vst2025} & 53.1 & \cellcolor{bestcolor}{\textbf{55.8}} & 53.3 & 84.5 & 55.5 & \cellcolor{secondcolor}{72.9} & 64.2 & 43.7 & 45.6 & 17.2 & 53.8 & 32.7 & 24.2 \\
Cambrian-S-3B~\cite{yang2025cambrian} & 41.4 & 33.5 & 46.2 & 52.8 & \cellcolor{secondcolor}{61.6} & 53.4 & 55.8 & 44.0 & 50.0 & 10.5 & 39.0 & 26.6 & 15.2 \\
Cambrian-S-7B~\cite{yang2025cambrian} & 45.0 & 43.3 & 46.1 & 58.3 & \cellcolor{bestcolor}{\textbf{64.9}} & 61.4 & 48.3 & 35.4 & 48.8 & 19.5 & 50.3 & 25.2 & 26.8 \\

\rowcolor{line-blue}\multicolumn{14}{l}{\textbf{Ours}} \\

\textbf{\name\textsubscript{Bagel-7B-MoT}} & 42.4 & 33.5 & 47.4 & 67.7 & 47.8 & 43.4 & 53.5 & 43.4 & 46.5 & 2.9 & 47.7 & 28.9 & 23.6 \\
\textbf{\name\textsubscript{Qwen3-VL-8B}} & 53.2 & \cellcolor{secondcolor}{52.2} & 55.9 & 85.0 & 58.4 & 66.6 & 61.9 & 41.0 & 49.1 & \cellcolor{secondcolor}{20.7} & 49.7 & 34.1 & 30.9 \\
\textbf{\name\textsubscript{InternVL3-2B}} & 50.5 & 50.0 & 51.6 & 66.8 & \cellcolor{bestcolor}{\textbf{64.9}} & 66.9 & \cellcolor{bestcolor}{\textbf{66.2}} & \cellcolor{secondcolor}{49.0} & 50.0 & \cellcolor{bestcolor}{\textbf{21.3}} & 45.9 & 28.9 & 27.7 \\
\textbf{\name\textsubscript{InternVL3-8B}} & \cellcolor{bestcolor}{\textbf{55.5}} & 49.1 & 51.5 & 83.8 & 59.3 & \cellcolor{bestcolor}{\textbf{75.7}} & 59.2 & 46.9 & \cellcolor{bestcolor}{\textbf{53.2}} & 16.9 & \cellcolor{bestcolor}{\textbf{60.2}} & \cellcolor{bestcolor}{\textbf{60.5}} & \cellcolor{secondcolor}{32.6} \\

\bottomrule
\end{tabular}
\end{adjustbox}
\vspace{-2pt}
\caption{\textbf{Evaluation on \textbf{3DSRBench}~\cite{ma20243dsrbench}.} All scores are \textbf{Acc} (circ\_eval setting).}
\label{tab:3dsrbench_oo}
\end{table*}

%% file: tables/supp_tables/embspatial.tex
% =================== EmbSpatialBench (auto-generated) ===================
\begin{table*}[ht!]
\centering\small
\setlength{\tabcolsep}{1.2pt}
\begin{adjustbox}{max width=\linewidth}
\begin{tabular}{l*{4}{c}}
\toprule
\textbf{Models} &
\textbf{Overall} &
\makecell{\textbf{MP3D}\\\textcolor{sensepurple}{\scriptsize SR}} &
\makecell{\textbf{AI2-THOR}\\\textcolor{sensepurple}{\scriptsize SR}} &
\makecell{\textbf{ScanNet}\\\textcolor{sensepurple}{\scriptsize SR}} \\

\midrule
Random Choice & 25.0 & 25.0 & 25.0 & 25.0 \\

\midrule
\rowcolor{line-blue}\multicolumn{5}{l}{\textbf{Proprietary Models}} \\

Seed-1.6-2025-06-15~\cite{seed2025seed1_5vl} & 75.4 & 74.3 & 75.1 & 76.7 \\
Gemini-2.5-pro-2025-06~\cite{team2023gemini} & 78.8 & 77.8 & 78.0 & 80.8 \\
Grok-4-2025-07-09~\cite{grok4_xai_2025} & 75.5 & 75.3 & 75.4 & 75.8 \\
GPT-5-2025-08-07~\cite{openai_gpt5_systemcard} & \cellcolor{secondcolor}{81.6} & \cellcolor{secondcolor}{81.3} & \cellcolor{secondcolor}{80.5} & \cellcolor{secondcolor}{83.0} \\
Gemini-3-Pro-Preview~\cite{gemini_3_pro_systemcard} & \cellcolor{bestcolor}{\textbf{84.3}} & \cellcolor{bestcolor}{\textbf{83.3}} & \cellcolor{bestcolor}{\textbf{84.5}} & \cellcolor{bestcolor}{\textbf{85.0}} \\

\midrule
\rowcolor{line-blue}\multicolumn{5}{l}{\textbf{Open-source General Models}} \\

Bagel-7B-MoT~\cite{deng2025bagel} & 73.1 & 72.2 & 71.4 & 75.8 \\
Qwen2.5-VL-3B-Instruct~\cite{Qwen2.5-VL} & 62.3 & 63.2 & 56.7 & 67.1 \\
Qwen2.5-VL-7B-Instruct~\cite{Qwen2.5-VL} & 71.8 & 74.6 & 70.5 & 70.4 \\
Qwen3-VL-2B-Instruct~\cite{Qwen3-VL} & 70.1 & 72.2 & 67.5 & 70.8 \\
Qwen3-VL-8B-Instruct~\cite{Qwen3-VL} & \cellcolor{bestcolor}{\textbf{77.7}} & \cellcolor{bestcolor}{\textbf{79.6}} & \cellcolor{bestcolor}{\textbf{75.1}} & \cellcolor{bestcolor}{\textbf{78.6}} \\
InternVL3-2B~\cite{zhu2025internvl3} & 60.1 & 60.8 & 55.8 & 63.9 \\
InternVL3-8B~\cite{zhu2025internvl3} & \cellcolor{secondcolor}{76.3} & \cellcolor{secondcolor}{76.3} & \cellcolor{secondcolor}{74.2} & \cellcolor{secondcolor}{78.5} \\

\rowcolor{line-blue}\multicolumn{5}{l}{\textbf{Open-source Spatial Intelligence Models}} \\

MindCube-3B-RawQA-SFT~\cite{mindcube} & 37.0 & 36.7 & 33.0 & 41.3 \\
SpatialLadder-3B~\cite{li2025spatialladder} & 58.2 & 60.0 & 54.1 & 60.8 \\
Spatial-MLLM-4B~\cite{wu2025spatial} & 50.0 & 51.5 & 46.9 & 51.5 \\
SpaceR-7B~\cite{ouyang2025spacer} & 66.9 & 67.8 & 66.7 & 66.3 \\
ViLaSR-7B~\cite{vilasr} & 67.3 & 69.5 & 67.9 & 64.5 \\
VST-3B-SFT~\cite{vst2025} & 69.0 & 71.7 & 63.3 & 72.2 \\
VST-7B-SFT~\cite{vst2025} & 73.7 & 76.2 & 71.1 & 74.0 \\
Cambrian-S-3B~\cite{yang2025cambrian} & 63.5 & 63.5 & 60.2 & 67.0 \\
Cambrian-S-7B~\cite{yang2025cambrian} & 72.8 & 71.3 & 72.9 & 74.3 \\

\rowcolor{line-blue}\multicolumn{5}{l}{\textbf{Ours}} \\

\textbf{\name\textsubscript{Bagel-7B-MoT}} & 69.0 & 68.2 & 63.8 & 75.1 \\
\textbf{\name\textsubscript{Qwen3-VL-8B}} & 72.5 & 75.8 & 66.4 & 75.4 \\
\textbf{\name\textsubscript{InternVL3-2B}} & 62.8 & 64.2 & 56.7 & 67.6 \\
\textbf{\name\textsubscript{InternVL3-8B}} & 72.0 & 71.0 & 70.5 & 75.1 \\

\bottomrule
\end{tabular}
\end{adjustbox}
\vspace{-2pt}
\caption{\textbf{Evaluation on \textbf{EmbSpatialBench}~\cite{du2024embspatial}.} All scores are \textbf{Acc}.}
\label{tab:embspatial_oo}
\end{table*}

%% file: figures/cot_case1.tex
\begin{figure*}[t]
    \centering
    \setlength{\belowcaptionskip}{-4pt}
    \includegraphics[width=\linewidth]{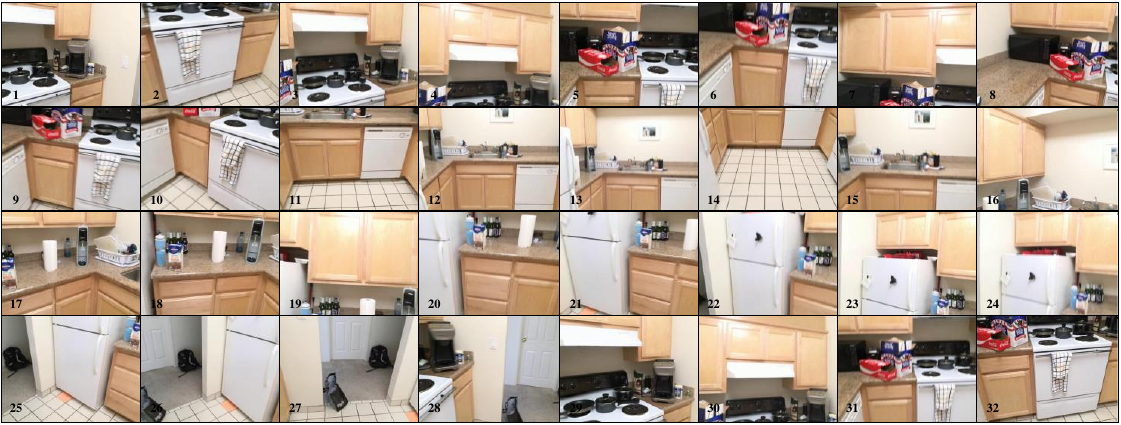}
    \caption{
    Successful case of trained \textit{CoT-\name-CGMap} output on VSI-Bench~\cite{vsi} Rel. Dir subset. The model maintains consistent and accurate reasoning through images and concludes to a \cotsuchl{correct} final answer. 
    }
    \label{fig:cot_case1}
\end{figure*}

%% file: figures/cot_case1_text_part1.tex
\textbf{Question:}

If I am standing by the refrigerator and facing the backpack, is the microwave to my front-left, front-right, back-left, or back-right?

The directions refer to the quadrants of a Cartesian plane (if I am standing at the origin and facing along the positive y-axis).

Options: A. back-left, B. back-right, C. front-left, D. front-right

\textbf{GT:} A

\textbf{Prediction(\textit{CoT-\name-CGMap}):} 

<think>

I will reason through this question step by step.

For one object in one image coordination system, its location is described as "{category\_id} at bbox [x1, y1, x2, y2], at {direction}, {distance from camera} and {x, y}".

x-aixs is the right side of the camera view perspective. y-axis is the forwad perspective.

The unit of distance is meters, and the unit of angle is degrees.

For describing camera motion, the movement from image `a` to image `b` is defined relative to the coordinate system of image `a`.

~

Step 1: **Clarify the queried objects in the related image**

From the question, I can infer that the queried objects are the refrigerator, the backpack and the microwave.

After looking over the provided images, I can get that the \cotsuchl{refrigerator\_6 shows in image 21, specifically at bbox [0.09, 0.0, 0.67, 0.98], at 5 degrees(CCW), dist 1.5, x -0.1, y 1.4}; the backpack\_19 shows in image 27, specifically at bbox [0.54, 0.17, 0.76, 0.45], at 10 degrees(CW), dist 2.6, x 0.4, y 2.3; the microwave\_28 shows in image 7, specifically at bbox [0.03, 0.72, 0.37, 1.0], at 15 degrees(CCW), dist 1.6, x -0.4, y 1.5.

Consequently, to solve this question, image [7, 21, 27] need to be paid more attentions to.
In this step, I have also assigned an id to each queried object.

I will keep this custom to distinguish objects that belong to the same category but are not the same instance.

~

Step 2: **Establish connections among the relevant images**

To analyze the relationships among the images identified in Step 1, I will select some images to connect them.

Two images are connected only if they share corresponding objects.

After looking over all these images, I derive the following chain that illustrates how the queried images are connected:

From image 1 to image 5 via image 2; image 3; image 4, the corresponding object is: stove\_3. Based on the object appearance change, I can infer that in this phase, the camera translates right and front, \cotsuchl{rotates to left and looks down}.

From image 5 to image 7 via image 6, the corresponding objects are: microwave\_28, box\_26. Based on the object appearance change, I can infer that in this phase, the camera does not move significantly, \cotsuchl{rotates to left slightly and looks up}.

%% file: figures/cot_case1_text_part2.tex
From image 7 to image 13 via image 8; image 11, the corresponding object is: \cotfailhl{sink\_0}. Based on the object appearance change, I can infer that in this phase, the camera translates right and back, \cotsuchl{rotates to left and looks down}.

From image 13 to image 21 via image 15; image 17, the corresponding object is: \cotfailhl{(dish rack)\_5}. Based on the object appearance change, I can infer that in this phase, the camera translates right and front, \cotsuchl{rotates to left and looks down}.

From image 21 to image 27 via image 25, the corresponding object is: refrigerator\_6. Based on the object appearance change, I can infer that in this phase, \cotsuchl{the camera translates left, rotates to left and looks down}.

Based on these analyse, I can infer that image 7 and image 21 can be related by image order \cotsuchl{[7, 8, 11, 13, 15, 17, 21]}. Then image 27 can be further connected by image order \cotsuchl{[21, 25, 27]}.

~

Step 3: **Construct Scene Layout**

The overall scene layout is constructed in the coordinate system of image 7 as follows.

How the main images are connected has been described in Step 1. The I will analyse the layout in detail.

** From image7 to image13 **

Image 7 contains objects: microwave\_28 at bbox [0.03, 0.72, 0.37, 1.0], at 15 degrees(CCW), dist 1.6, x -0.4, y 1.5; sink\_0 at bbox [0.43, 0.52, 0.56, 0.59], at 0 degrees(CW), dist 1.9, x 0.1, y 1.8; (dish rack)\_5 at bbox [0.03, 0.26, 0.33, 0.47], at 15 degrees(CCW), dist 1.7, x -0.5, y 1.6.

Image 13 contains objects: sink\_0 at bbox [0.41, 0.48, 0.64, 0.6], at 0 degrees(CW), dist 2.0, x 0.1, y 1.9.

The correspond objects between image 7 and image 13 is sink\_0.

According to the objects motion changes, \cotsuchl{from image7 to image13, the camera motion is: translation (x 0.2, y -0.1), rotation (left 25 degrees, down 10 degrees).}

** From image13 to image21 **

Image 13 contains objects: (dish rack)\_5 at bbox [0.03, 0.26, 0.33, 0.47], at 15 degrees(CCW), dist 1.7, x -0.5, y 1.6.

Image 21 contains objects: (dish rack)\_5 at bbox [0.85, 0.01, 1.0, 0.37], at 35 degrees(CW), dist 1.5, x 0.8, y 1.2.

The correspond objects between image 13 and image 21 is (dish rack)\_5.

According to the objects motion changes, from image13 to image21, the camera motion is: translation (x 0.4, y 0.4), \cotsuchl{rotation (left 70 degrees, down 10 degrees)}.

As mentioned above, from image7 to image13, the camera motion is: translation (x 0.2, y -0.1), rotation (left 25 degrees, down 10 degrees).

Consequently, by accumulating the transformations, from image7 to image21, the camera motion is: translation (x 0.4, y 0.4), rotation (left 95 degrees, down 20 degrees).

** From image21 to image27 **

Image 21 contains objects: refrigerator\_6 at bbox [0.34, 0.0, 0.79, 0.87], at 5 degrees(CW), dist 1.6, x 0.1, y 1.5.

Image 27 contains objects: backpack\_19 at bbox [0.54, 0.17, 0.76, 0.45], at 10 degrees(CW), dist 2.6, x 0.4, y 2.3; refrigerator\_6 at bbox [0.38, 0.0, 0.79, 0.88], at 5 degrees(CW), dist 1.6, x 0.1, y 1.4.

The correspond objects between image 21 and image 27 is refrigerator\_6.

According to the objects motion changes, from image21 to image27, the camera motion is: translation (x -0.2, y 0.0), rotation (left 50 degrees, down 15 degrees).

As mentioned above, from image7 to image21, the camera motion is: translation (x 0.4, y 0.4), rotation (left 95 degrees, down 20 degrees).

Consequently, \cotsuchl{by accumulating the transformations, from image7 to image27, the camera motion is: translation (x 0.4, y 0.2), rotation (left 145 degrees, down 35 degrees)}.

In this phase, image 27 has introduced some new objects in its own coordinate system: backpack\_19 at bbox [0.54, 0.17, 0.76, 0.45]; refrigerator\_6 at bbox [0.38, 0.0, 0.79, 0.88].

Transfer these objects from image 27 into image 7 coordination system: \cotsuchl{backpack\_19 at 140 degrees(CCW), dist 2.5, x -1.2, y -1.5; refrigerator\_6 at 70 degrees(CCW), dist 1.4, x -1.0, y 0.4};

~

Step 4: **Infer object spatial relation and answer the question**

After localizing the objects from questions in the scene layout from step2, we can know that \cotsuchl{"In image 7 coordination system: refrigerator\_6 at 70 degrees(CCW), dist 1.4, x -1.0, y 0.4; backpack\_19 at 140 degrees(CCW), dist 2.5, x -1.2, y -1.5; microwave\_28 at 15 degrees(CCW), dist 1.6, x -0.4, y 1.5".}

Considering the **view perspective** mentioned in the question, the right answer this question is the right answer is \cotsuchl{`A. back-left`}

</think>

<answer>A</answer>

%% file: figures/cot_case2.tex
\begin{figure*}[t]
    \centering
    \setlength{\belowcaptionskip}{-4pt}
    \includegraphics[width=\linewidth]{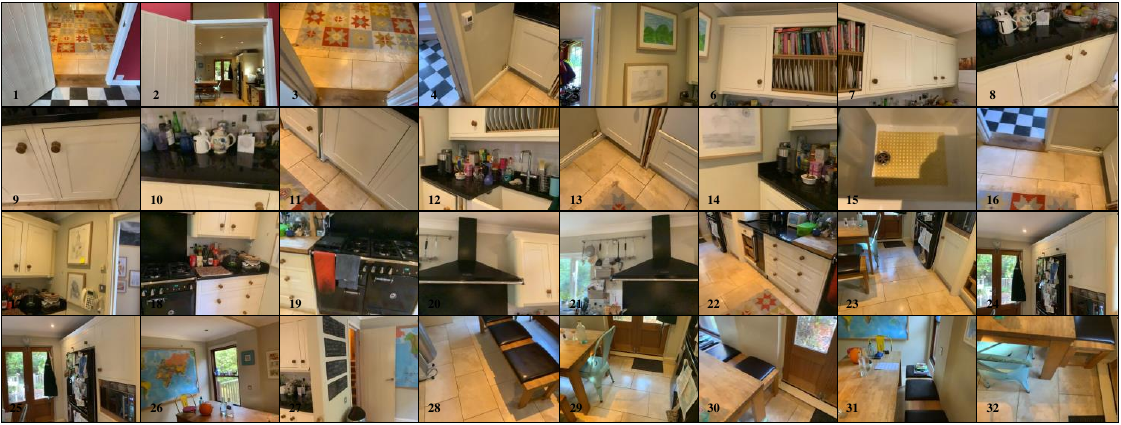}
    \caption{
     Failure case of trained \textit{CoT-\name-CGMap} output on VSI-Bench~\cite{vsi} Rel. Dir subset. Multiple local inconsistencies accumulate through the reasoning trajectory and ultimately lead to an \cotfailhl{incorrect} answer.
    }
    \label{fig:cot_case2}
\end{figure*}

%% file: figures/cot_case2_text_part1.tex
\textbf{Question:} 

If I am standing by the stove and facing the table, is the refrigerator to my front-left, front-right, back-left, or back-right?

The directions refer to the quadrants of a Cartesian plane (if I am standing at the origin and facing along the positive y-axis).

Options: A. front-right, B. front-left, C. back-right, D. back-left

\textbf{GT:} A

\textbf{Prediction(\textit{CoT-\name-CGMap}):} 

<think>

I will reason through this question step by step.

For one object in one image coordination system, its location is described as "{category\_id} at bbox [x1, y1, x2, y2], at {direction}, {distance from camera} and {x, y}".

x-aixs is the right side of the camera view perspective. y-axis is the forwad perspective.

The unit of distance is meters, and the unit of angle is degrees.

For describing camera motion, the movement from image `a` to image `b` is defined relative to the coordinate system of image `a`.

~

Step 1: **Clarify the queried objects in the related image**

From the question, I can infer that the queried objects are the stove, the table and the refrigerator.

After looking over the provided images, I can get that the stove\_18 shows in image 19, specifically at bbox [0.21, 0.01, 1.0, 1.0], at 5 degrees(CW), dist 1.6, x 0.1, y 1.4; \cotfailhl{the table\_25 shows in image 21, specifically at bbox [0.02, 0.82, 0.2, 1.0], at 25 degrees(CCW), dist 2.6, x -0.9, y 2.1;} the refrigerator\_10 shows in image 25, specifically at bbox [0.41, 0.26, 0.6, 0.98], at 5 degrees(CW), dist 2.4, x 0.1, y 2.3.

Consequently, to solve this question, image [19, 21, 25] need to be paid more attentions to.

In this step, I have also assigned an id to each queried object.

I will keep this custom to distinguish objects that belong to the same category but are not the same instance.

~

Step 2: **Establish connections among the relevant images**

To analyze the relationships among the images identified in Step 1, I will select some images to connect them.

Two images are connected only if they share corresponding objects.

After looking over all these images, I derive the following chain that illustrates how the queried images are connected:

From image 1 to image 2 directly, the corresponding object is: doorframe\_24. Based on the object appearance change, I can infer that in this phase, the camera translates right, rotates to right slightly and looks up.

From image 1 to image 4 directly, the corresponding object is: door\_21. Based on the object appearance change, I can infer that in this phase, the camera translates left and front, rotates to left slightly and looks up.

%% file: figures/cot_case2_text_part2.tex
From image 4 to image 16 directly, the objects are mostly not fully visible. But Based on the object appearance change, I can infer that in this phase, the camera translates left, rotates to left and looks down.

From image 16 to image 19 directly, the corresponding objects are: towel\_16, stove\_18. Based on the object appearance change, I can infer that in this phase, \cotsuchl{the camera translates right and front, rotates to back-left and looks up}.

From image 2 to image 21 via image 19, the corresponding object is: (kitchen cabinets)\_4. Based on the object appearance change, I can infer that in this phase, the camera translates right and front, \cotfailhl{rotates to left and looks up}.

From image 21 to image 25 via image 24, \cotfailhl{the corresponding object is: doors\_0}. Based on the object appearance change, I can infer that in this phase, the camera translates left and rears, \cotfailhl{rotates to right and looks up}.

Based on these analyse, I can infer that image 19 and image 21 can be related by image order \cotfailhl{[19, 16, 4, 1, 2, 19]}. Then image 25 can be further connected by image order [19, 21, 24, 25].

~

Step 3: **Construct Scene Layout**

The overall scene layout is constructed in the coordinate system of image 19 as follows.

How the main images are connected has been described in Step 1. The I will analyse the layout in detail.

** From image19 to image4 **

Image 19 contains objects: stove\_18 at bbox [0.21, 0.01, 1.0, 1.0], at 5 degrees(CW), dist 1.6, x 0.1, y 1.4; towel\_16 at bbox [0.24, 0.35, 0.41, 0.78], at 10 degrees(CCW), dist 1.4, x -0.2, y 1.3; towel\_15 at bbox [0.31, 0.32, 0.49, 0.77], at 5 degrees(CCW), dist 1.4, x -0.1, y 1.3.

Image 4 contains objects: \cotfailhl{towel\_16 at bbox [0.9, 0.65, 1.0, 0.99], at 40 degrees(CW), dist 1.4, x 0.7, y 0.9; towel\_15 at bbox [0.89, 0.34, 1.0, 0.83], at 35 degrees(CW), dist 1.4, x 0.7, y 1.0.}

The correspond objects between image 19 and image 4 are towel\_16, towel\_15.

According to the objects motion changes, from image19 to image4, the camera motion is: translation (x -0.1, y 0.1), \cotsuchl{rotation (right 50 degrees, down 30 degrees)}.

** From image4 to image16 **

According to the objects motion changes, from image4 to image16, the camera motion is: translation (x -0.2, y 0.0), rotation (left 30 degrees, down 15 degrees).

As mentioned above, from image19 to image4, the camera motion is: translation (x -0.1, y 0.1), rotation (right 50 degrees, down 30 degrees).

Consequently, by accumulating the transformations, from image19 to image16, the camera motion is: translation (x -0.2, y 0.3), rotation (\cotfailhl{right 20 degrees}, down 45 degrees).

** From image16 to image19 **

Image 16 contains objects: stove\_18 at bbox [0.02, 0.0, 0.56, 0.51], at 20 degrees(CCW), dist 1.8, x -0.5, y 1.6; towel\_16 at bbox [0.92, 0.0, 1.0, 0.11], at 30 degrees(CW), dist 1.6, x 0.7, y 1.3.

Image 19 contains objects: stove\_18 at bbox [0.21, 0.01, 1.0, 1.0], at 5 degrees(CW), dist 1.6, x 0.1, y 1.4; towel\_16 at bbox [0.24, 0.35, 0.41, 0.78], at 10 degrees(CCW), dist 1.4, x -0.2, y 1.3.

The correspond objects between image 16 and image 19 are stove\_18, towel\_16.

According to the objects motion changes, from image16 to image19, the camera motion is: translation (x 0.7, y 1.2), \cotsuchl{rotation (left 100 degrees, up 30 degrees)}.

As mentioned above, from image19 to image16, the camera motion is: translation (x -0.2, y 0.3), rotation (right 20 degrees, down 45 degrees).

Consequently, by accumulating the transformations, \cotfailhl{from image19 to image16, the camera motion is: translation (x 0.8, y 1.2), rotation (left 80 degrees, down 15 degrees)}.

** From image2 to image21 **

Image 2 contains objects: (kitchen cabinets)\_4 at bbox [0.6, 0.7, 0.7, 0.96], at 10 degrees(CW), dist 2.9, x 0.5, y 2.6.

Image 21 contains objects: (kitchen cabinets)\_4 at bbox [0.19, 0.71, 0.36, 1.0], at 10 degrees(CCW), dist 2.2, x -0.4, y 1.9; table\_25 at bbox [0.02, 0.82, 0.2, 1.0], at 25 degrees(CCW), dist 2.6, x -0.9, y 2.1.

The correspond objects between image 2 and image 21 is (kitchen cabinets)\_4.

According to the objects motion changes, from image2 to image21, the camera motion is: translation (x 2.0, y 1.2), rotation (\cotfailhl{left 45 degrees}, up 25 degrees).

As mentioned above, from image19 to image2, the camera motion is: translation (x 0.1, y 0.1), rotation (right 50 degrees, down 30 degrees).

Consequently, by accumulating the transformations, from image19 to image21, the camera motion is: translation (x 2.4, y -0.5), rotation (right 5 degrees, down 0 degrees).

In this phase, image 21 has introduced some new objects in its own coordinate system: (kitchen cabinets)\_4 at bbox [0.19, 0.71, 0.36, 1.0]; table\_25 at bbox [0.02, 0.82, 0.2, 1.0].

\cotfailhl{Transfer these objects from image 21 into image 19 coordination system: (kitchen cabinets)\_4 at 60 degrees(CW), dist 3.1, x 2.5, y 1.4; table\_25 at 50 degrees(CW), dist 3.2, x 2.3, y 1.9;}

%% file: figures/cot_case2_text_part3.tex
** From image21 to image25 **

Image 21 contains objects: doors\_0 at bbox [0.01, 0.23, 0.22, 0.85], at 25 degrees(CCW), dist 3.1, x -1.3, y 2.8.

Image 25 contains objects: doors\_0 at bbox [0.0, 0.36, 0.32, 0.97], at 20 degrees(CCW), dist 3.7, x -1.2, y 3.5.

The correspond objects between image 21 and image 25 is doors\_0.

According to the objects motion changes, from image21 to image25, the camera motion is: translation (x -1.0, y -0.2), rotation (\cotfailhl{right 30 degrees}, up 10 degrees).

As mentioned above, from image19 to image21, the camera motion is: translation (x 2.4, y -0.5), rotation (right 5 degrees, down 0 degrees).

Consequently, by accumulating the transformations, from image19 to image25, the camera motion is: translation (x 1.4, y -0.7), rotation (\cotfailhl{right 35 degrees}, up 10 degrees).

In this phase, image 25 has introduced some new objects in its own coordinate system: doors\_0 at bbox [0.0, 0.36, 0.32, 0.97].

\cotfailhl{Transfer these objects from image 25 into image 19 coordination system: doors\_0 at 40 degrees(CW), dist 3.6, x 2.3, y 2.8;}

~

Step 4: **Infer object spatial relation and answer the question**

After localizing the objects from questions in the scene layout from step2, we can know that "In image 19 coordination system: the stove\_18 at 5 degrees(CW), dist 1.6, x 0.1, y 1.4; \cotfailhl{the table\_25 at 50 degrees(CW), dist 3.2, x 2.3, y 1.9; the refrigerator\_10 at 5 degrees(CW), dist 2.4, x 0.1, y 2.3"}.

Considering the **view perspective** mentioned in the question, the right answer this question is the right answer is \cotfailhl{`C. back-right`}

</think> 
<answer>C</answer>